\pdfoutput=1

\documentclass[11pt]{article}

\usepackage[preprint]{acl}

\usepackage{times}
\usepackage{latexsym}

\usepackage[T1]{fontenc}

\usepackage[utf8]{inputenc}

\usepackage{microtype}

\usepackage{inconsolata}

\usepackage{graphicx}

\usepackage{amsmath}
\usepackage{amssymb}
\usepackage{mathtools}
\usepackage{amsthm}

\usepackage{dsfont}
\usepackage{mathrsfs}
\usepackage{booktabs}
\usepackage{multirow}
\usepackage{colortbl}
\usepackage{xcolor}

\usepackage{soul}
\usepackage{arydshln} 
\usepackage{subcaption}

\usepackage{tipa}

\definecolor{softblue}{rgb}{0.88, 0.95, 1.0} 
\definecolor{softyellow}{rgb}{0.98, 0.98, 0.82} 

\definecolor{plusgreen}{RGB}{17, 159, 88}

\usepackage{listings}
\usepackage{tcolorbox}
\usepackage{textcomp}

\DeclareMathOperator*{\argmax}{argmax}

\title{DOCE: Finding the Sweet Spot for Execution-Based Code Generation}

\author{
  \textbf{Haau-Sing Li\textsuperscript{1,2}},
  \textbf{Patrick Fernandes\textsuperscript{2,3,4}},
  \textbf{Iryna Gurevych\textsuperscript{1}},
  \textbf{André F. T. Martins\textsuperscript{2,4,5}},
\\
  \textsuperscript{1}UKP Lab, Department of Computer Science and Hessian Center for AI (hessian.AI), TU Darmstadt\\
  \textsuperscript{2}Instituto de Telecomunicações
  \textsuperscript{3}Language Technologies Institute, Carnegie Mellon University\\
  \textsuperscript{4}Instituto Superior Técnico, Universidade de Lisboa \& ELLIS Unit Lisbon
  \textsuperscript{5}Unbabel
\\
  \small{
    \textbf{Correspondence:} \href{hli@ukp.tu-darmstadt.de}{hli@ukp.tu-darmstadt.de}
    }
}

\begin{document}
\maketitle
\begin{abstract}
 A diverse set of decoding, reranking, and self-improvement procedures have been shown effective for code generation with LLMs. However, a unified framework that links and experimentally compares these methods is missing. We address this by proposing \textbf{D}ecoding \textbf{O}bjectives for \textbf{C}ode \textbf{E}xecution, a framework with candidate generation, $n$-best reranking, minimum Bayes risk (MBR) decoding, and self-debug as the core components. We then systematically study the contributions of these components through execution-based evaluation. Our findings suggest the importance of sampling with high temperature, reranking with execution-based methods using only a few high-quality unit test inputs, and robust self-debugging on multiple candidates. Specifically, we highlight filtering based on trial unit tests, a simple and effective strategy that has often been overlooked in prior works\footnote{Our code is available at \url{https://github.com/deep-spin/doce}}. 
\end{abstract}

\section{Introduction}

Despite impressive performances of large language models (LLMs) in code generation \citep{codex, alphacode, li2023starcoder, rozière2024code, luo2024wizardcoder, guo2024deepseek, lozhkov2024starcoder,zhu2024deepseek} on execution-based metrics \citep{apps,codex,mbpp,ds1000,livecodebench,bigcodebench} based on exploration of training approaches, predominant methods during inference still heavily rely on techniques such as greedy decoding, beam search, and $p$-nucleus sampling \citep{nucleus} followed directly by evaluation on execution accuracy.


An alternative approach involves \textit{sampling} multiple candidates with an LLM and selecting one using $n$-best reranking \citep{ng-etal-2019-facebook,bhattacharyya-etal-2021-energy,cobbe2021trainingverifierssolvemath,shen-etal-2021-generate-rank} or Minimum Bayes Risk (MBR) decoding \citep{eikema-aziz-2020-map,eikema-aziz-2022-sampling}, with potential improvements on candidates using the same LLM \citep{madaan2023selfrefine}. These methods have been systematically studied in domains like machine translation \citep{eikema-aziz-2020-map,freitag-etal-2022-high,fernandes-etal-2022-quality,farinhas-etal-2023-empirical}. However, despite the proposal of reranking methods, based on likelihoods \citep{coder-reviewer}, execution agreements \citep{shi-etal-2022-natural} or predicting execution accuracy \citep{inala2022faultaware,ni2023lever}, and improvement methods simulating the debugging process \citep{selfpair,chen2024teaching}, a systematic study that helps understand the effects and provide an apples-to-apples comparison of these methods is still missing.

\begin{figure*}[htbp]
    \centering
    \includegraphics[width=1.0\linewidth]{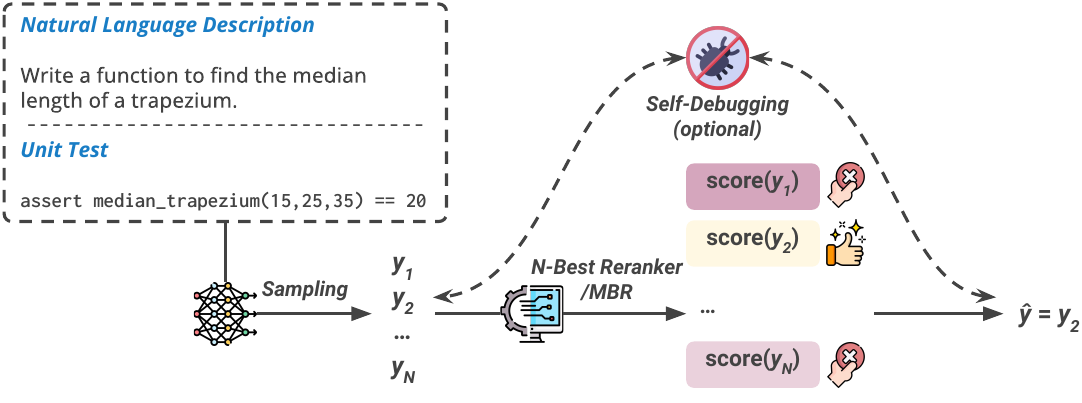}
    \caption
{The \textbf{D}ecoding \textbf{O}bjectives for \textbf{C}ode \textbf{E}xecution (\textbf{DOCE}) Framework. Firstly, multiple candidates are generated through sampling. Each candidate then is assigned a score using an $n$-best reranker or MBR, before the candidate with the highest score is returned. Self-Debug can be applied to multiple candidates before scoring as we propose, or the highest score candidate as proposed by \citet{chen2024teaching}.}
    \label{fig:framework}
\end{figure*}

Bridging this gap is crucial to current research. First of all, lately released open-source code LLMs \citep{rozière2024code,luo2024wizardcoder,guo2024deepseek,zhu2024deepseek} excelling on a range of tasks lead to a demand for a renewed understanding of the effects of sampling and reranking methods. Example methods include choosing a relatively low sampling temperature \citep{alphacode,shi-etal-2022-natural,huang-etal-2024-enhancing,chen2023codet}, which is originally suggested with LLMs having lower performances on these tasks \citep{incoder,codegen}, and filtering on trial unit tests \citep{alphacode}, a method originally proposed to retain sensible programs when a code LLM can hardly generate. Additionally, strong open-source code LLMs also allow a systematic study of sampling, reranking, and candidate improvement with a low budget compared to models requesting API calls \citep{openai2023gpt}. 
Last but not least, lately proposed neural metrics for code \citep{coder-reviewer,zhou-etal-2023-codebertscore, dong2023codescore} provide an opportunity to systematically understand the execution component of reranking compared with execution-free approaches.

Therefore, we propose a unified framework for execution-based code generation to improve decoding performance on execution-based metrics.
Our contributions are listed below:

\begin{itemize}
    \item We propose the \textbf{D}ecoding \textbf{O}bjectives for \textbf{C}ode \textbf{E}xecution (\textbf{DOCE})  framework for code generation (see Figure~\ref{fig:framework}), including candidate generation, reranking, and self-debug.
    \item We provide analysis on the number of generated candidates and sampling temperature. Specifically, sampling with previously unseen high temperatures allows high oracle and reranking performance.
    \item With experiments on reranking, we emphasize the importance of execution-based approaches. Specifically, our analysis reveals the importance of filtering based on trial unit tests, a commonly used technique whose effect has been rarely reported in previous works. 
    \item Execution-based methods lead to close-to-Oracle performance, with further improvements using Self-Debug. Effective and robust Self-Debug only happens when applied to multiple candidates before choosing the final candidate.
\end{itemize}


\section{Candidate Generation, Self-Debugging, and Reranking}


\subsection{Candidate generation}\label{sec:generation}
An LLM-based code generation model defines a probability distribution $p_{\theta_{\mathrm{LLM}}}(y|x)$ over the space of possible programs, conditioned on a description of the task $x$, where $\theta_{\mathrm{LLM}}$ refers to learned parameters of the LLM. An $x$ often contains a natural language description; some unit test cases, namely \textit{trial unit tests}, are used to regularize and generate the corresponding code.

Generated code is usually predicted using greedy decoding or $p$-nucleus sampling \citep{nucleus}. The latter approach allows obtaining diverse generated candidates $\mathcal{Y}$ compared to greedy decoding and is often accompanied by temperature sampling \citep{dabre-fujita-2021-investigating}. Previous studies \citep{shi-etal-2022-natural,chen2023codet} found that sampling leads to higher \textit{oracle} performance, denoted by \textit{pass@k}, assuming always choosing the best among \textit{k} candidates. However, deciding which candidate to pick is still unclear.

\subsection{Reranking}

We now assume access to $\mathcal{Y}$ containing $N$ generated candidates for a given $x$, as mentioned in Section~\ref{sec:generation}. To provide a final output $\hat{y}$, we explore how to rank and pick candidates for code generation effectively\footnote{With outputs that have the same score, we choose the output with the smallest index.}.

\subsubsection{\textit{n}-Best Reranking.}

In its simplest form, $n$-best reranking uses a single feature $f$ and ranks candidates, with the chosen candidate maximizing this feature:

\begin{equation}
    \label{eq:qe_argmax}
    \hat{y} = \argmax_{y' \in \mathcal{Y}} f(y').
\end{equation}

\paragraph{Likelihood-based Feature.} We can use the likelihood as the feature $f(y') = p_{\theta_{LLM}}(y' | x)$. In practice of code generation, the likelihood is often normalized by sequence length $p_{\theta_{LLM}}(y' | x)^{\frac{1}{len(y')}}$.
Alternatively, \citet{coder-reviewer} proposed an average of likelihood combined with a ``reviewer-likelihood'', normalized by sequence length as well:

\begin{equation}
    \label{eq:qe_argmax}
    f(y') = (p_{\theta_{LLM}}(y' | x) p_{\theta_{LLM}}(x | I, y'))^{\frac{1}{len(y')}},
\end{equation}

where $I$ is the set of the few-shot demonstrations.

\paragraph{Execution-based Feature.}
Previous works \citep{alphacode,chen2024teaching} have experimented with filtering based on trial unit tests. It filters out outputs that do not pass any single unit test $t(y) \in \{\text{pass,fail}\}$ of the set of trial unit tests $T_\text{trial}$:

\begin{equation}
    \label{eq:filtering}
    f(y') = \prod_{t \in T_\text{trial}} \mathds{1}_{t(y')=\text{pass}}.
\end{equation}

\paragraph{Scores from External Models.} We can also use a single model $\theta_{ext}$. Specifically, previous efforts include training models using execution success/failure as signals \citep{inala2022faultaware,ni2023lever}, given the input $x$ and generated candidate $y'$. During inference, a score $f_{\theta_{ext}}(x,y')$ is given for every generated candidate. 


\subsubsection{MBR Decoding}
While $n$-best list reranking can rank candidates based on \textit{reference-free} features, it is also possible to rank candidates with \textit{reference-based} metrics using MBR decoding. With a utility function $U(y^\star, y)$ measuring the \textit{similarity} between a candidate $y$ and some \textit{reference} (correct) code $y^\star$ MBR selects the candidate in $\mathcal{Y}$ that maximizes the \textit{expected} utility considering all candidates as possible references:

\begin{equation}
    \label{eq:mbr-mc-gain}
    \hat{y} =  \argmax_{y' \in \mathcal{Y}} \frac{1}{|\mathcal{Y}|} \sum_{y \in \mathcal{Y}} U(y,y').
\end{equation}

\paragraph{Execution-based Metrics.}
If we can feed unit test inputs to candidates, a basic utility function \citep{shi-etal-2022-natural,alphacode} is to consider if a candidate matches the outputs of some reference code with the set of unit tests used for evaluation $\mathcal{E}_{\mathcal{T}}$, the utility function is the exact match of outputs between the reference and the candidates
\begin{equation}
    \label{eq:mbr_exec}
    U(y,y') = \mathds{1}_{\mathcal{E}_{\mathcal{T}}(y) = \mathcal{E}_{\mathcal{T}}(y')}.
\end{equation}

In practice, this utility function reduces the MBR decision to a majority vote for the candidate that produces outputs that agree the most with other candidates' outputs \citep{wang2023selfconsistency}.

Note that in \citet{shi-etal-2022-natural} execution-based MBR both on the cases of using only inputs from trial unit tests only and from all unit tests used for evaluation. Since we decide to use these unit tests as filtering, we use all unit tests given but only their inputs, which aligns with previous practices \citep{shi-etal-2022-natural,ni2023lever,chen2024teaching}.

\paragraph{External Models for MBR.}
We can also use an external neural model \citep{zhou-etal-2023-codebertscore,dong2023codescore} used for evaluation $\theta_{ext}$ to compute $U_{\theta_{ext}}(x,y,y')$, obtaining the utility by inputting input $x$ and two generated candidates $y$ and $y'$.

\subsection{Self-Debugging}

Reranking is oracleed by the best possible candidate's performance. An approach for further quality improvement of generated codes is to \textit{debug} it with the same model used for generation \citep{chen2024teaching}. Given few-shot demonstrations $I$, an input $x$, a candidate $y'$, and its execution feedback with the trial unit tests $exec(y',T_\text{trial})$ , we obtain a corrected generation $y'_{d}$ using $\theta_\mathrm{LLM}$:

\begin{equation}
    \label{eq:debugging}
    y'_{d} =  \mathrm{debug}_{\theta_\mathrm{LLM}}(I, x, y', exec(y',T_\text{trial})).
\end{equation}



\subsection{\textbf{DOCE} Framework}

We define \textbf{DOCE} framework. Firstly, with only reranking and MBR, we have

\begin{equation}
    \label{eq:framework_1}
    \hat{y} =  \argmax_{y' \in \mathcal{Y}} \frac{f(y')}{|\mathcal{Y}|} \sum_{y \in \mathcal{Y}} U(y,y').
\end{equation}

\paragraph{Self-Debugging with Single Selected Candidate from Reranking.} If we only debug selected candidates from reranking, namely \textbf{SD-1} \citep{chen2024teaching}, we have the final code answer $\textit{debug}_{\theta}(x, \hat{y}, T_\text{trial})$.

\paragraph{Self-Debugging on All Candidates before Reranking.} We can also perform self-debugging on \textit{all} generated candidates before reranking, namely \textbf{SD-Multi}. We presume this outperforms self-debugging on one candidate with more candidates passing trial unit tests, leading to better candidates to be reranked and possibly better reranking results. The final code answer is $\argmax_{y'_{d} \in \mathcal{Y}_{d}} \frac{f(y'_{d})}{|\mathcal{Y}_{d}|} \sum_{y_{d} \in \mathcal{Y}_{d}} U(y_{d},y'_{d})$.

\section{Experiments}


\subsection{Datasets}

We first conduct experiments using two widely recognized execution-based datasets: HumanEval \citep{codex} and MBPP \citep{mbpp}. We employ the sanitized, higher-quality subset of the MBPP dataset, namely MBPP-S. We utilize EvalPlus \citep{evalplus} that includes more than 35 times unit tests than the original benchmark, ensuring rigorous execution-based evaluation. We then extend experiments to competitive programming using LiveCodeBench \cite{livecodebench}, a dataset containing up-to-date competitive programming problems to avoid data contamination. See Appendix~\ref{sec:dataset_stats} for further details.


\subsection{Code LLMs}

We generate candidates with the CodeLlama-\{7B,13B\}-Instruct \citep{rozière2024code} and DeepSeekCoder-\{6.7B,V2-Lite\}-Instruct \citep{guo2024deepseek,zhu2024deepseek}. We use instruction-tuned models that enable simple processing of generated codes for both candidate generation and Self-Debug.


\subsection{Candidate Generation}

We generate 5 to 50 candidates for each task. We first explore sampling temperature by varying it between 0.2 and 2.0 with $p$-nucleus sampling $p$ 0.95, before deciding what sampling temperature we use for subsequent experiments. With MBPP-S and LiveCodeBench, we vary temperature between 0.2 and 1.8. When generating multiple candidates, we use the open-source vLLM \citep{kwon2023efficient} for fast inference. To answer research questions related to candidate generation and reranking, we present the results from CodeLlama-7B-Instruct. 

\subsection{Reranking}

For $n$-best reranking, we study three metrics types: likelihood-based, execution-based, and neural. For likelihood-based metrics, we use the likelihood feature and CoderReviewer, both normalized by sequence length. 
For execution-based metrics, we consider filtering based on trial unit tests.
For neural metrics, we use CodeScore \citep{dong2023codescore} as it is so far the only publicly available general-purpose neural metric trained to predict execution accuracy. We do not consider task-specific and model-specific neural metrics \citep{inala2022faultaware,ni2023lever}.

For MBR, we consider execution-based metrics and neural metrics. Specifically, among execution-free metrics, we include CodeBertScore \citep{zhou-etal-2023-codebertscore}, which is not trained with execution signals, and CodeScore \citep{dong2023codescore}, which is trained with execution signals. For execution-based metrics, we consider MBR-Exec \citep{shi-etal-2022-natural} mentioned in Equation~\ref{eq:mbr_exec}. 
Note that we assume access to the \textit{inputs} of unit tests used for evaluation, which aligns with one setting of MBR-Exec \citep{shi-etal-2022-natural} and the setting of LEVER \citep{ni2023lever} and Self-Debug \citep{chen2024teaching}.

\subsection{Self-Debug}

We consider the simple setting of Self-Debug proposed by \citet{chen2024teaching}, using only unit test (UT) feedback, i.e., the feedback obtained from execution when a generated candidate is tested on trial unit tests. We only consider this simple setting because 1) our main focus is to test generally the impact of Self-Debug on improving the oracle's performance, 2) it does not require steps of generation using the LLM other than Self-Debug, thus requires less computation for inference and 3) it is the feedback that gives the largest gain to execution accuracy post-debugging according to \citet{chen2024teaching}. We perform 3 rounds of Self-Debug for candidates generated with sampling because \citet{chen2024teaching} demonstrated that most debugging can be finished in 3 rounds. In each round, the LLM debugs the candidates generated in the last round.

\section{Results}

In this section, we provide results and discuss the effects of sampling, reranking, and Self-Debug. Firstly, we analyze the impact of candidate generation on the model's \textit{oracle} performance (assuming we pick the best possible candidate) and the performance of execution-based reranking methods. We then compare different reranking methods and discuss the importance of execution for reranking. With the answers to these questions in mind, we study the effect of Self-Debug and discuss the best way to combine all these methods to improve reranking performance. We provide a subset of all our experimental results to answer these research questions. For results on more experiments, see Appendix~\ref{sec:complete_results}.

\subsection{Candidate Generation}\label{sec:results_generation}

First, we study the impact of candidate generation varied by candidate numbers and sampling temperature using CodeLlama-7B-Instruct. We provide average execution accuracies, reranking results using MBR-Exec with and without filtering based on trial unit tests, and estimates of reranking oracle using \textit{Pass@k}. We hypothesize that filtering gives reranking better alignment with the (potential) ground-truth function as it is a strong signal.


\begin{figure}[tbp]
    \centering
    \begin{subfigure}{0.165\textwidth}
        \centering
        \includegraphics[width=\linewidth]{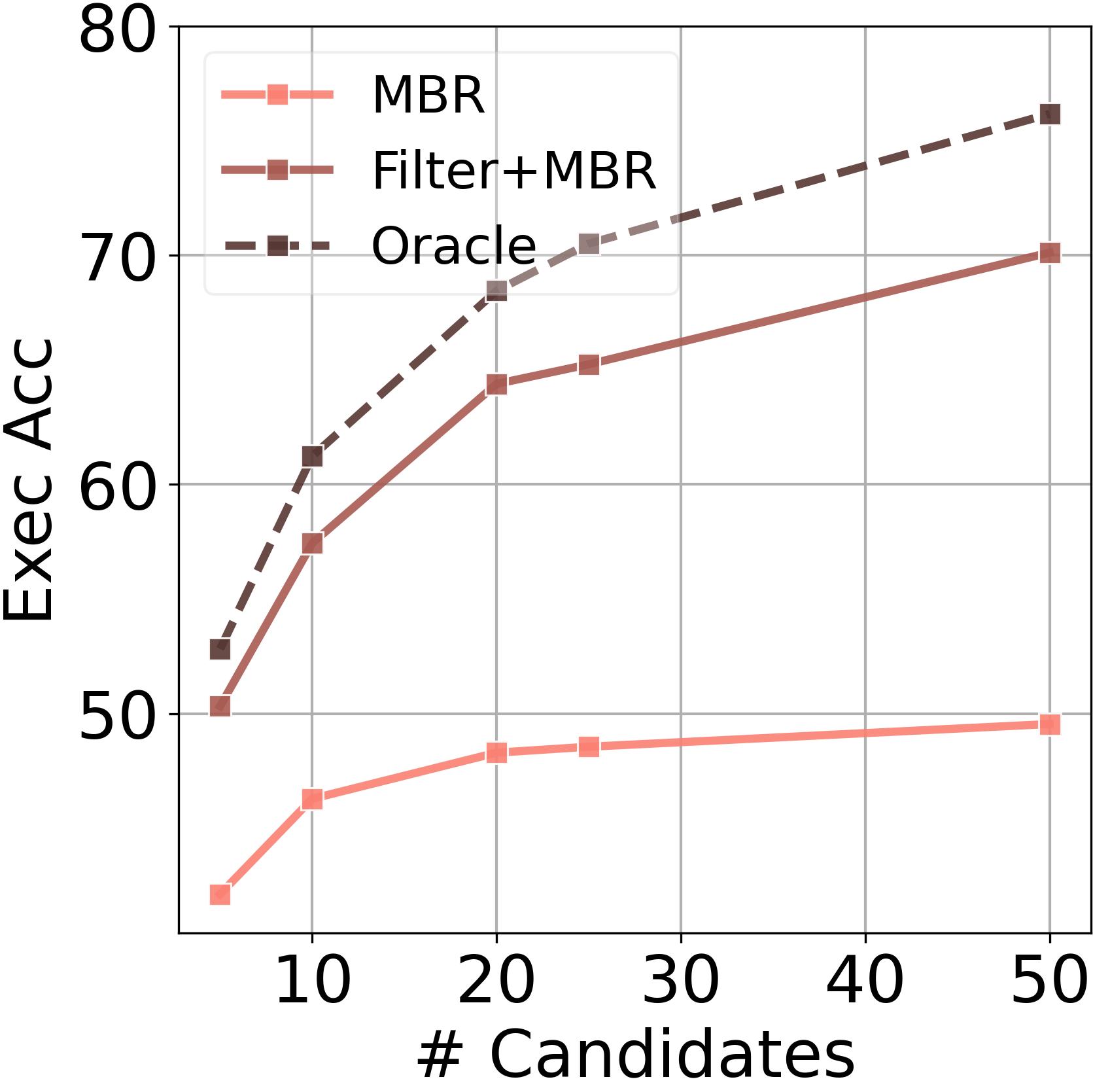}
        \caption{HumanEval+}
        \label{fig:num_candidates_humaneval_plus}
    \end{subfigure}\hfill
    \begin{subfigure}{0.154\textwidth}
        \centering
        \includegraphics[width=\linewidth]{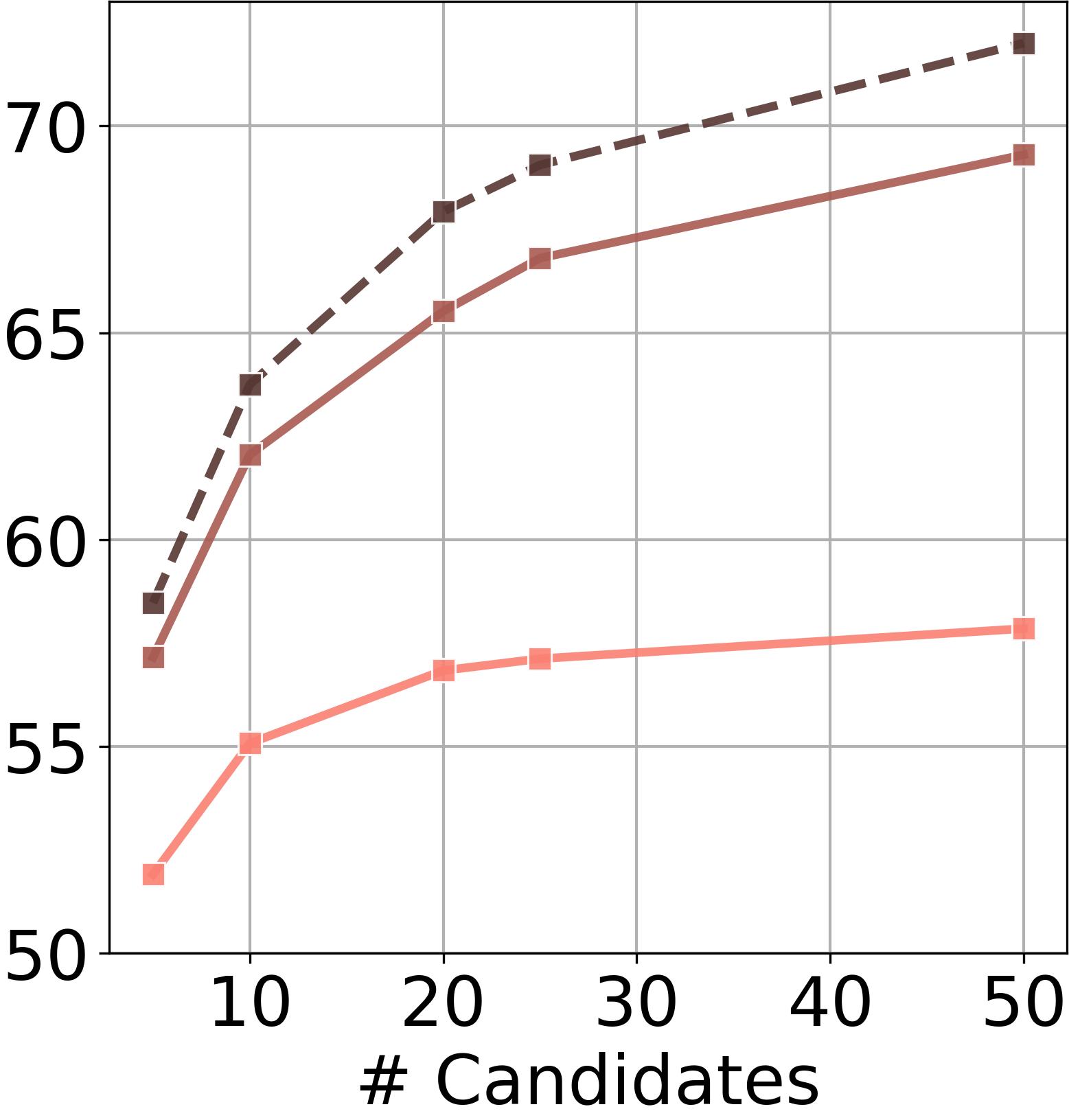}
        \caption{MBPP-S+}
        \label{fig:num_candidates_mbpp_plus}
    \end{subfigure}\hfill
    \begin{subfigure}{0.154\textwidth}
        \centering
        \includegraphics[width=\linewidth]{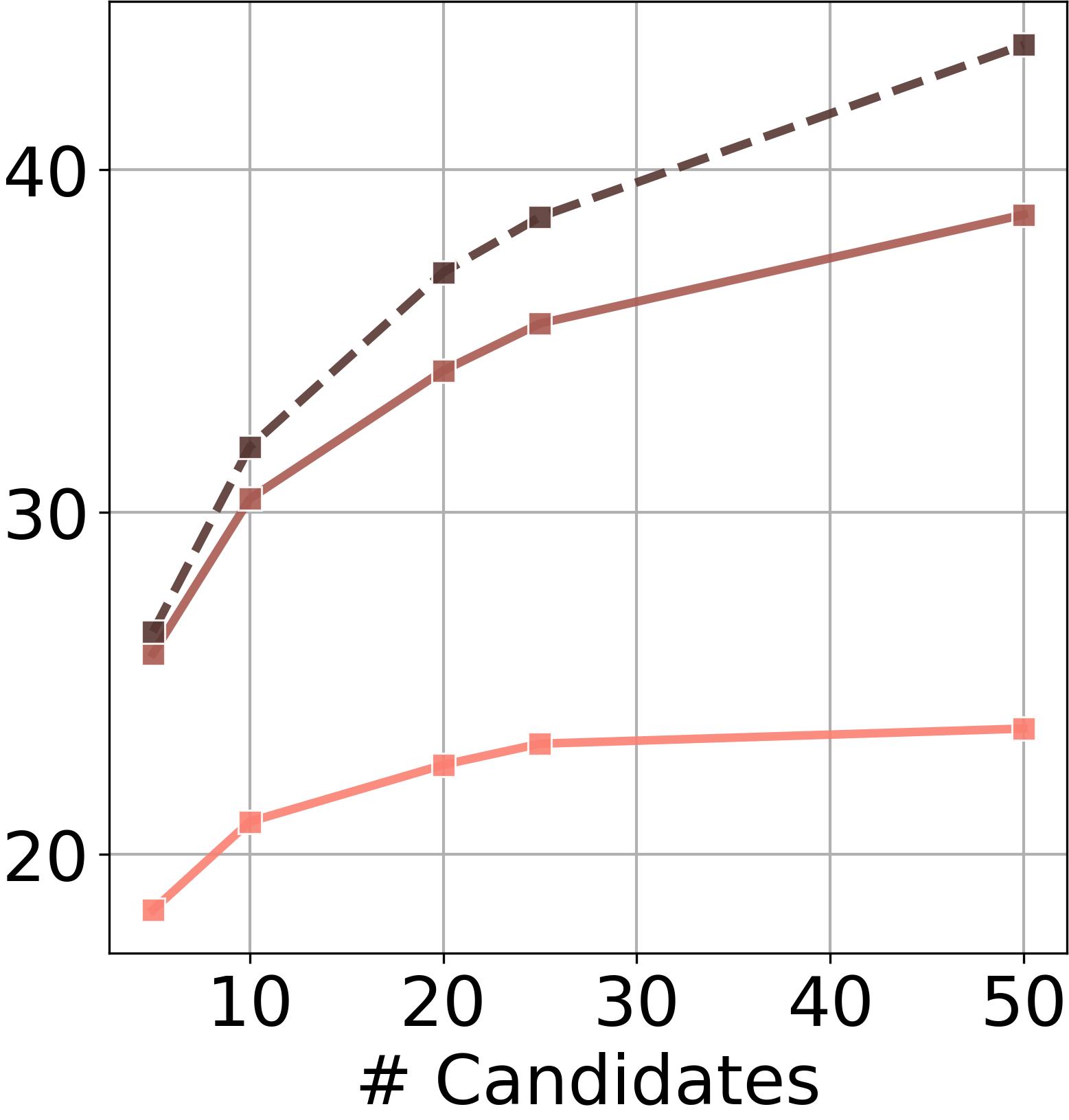}
        \caption{LiveCodeBench}
        \label{fig:num_candidates_lcb}
    \end{subfigure}

    \caption{Performance of reranking and oracle performance over different numbers of generated candidates using CodeLlama-7B-Instruct with temperature 1.6 for HumanEval+ and MBPP+, and 1.2 for LiveCodeBench. Results are averaged across at least 2 runs for LiveCodeBench and 4 runs for the rest.
    }
    \label{fig:num_candidates}
\end{figure}

\paragraph{What is the impact of the number of candidates?} We generate candidates using CodeLlama-7B-Instruct by setting temperature 1.6 and \textit{p} 0.95. 

Figure~\ref{fig:num_candidates} shows the oracle performance and the reranking performance. First of all, when filtering is not included, MBR-Exec only shows moderate improvement with more generated candidates across different numbers of unit tests for evaluation and two benchmarks, aligning with the analysis from \citet{shi-etal-2022-natural}. However, with filtering, MBR-Exec shows considerable improvements over the case without filtering, with performance scaling much better with more candidates generated, which aligns with our hypothesis. Secondly, the performance of MBR-Exec with filtering consistently approaches the oracle performance even as we increase the number of candidates, providing further evidence supporting our hypothesis. 

\begin{figure}[tbp]
    \centering
    \begin{subfigure}{0.16\textwidth}
        \centering
        \includegraphics[width=\linewidth]{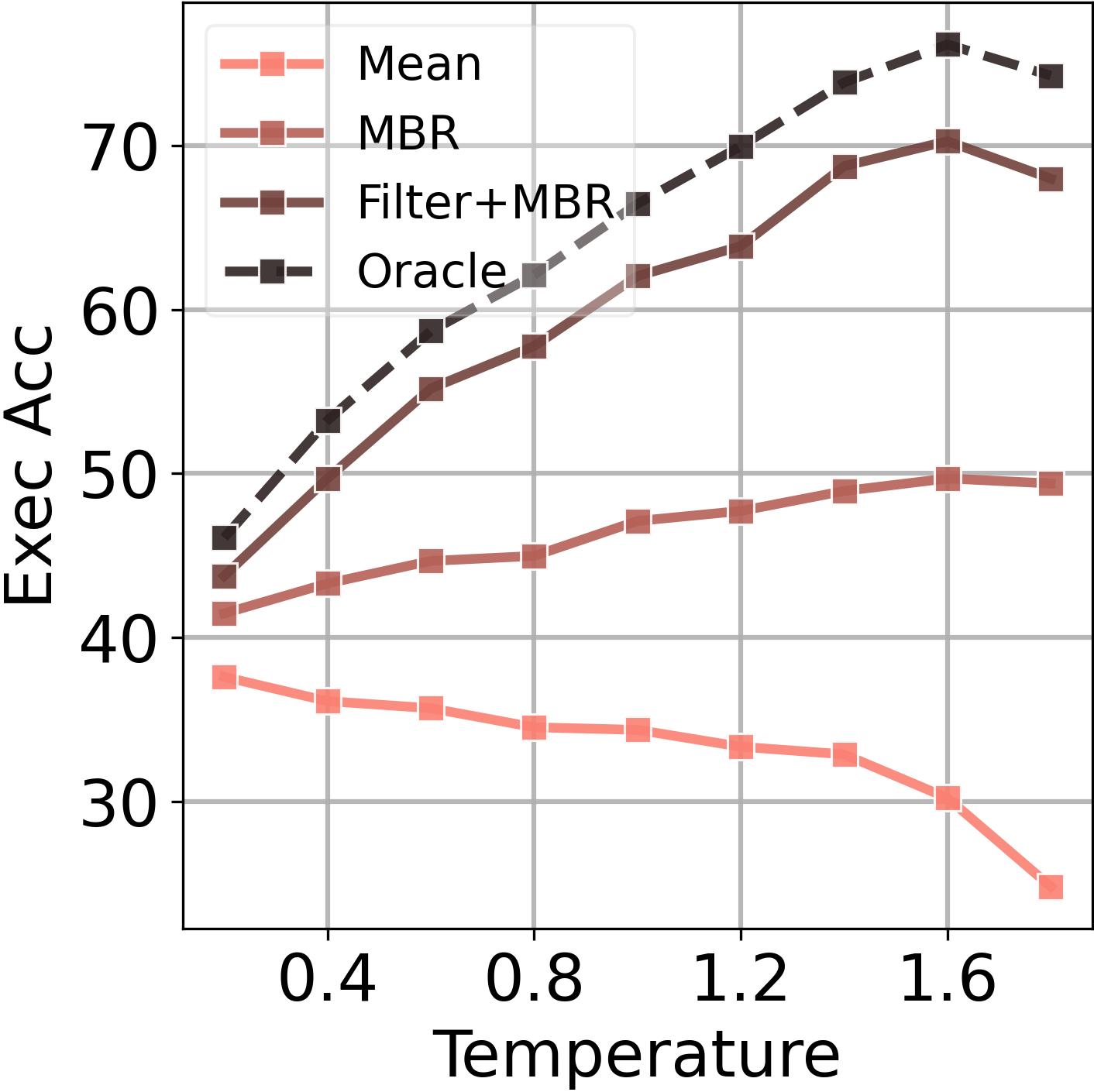}
        \caption{HumanEval+}
        \label{fig:temp_humaneval}
    \end{subfigure}\hfill
    \begin{subfigure}{0.16\textwidth}
        \centering
        \includegraphics[width=\linewidth]{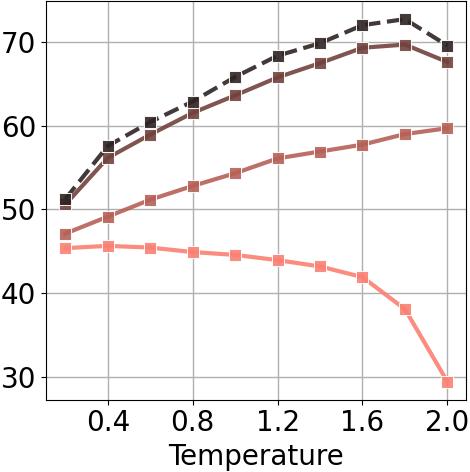}
        \caption{MBPP-S+}
        \label{fig:temp_mbpp}
    \end{subfigure}\hfill
    \begin{subfigure}{0.16\textwidth}
        \centering
        \includegraphics[width=\linewidth]{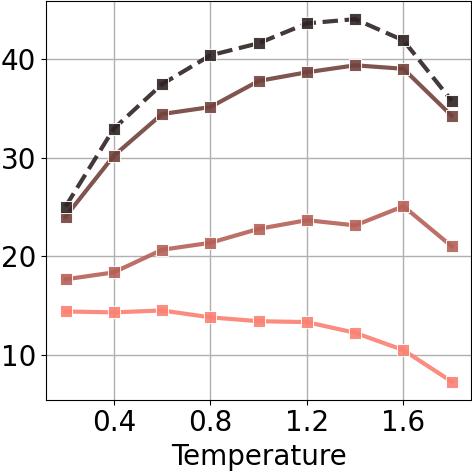}
        \caption{LiveCodeBench}
        \label{fig:temp_lcb}
    \end{subfigure}
    \caption{Performance of reranking and oracle over sampling temperatures using CodeLlama-7B-Instruct with 50 generated candidates over 4 runs.}
    \label{fig:temperature}
\end{figure}

\paragraph{What is the impact of sampling temperature?} We compare generations with temperatures using CodeLlama-7B-Instruct with nucleus $p$ 0.95 50 generated candidates. We show results in Figure~\ref{fig:temperature}.  

We find that sampling with a previously unseen high gives a higher oracle of reranking performance, peaking between 1.6 and 1.8 for HumanEval+ and MBPP+, and between 1.2 and 1.4 for LiveCodeBench. This suggests that sampling with a lower temperature is suggested when applied to tasks that are more difficult, e.g. competitive programming when compared to basic programming. However, sampling in practice generally can be done with a higher temperature than previously suggested \citep{codex,alphacode,shi-etal-2022-natural,evalplus,livecodebench}. Additionally, when filtering is applied, reranking performance approaches the oracle, which again supports our aforementioned hypothesis.

For later experiments, we choose the number of generated candidates to be  50. We adopt sampling temperature 1.6 for HumanEval and MBPP-S and 1.2 for LiveCodeBench, with nucleus $p$ 0.95. For DeepSeek models, we use temperature 1.2\footnote{With temperature >1.2 on DeepSeek models, sampling is more likely to generate token indices that are not defined in the vocabulary, see \url{https://github.com/vllm-project/vllm/pull/3685}.}.

\begin{table}[!tb]
\small
\centering
\scalebox{1}{
\begin{tabular}{lccc@{}}
\toprule
& \multirow{1}{*}{HE+} & \multirow{1}{*}{MBPP-S+}  & \multirow{1}{*}{LCB} \\\midrule
Random & 30.2 & 41.9 & 13.3 \\ 
Greedy & 39.0 & 44.8 & 13.4 \\ 
Oracle & 76.2 & 72.0 & 44.1 \\
\midrule
\multicolumn{3}{c}{N-Best Reranking} \\ \midrule
LL & 38.7 & 42.6 & 16.3 \\
CR & 40.5 & 43.1 & 15.9 \\
CS & 30.2 & - & 14.1  \\

\rowcolor{softblue}
Filter & 59.7 & \textbf{60.8} & 29.6  \\
\rowcolor{softblue}
Filter + LL & 60.4 & 59.7 & \textbf{32.5}  \\
\rowcolor{softblue}
Filter + CR & \textbf{61.3} & \underline{60.2} & \underline{32.2}  \\
\rowcolor{softblue}
Filter + CS & \underline{61.0} & - & 30.2 \\ 
\midrule

\multicolumn{3}{c}{MBR} \\ \midrule
MBR-CBS & 35.5 & 45.4 & - \\
MBR-CS & 31.9 & - & - \\
MBR-Exec & 49.5 & 57.8& 23.7 \\

\midrule
\multicolumn{3}{c}{N-Best Reranking + MBR} \\ \midrule
LL + MBR-Exec & 49.4 & 58.1& 24.2  \\
CR + MBR-Exec & 48.9 & 58.7 & 24.2 \\
\rowcolor{softblue}
Filter + MBR-CBS & 60.1 & 61.3& -  \\
\rowcolor{softblue}
Filter + MBR-CS & 59.9 & -& - \\
\rowcolor{softblue}
Filter + MBR-Exec & \underline{70.1} & \underline{69.3}& \textbf{38.7}  \\
\rowcolor{softblue}
Filter + LL + MBR-Exec & \textbf{70.6} & 69.2 & \underline{33.4}  \\
\rowcolor{softblue}
Filter + CR + MBR-Exec & 70.0 & \textbf{69.4} & 33.2  \\
\bottomrule
\end{tabular}}
\caption{Comparison of reranking methods on HumanEval (HE)+, MBPP-S+, and LiveCodeBench(LCB). For N-Best Reranking, we compare Likelihood (LL), Coder-Reviewer (CR), and CodeScore (CS). For MBR, we compare CodeScore, CodeBertScore (CBS), and MBR-Exec. We highlight \textbf{best} and \underline{second best} reranking results of the class of reranking methods.}
\label{tab:compare_reranking}
\end{table}

\subsection{Comparing Reranking Methods}\label{sec:compare_reranking}

With the best setting we observe from Section~\ref{sec:results_generation}, we now try to provide an apples-to-apples comparison between different reranking methods that either directly utilize execution, predict executions, or utilize other information to understand the effect of filtering or execution more generally in reranking and provide suggestions for reranking methods.

We conduct experiments with $n$-best reranking and MBR to compare different raking approaches with and without filtering based on trial unit tests. We show results in Table~\ref{tab:compare_reranking}. 

\paragraph{Filtering is highlighted.}
According to the group ``N-Best Reranking'' (see Table~\ref{tab:compare_reranking}), filtering based on trial unit tests gives the largest gain from the random baseline compared to the likelihood feature, CoderReviewer, and CodeScore.

Not only is filtering powerful in itself, but it also shows further improvements when combined with $n$-best reranking methods. Therefore, we again emphasize the importance of filtering based on unit tests for reranking.

\paragraph{MBR-Exec with filtering still rocks.} \citet{coder-reviewer} shows that Coder-Reviewer outperforms MBR-Exec with a weaker type of filtering, i.e., retaining a code candidate if it can produce outputs given the inputs provided in trial unit tests. In our experimental settings that include a more sensible definition of filtering and assume access to all unit test inputs for evaluation, likelihood feature and CoderReviewer underperform MBR-Exec with filtering. Additionally, when combined with filtering and MBR-Exec, they do not substantially outperform MBR-Exec with filtering either, suggesting that the likelihood feature and CoderReviewer \citep{coder-reviewer} do not substantially help reranking when sufficient direct execution is applied, giving close-to-oracle performances.

Additionally, when combined with filtering, MBR based on neural metrics does not significantly outperform filtering itself on basic programming tasks (HumanEval and MBPP). Therefore, we do not recommend compared to using execution-based MBR. Acknowleding the huge computational cost of applying neural metrics for MBR, we do not apply it to competitive programming with long generated sequences.

\subsection{Efficient Reranking with Fewer Unit Tests}

While MBR-Exec performs well with filtering, it requires access to execution unit test inputs, which is impractical considering the execution cost. Therefore, we ask: \textit{how many high-quality unit tests should we execute candidates on?} We experiment with 50 generated candidates using MBR-Exec using increasing unit tests. Note that as we use the extended unit tests \citep{evalplus}, we should also evaluate cases with the extended unit tests. We do not test this on LiveCodeBench due to the limit of unit tests. Additionally, when the number of unit tests is lower than those given in the original dataset, we only use unit tests from the original dataset instead of the extended partition. 

\begin{figure}[tbp]
    \centering
    \begin{subfigure}{0.243\textwidth}
        \centering
        \includegraphics[width=\linewidth]{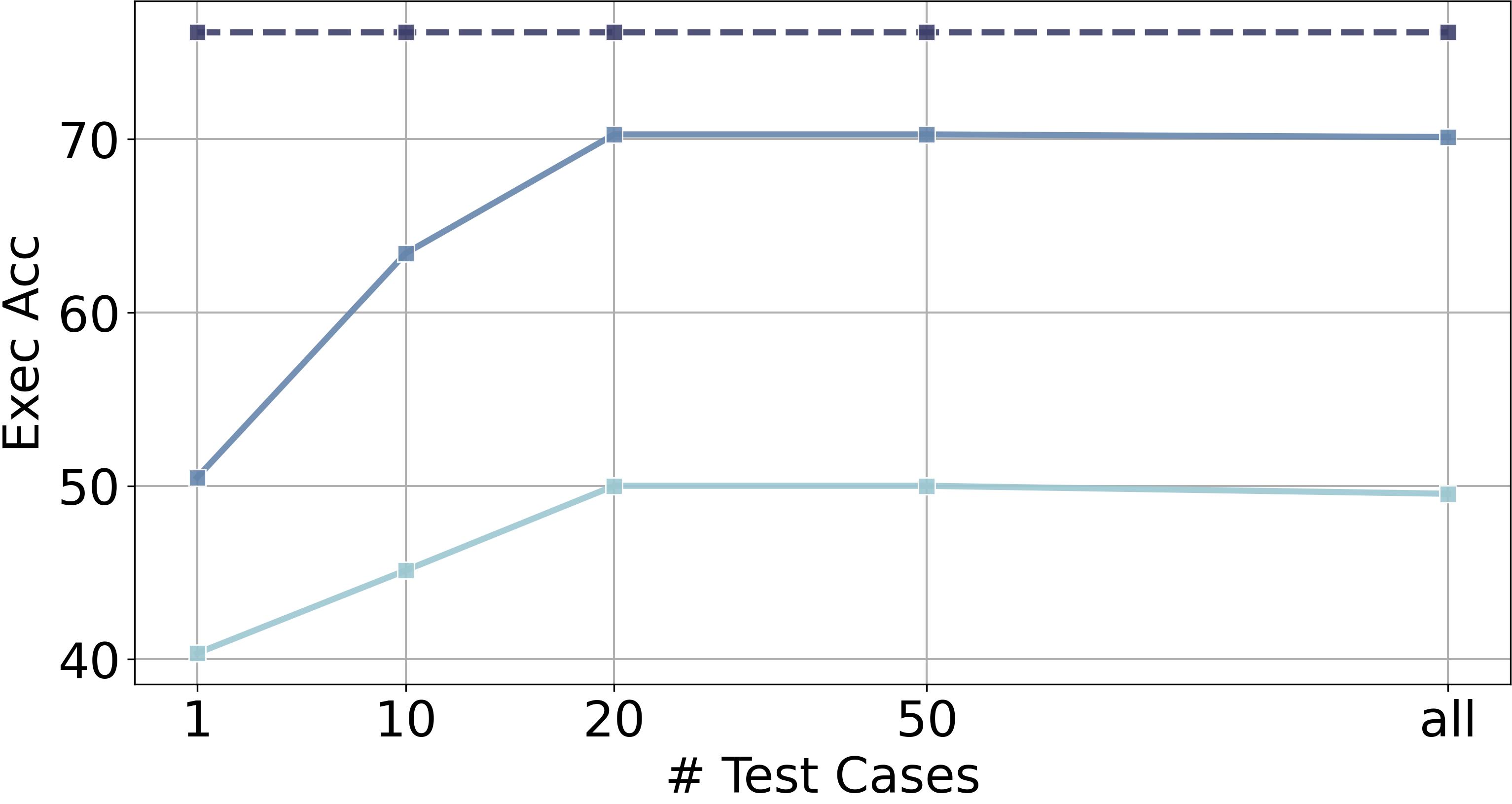}
        \caption{HumanEval+}
        \label{fig:num_uts_humaneval_plus}
    \end{subfigure}
    \begin{subfigure}{0.232\textwidth}
        \centering
        \includegraphics[width=\linewidth]{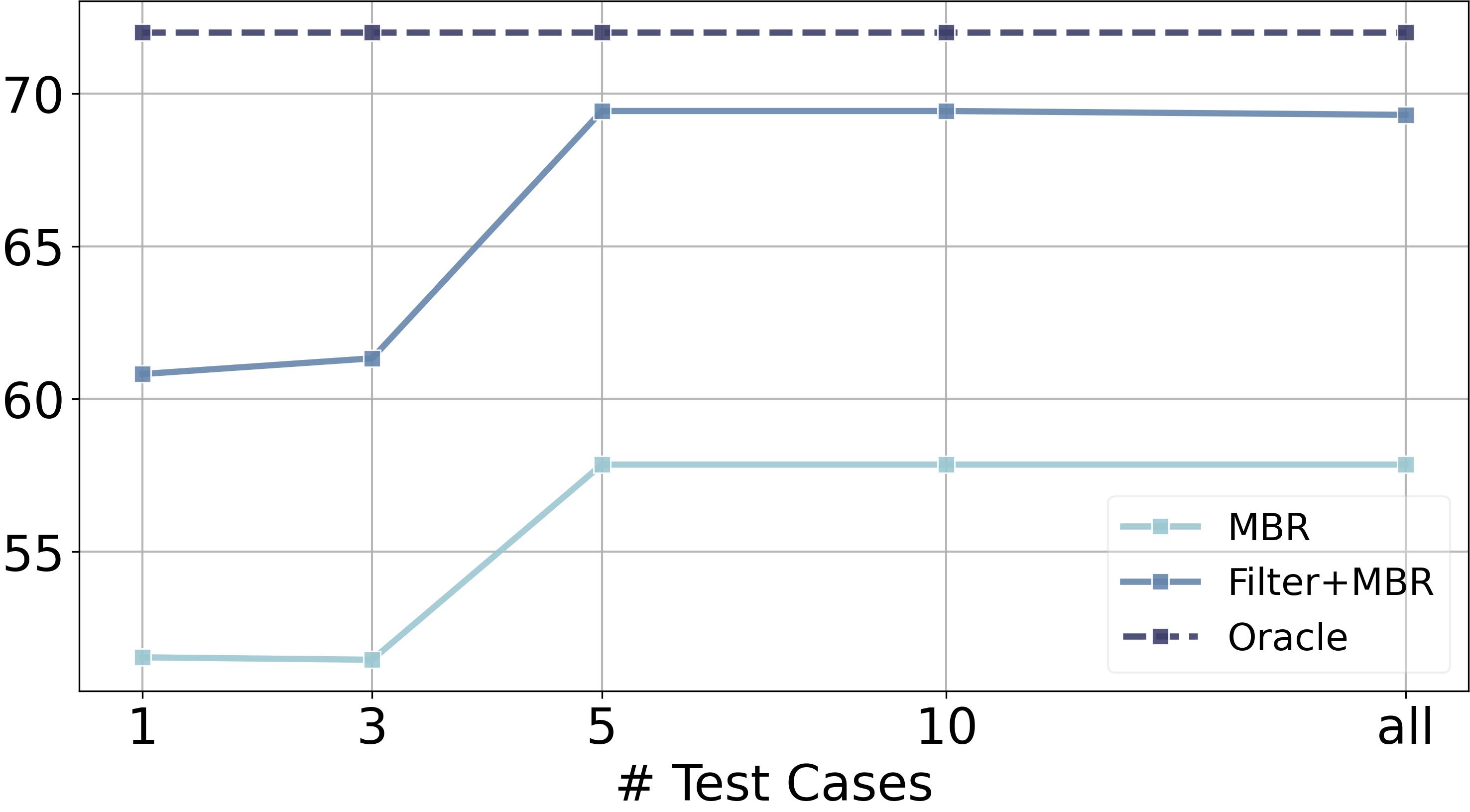}
        \caption{MBPP-S+}
        \label{fig:num_uts_mbpp_plus}
    \end{subfigure}
    \caption{Performance of MBR-Exec with fewer unit tests.}
    \label{fig:num_uts}
\end{figure}

\paragraph{With only a few more unit tests than from the original partition, MBR-Exec with filtering reaches peak performance.} According to Figure~\ref{fig:num_uts}, for HumanEval+, we only need 20 unit tests, while for MBPP-S+, MBR-Exec with 5 unit tests already gives optimal performance. This suggests that MBR-Exec can already reach its optimal performance with a few unit tests provided.

\subsection{Improving Oracle with Self-Debug}\label{sec:sd}

\begin{table*}[!tb]
\small
\centering
\scalebox{0.89}{
\begin{tabular}{lccccccccc@{}}
\toprule
\multirow{2}{*}{Model} & & \multirow{1}{*}{HumanEval+} & & & \multirow{1}{*}{MBPP-S+} & & & \multirow{1}{*}{LiveCodeBench} &\\
 &  MBR & SD-1 & SD-Multi & MBR & SD-1 & SD-Multi & MBR & SD-1 & SD-Multi \\ \midrule
\multicolumn{10}{c}{$\sim$7B Scale} \\ \midrule

CL-7B-Instruct& \underline{49.5} & \underline{49.5} & \textbf{50.0} & 57.8 & \underline{59.1} & \textbf{60.4} & \underline{23.7} & 23.3 & \textbf{24.0} \\

\rowcolor{softblue}
\quad + Filtering & \underline{70.1} & \underline{70.1} & \textbf{72.3} & \underline{69.3} & 69.2 & \textbf{72.1} & \underline{38.7} & 38.0 & \textbf{39.9} \\

DS-6.7B-Instruct& 85.5 & \underline{86.0} & \textbf{87.0} & 78.6 & \underline{78.8} & \textbf{79.3} & \underline{30.6} & \underline{30.6}  & \textbf{31.3} \\
\rowcolor{softblue}
\quad + Filtering & \textbf{90.7} & \underline{90.4} & \underline{90.4} & \underline{82.1} & 81.7 & \textbf{82.8} & \underline{41.9} & 41.2 & \textbf{42.0}\\

\midrule
\multicolumn{10}{c}{$\sim$13B Scale} \\ \midrule
CL-13B-Instruct& 59.6 & \underline{60.1} & \textbf{61.4} & \underline{67.2} & 67.0 & \textbf{68.4} & 27.0 & \textbf{27.6} & \textbf{27.6}\\
\rowcolor{softblue}
\quad + Filtering & 76.1 & \underline{76.5} & \textbf{77.6} & \underline{74.7} & 74.2 & \textbf{76.2} & 40.6 & \underline{40.8} & \textbf{46.8} \\

\midrule
\multicolumn{10}{c}{$\sim$16B Scale} \\ \midrule
DS-V2-Lite-Instruct& \underline{83.5} & 83.4 & \textbf{85.1} & 80.1 & \underline{81.0} & \textbf{81.4} & \underline{45.8} & 45.4 & \textbf{46.6}\\
\rowcolor{softblue}

\quad + Filtering & \textbf{89.6} & 88.4 & \textbf{89.6} & 81.8 & \underline{82.2} & \textbf{82.8} & \textbf{58.7} & 58.3 & \underline{58.5}\\

\bottomrule
\end{tabular}}%
\caption{Comparison of self-debugging methods with 50 candidates generated by CodeLlama-\{7,13\}B-Instruct and DeepSeekCoder-\{6.7B,V2-Lite\}-Instruct, and debugged over \{1, \textbf{Multi}\} candidates. We also provide the upper bound after debugging. Results are averaged across 2 runs for LiveCodeBench and 4 runs for the rest.}
\label{tab:sd_results}
\end{table*}

Applying MBR-Exec with filtering gives close-to-oracle performances, suggesting that improving the oracle performance with Self-Debug \citep{chen2024teaching} is a more sensible choice than providing better reranking methods. In this section, we first analyze the improvement in oracle performance. We then compare our proposed SD-Multi and SD-1 proposed by \citet{chen2024teaching}. For SD-Multi, we only use results with one round of debugging, while we use results with 3 rounds of debugging for SD-1.

\begin{figure}[tbp]
    \centering
    \begin{subfigure}{0.165\textwidth}
        \centering
        \includegraphics[width=\linewidth]{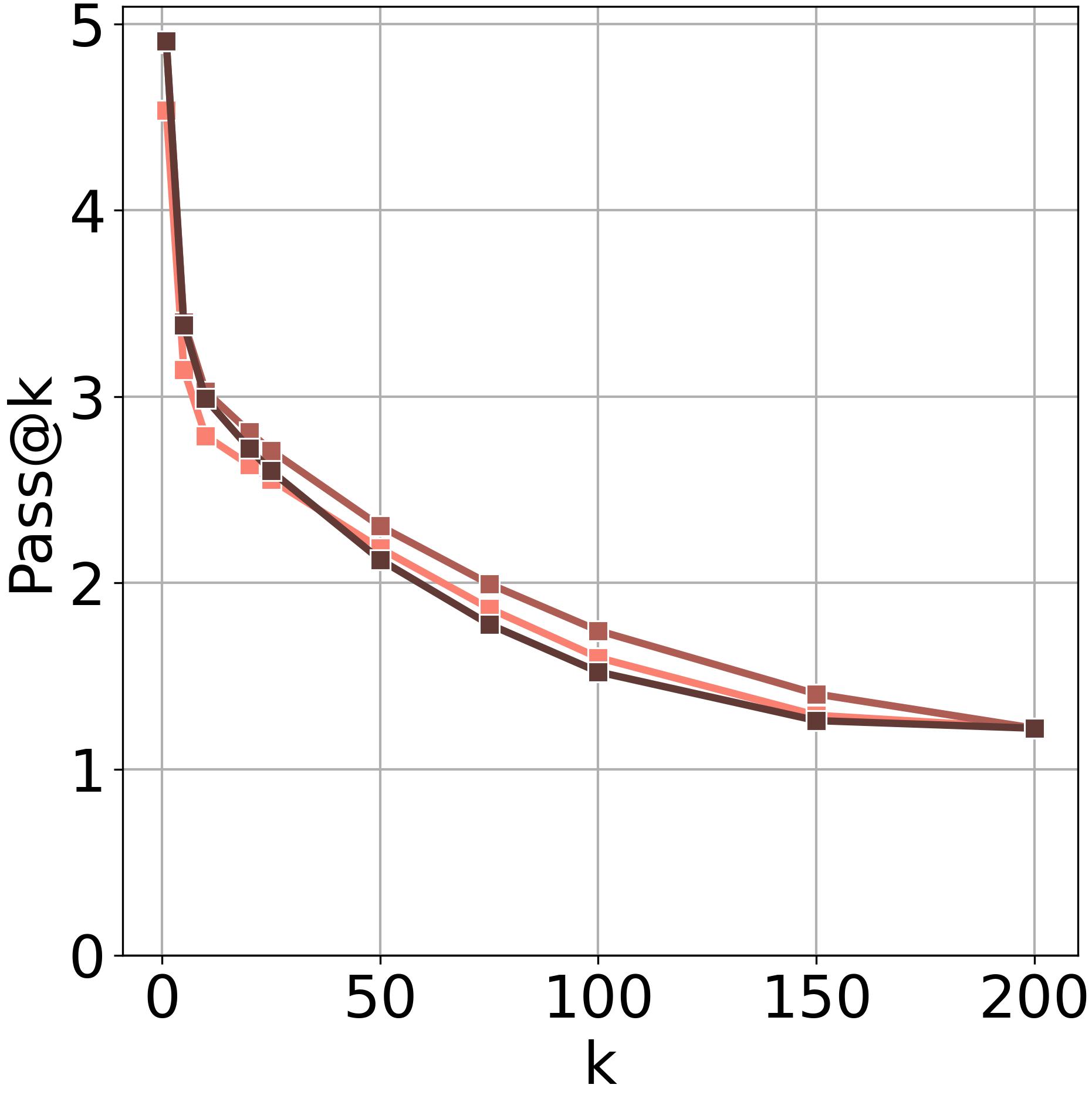}
        \caption{HumanEval+}
        \label{fig:sd_upperbound_humaneval_plus}
    \end{subfigure}\hfill
    \begin{subfigure}{0.154\textwidth}
        \centering
        \includegraphics[width=\linewidth]{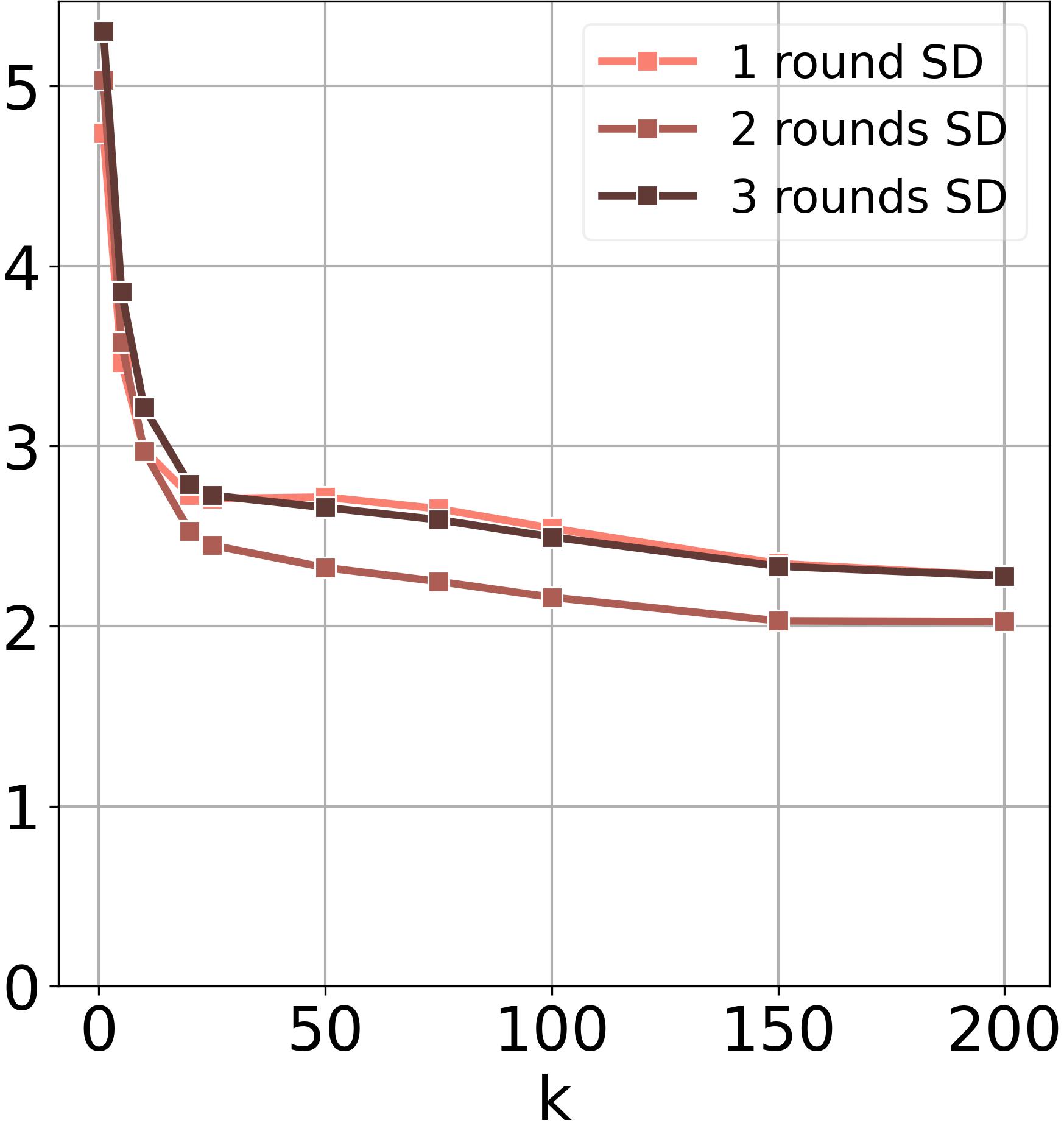}
        \caption{MBPP-S+}
        \label{fig:sd_upperbound_mbpp_plus}
    \end{subfigure}\hfill
    \begin{subfigure}{0.154\textwidth}
        \centering
        \includegraphics[width=\linewidth]{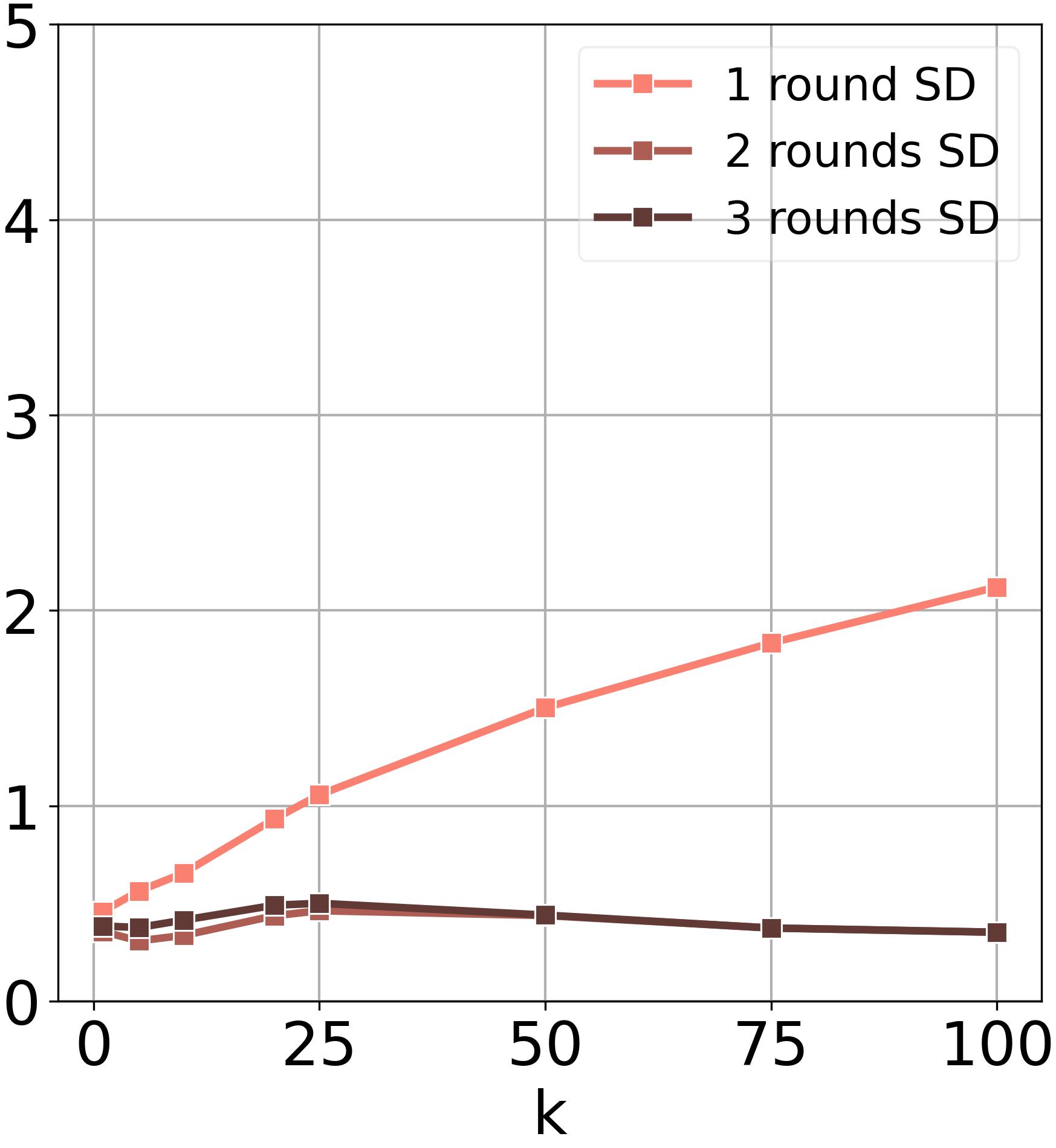}
        \caption{LiveCodeBench}
        \label{fig:sd_upperbound_lcb}
    \end{subfigure}
    \caption
{Improvement in Pass@k of CodeLlama-7B-Instruct after Self-Debug compared to no Self-Debug applied.}
    \label
{fig:sd_upperbound}
\end{figure}

\paragraph{Self-Debug improves oracle performance of reranking, and one single round is enough.} According to Figure~\ref{fig:sd_upperbound}, Self-Debug helps improve the oracle of reranking, represented by the improvement on \textit{Pass@k} across different \textit{k}s. However, later rounds of Self-Debug do not show substantial improvement in the oracle compared to the first round. Moreover, Self-Debug with one single round LiveCodeBench shows larger \textit{Pass@k} improvement with larger \textit{k}s. A sensible guess for this trend is related to the higher level of difficulty of problems in LiveCodeBench, with Self-Debug regenerating few sensible programs on problems that originally had no sensible candidates. The disappearance of this improvement suggests that since most code LLMs have not been trained extensively on debugging objectives, Self-Debug regenerates candidates close to greedy decoding.

\paragraph{SD-Multi outperforms SD-1.} According to Table~\ref{tab:sd_results}, SD-Multi consistently outperforms SD-1 in most experiments, with better-performing models having smaller margins of improvement with Self-Debug. The only exceptions we find are on experiments with candidates generated and self-debugged with DeepSeekCoder-\{6.7B, V2-Lite\}-Instruct, where either the task is considered more difficult compared to basic programming, or that the baseline performance with MBR-Exec is already above 91.

\begin{figure}[tbp]
    \centering
    \begin{subfigure}{0.165\textwidth}
        \centering
        \includegraphics[width=\linewidth]{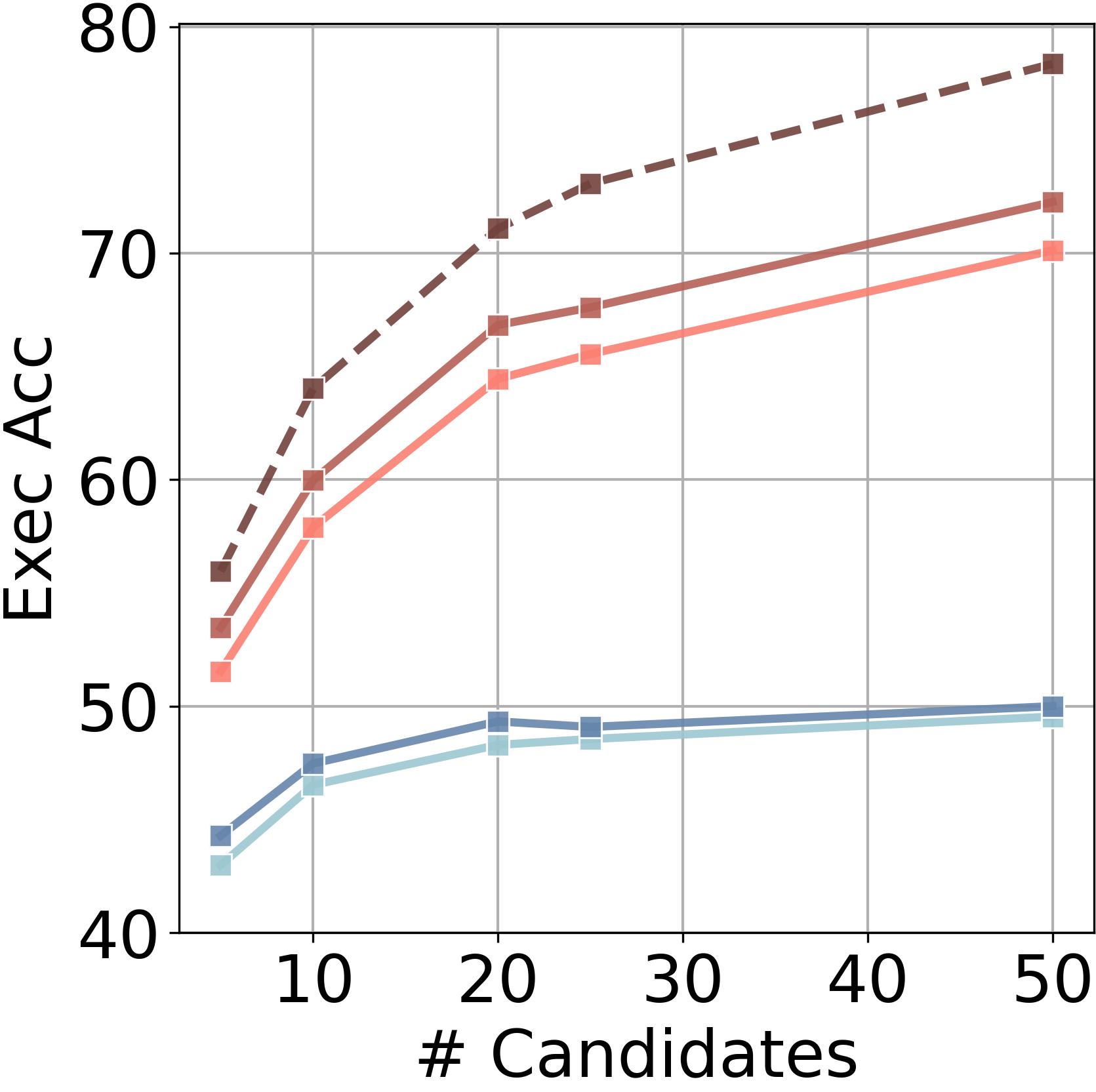}
        \caption{HumanEval+}
        \label{fig:sd_candidates_humaneval_plus}
    \end{subfigure}\hfill
    \begin{subfigure}{0.154\textwidth}
        \centering
        \includegraphics[width=\linewidth]{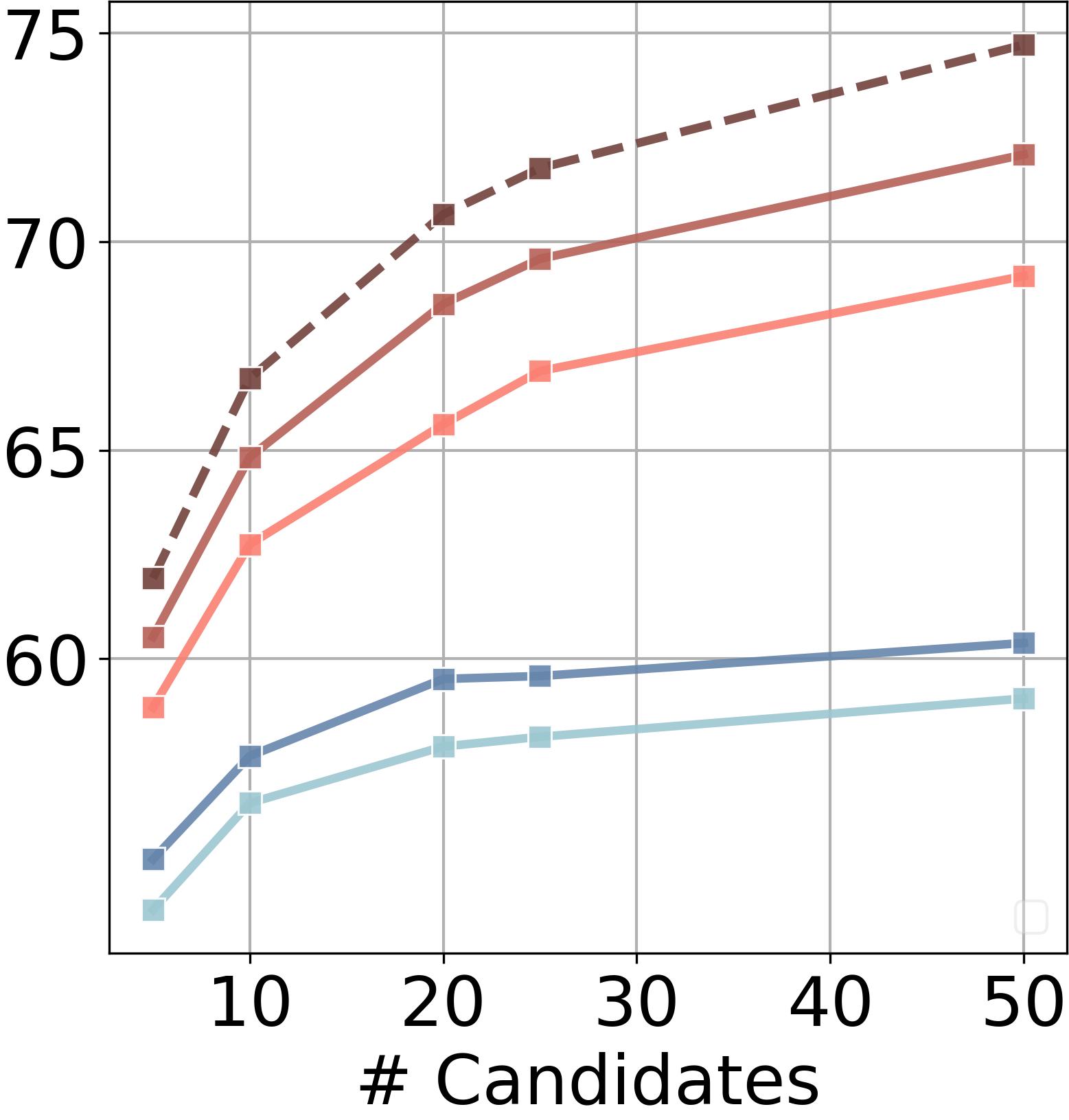}
        \caption{MBPP-S+}
        \label{fig:sd_candidates_mbpp_plus}
    \end{subfigure}\hfill
    \begin{subfigure}{0.154\textwidth}
        \centering
        \includegraphics[width=\linewidth]{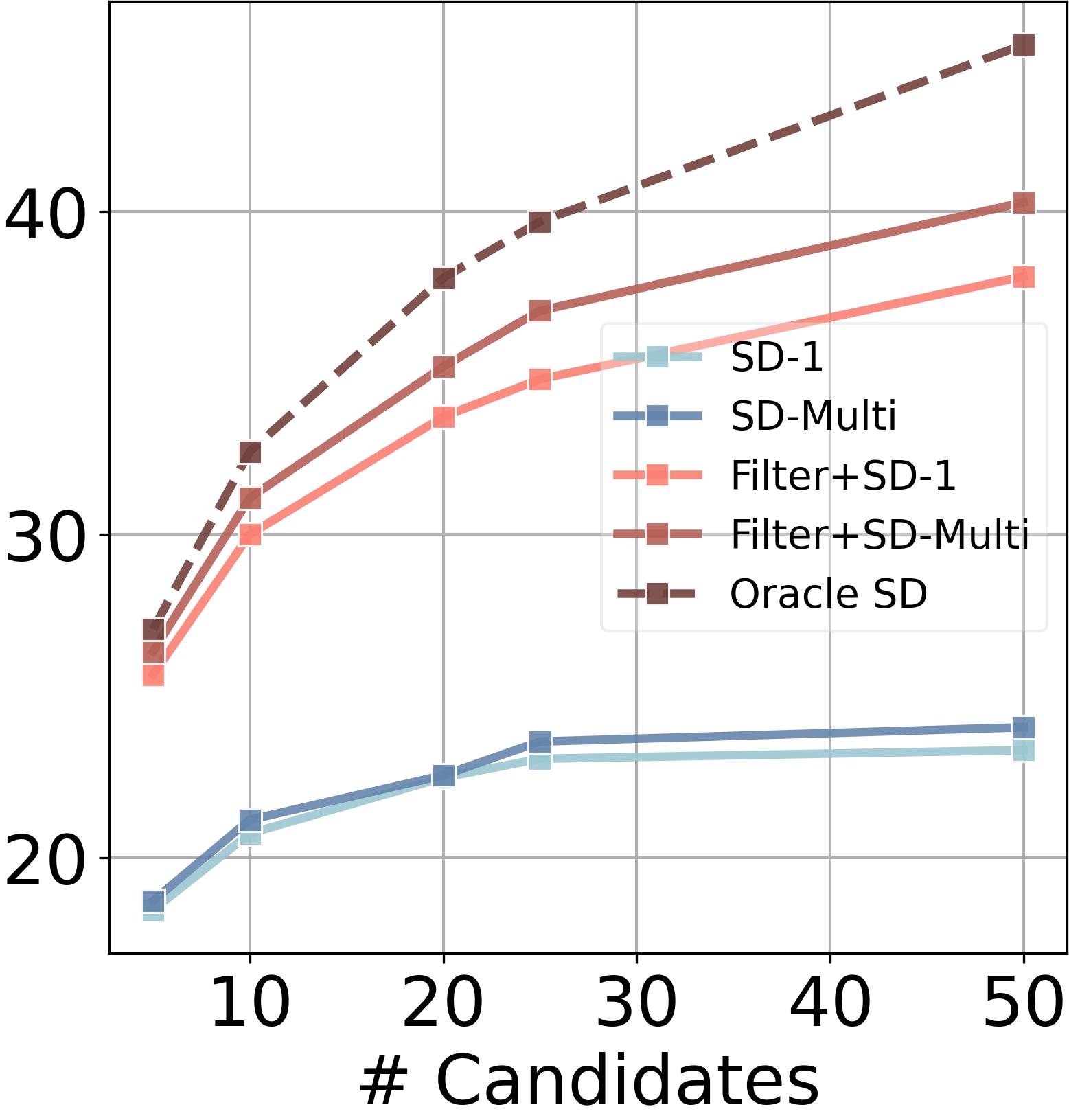}
        \caption{LiveCodeBench}
        \label{fig:sd_candidates_lcb}
    \end{subfigure}
    \caption{Comparison of Self-Debug methods over different numbers of candidates generated by CodeLlama-7B-Instruct, and debugged over $\{1,\textbf{Multi}\}$ candidates. Results are averaged across at least 4 runs (and 2 runs for LiveCodeBench).}
    \label{fig:sd_candidates}
\end{figure}

\paragraph{Filtering matters in Self-Debug.} With 50 candidates generated and self-debugged, more experiments of MBR-Exec outperforming SD-1 with filtering compared to without filtering (see Table~\ref{tab:sd_results}), suggesting that improvement brought by Self-Debug can be partly accomplished by filtering, a finding that is missing from \citet{chen2024teaching} where filtering is also applied but not analyzed. Additionally, when filtering is applied, SD-1 only outperforms MBR-Exec with samples generated and debugged with CodeLlama-13B-Instruct.

Figure~\ref{fig:sd_candidates} shows that SD-Multi (1 round) consistently outperforms SD-1 (3 rounds) with different numbers of candidates generated. 
Last but not least, Self-Debug leaves a larger gap between SD-Multi and the oracle.

\section{Related Work}

\paragraph{Large Language Models for Programming Languages.} Recent years have witnessed the success of pretrained large language models for code generation \citep{codex, alphacode,li2023starcoder,codesurvey,zhuo-2024-ice} that have been successful on multiple execution-based benchmarks \citep{apps,mbpp,codex,evalplus,ds1000,odex}. More recently, strong-performing open-source pretrained Code LLMs \citep{rozière2024code,luo2024wizardcoder,guo2024deepseek,lozhkov2024starcoder,zhu2024deepseek} have been available, and they can perform tasks such as filling-in-the-middle \citep{rozière2024code}, code debugging \citep{guo2024codeeditorbench}. Our work uses representative open-source code LLMs to study generation, reranking, and Self-Debug.

\paragraph{Reranking in Code Generation} \citet{shi-etal-2022-natural} and \citet{alphacode} proposed MBR decoding based on agreement on execution results. \citet{alphacode} further proposed filtering based on trial unit tests, and \citet{shi-etal-2022-natural} proposed filtering out samples that cannot get execution outputs, followed by \citet{coder-reviewer} combining log-likelihoods of the code candidate and the instruction. \citet{reranking_laws} further studied the reranking laws based on communication theory. Our work systematically studies proposed methods.

Other works adopt a setting without unit test inputs for evaluation available, and proposed methods with generated unit test cases utilizing generated unit tests \citep{alphacode,chen2023codet,to2024neural} for reranking and \citet{huang-etal-2024-enhancing} incorporating both generated unit test cases and functional specifications Our work differs from these efforts as we adopt a setting with unit test inputs for evaluation available. 

Additionally, previous works have also studied training N-Best rerankers. Early efforts include training classifiers that predict program success in execution \citep{inala2022faultaware}, followed by \citet{ni2023lever} proposing to train both model-specific and task-specific classifiers for reranking. Our work differs from \citet{ni2023lever} as we evaluate with more general-purpose metrics \citep{dong2023codescore} that do not rely on execution during inference time.

Generally, our work differs as we systematically study the reranking with representative methods, along with sampling strategies and Self-Debug.

\paragraph{Self-Debug with LLMs}

\citet{chen2024teaching} studied the effect of using the same Code LLM to debug a single generated code candidate by varying feedback information such as execution errors and code explanation. \citet{selfpair} also studied Self-Debug but focus on different sampling methods and their impacts on Self-Debug. \citet{zhong2024ldb} additionally integrated information regarding intermediate states to provide additional information for debugging. Our work differs from theirs as we study the joint effect of both Self-Debug and reranking. We also perform Self-Debug on multiple candidates.

\section{Conclusion and Future Work}

We propose \textbf{DOCE}, a unified inference framework for code generation. We systematically study candidate generation, reranking, and self-debugging. Importantly, we highlight the impact of high-temperature sampling, execution-based reranking with few high-quality unit tests, and self-debugging with multiple candidates. Potential future works include analyzing unit test quality for reranking and proposing better methods for generating code candidates, exploring the current framework with generated unit test cases, and exploring general-purpose rerankers.

\section*{Limitation}
First of all, our paper is limited by the scale of experiments as we cannot experiment with all available large language models due to the limit of computation, and our solution is to select representative classes of open-source models and experiments on models with parameters at most 16 billions. Additionally, this paper is limited by potential risks of data contamination from the dataset, despite that HumanEval and MBPP are commonly used in evaluating code generation models. Our solution is to include experiments with LiveCodeBench, which contains a lower risk of data contamination.

\section*{Ethical Considerations}
We do not consider the existence of ethical issues related to the paper, due to the nature of code generation and our usage of publicly available datasets that have been verified. However, we noticed the risk of ethical concern due to our choice of sampling temperature. We checked generations and found no ethical issues in the generated content.


\bibliography{custom}
\clearpage

\newpage

\appendix

\clearpage

\begin{figure*}[tbp]
    \centering
    \begin{subfigure}{0.237\textwidth}
        \centering
        \includegraphics[width=\linewidth]{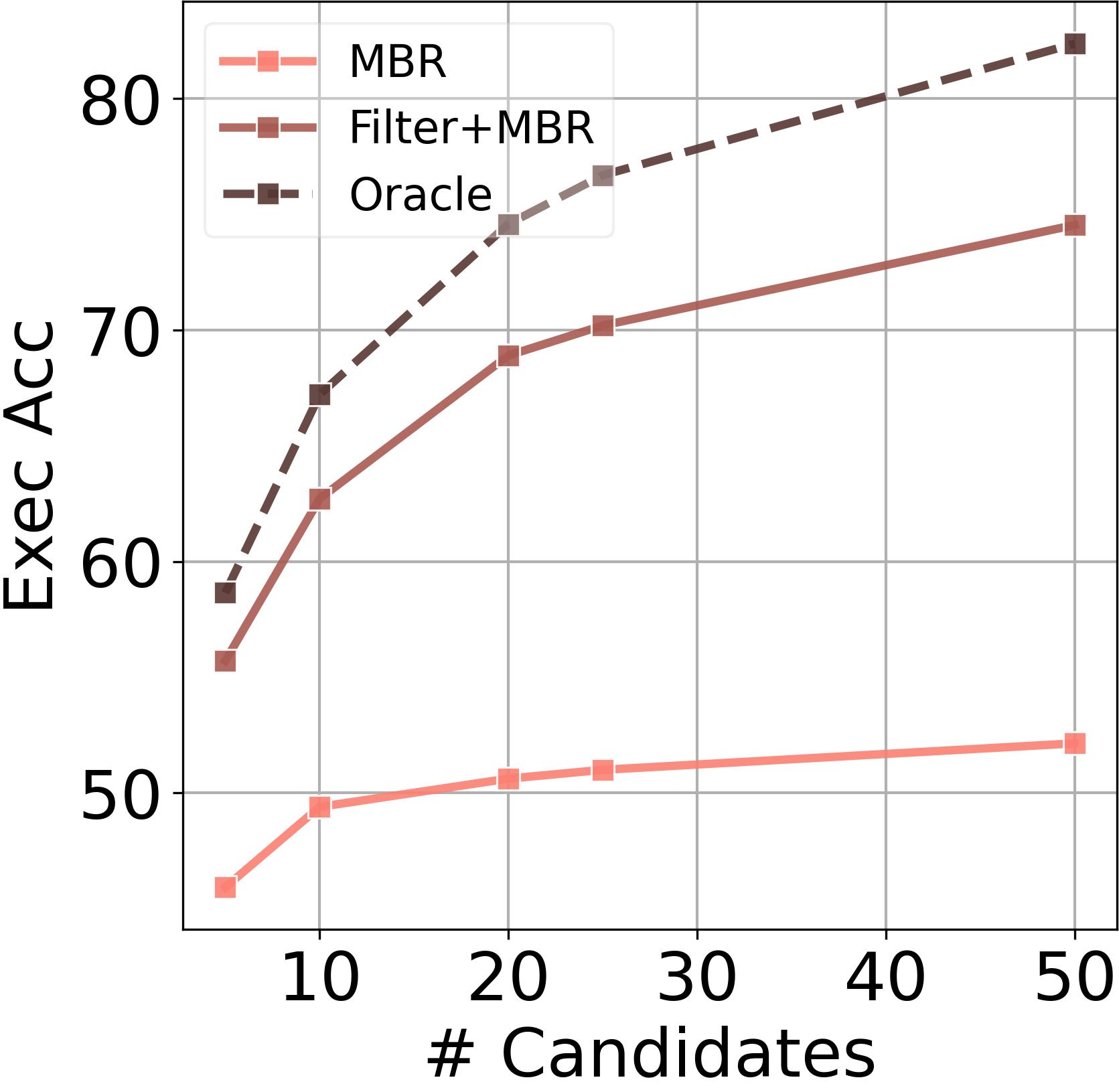}
        \caption{HumanEval}
        \label{fig:cl_7b_num_candidates_humaneval_base}
    \end{subfigure}\hfill
    \begin{subfigure}{0.22\textwidth}
        \centering
        \includegraphics[width=\linewidth]{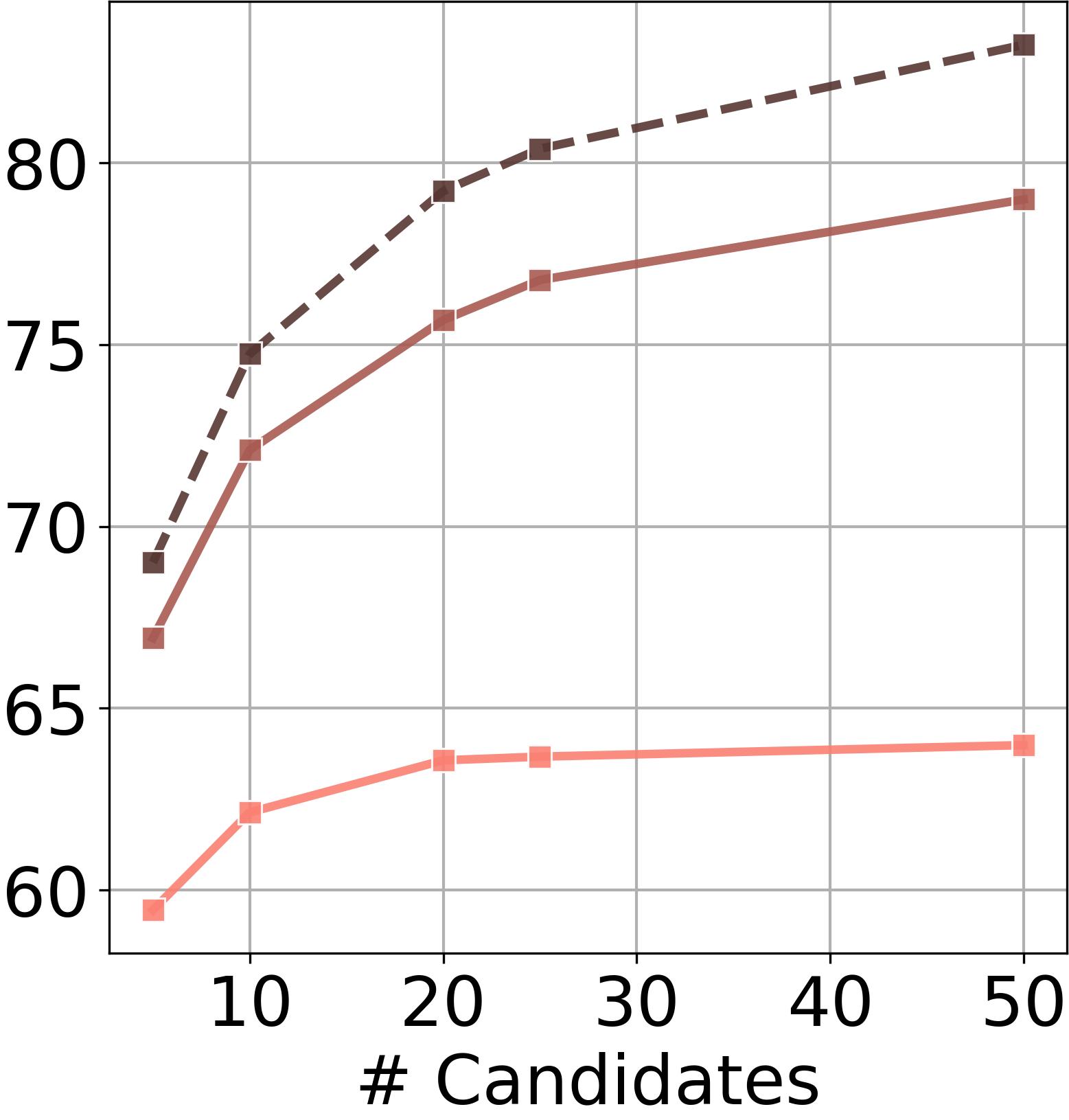}
        \caption{MBPP-S}
        \label{fig:cl_7b_num_candidates_mbpp_base}
    \end{subfigure}\hfill
    \begin{subfigure}{0.237\textwidth}
        \centering
        \includegraphics[width=\linewidth]{figures/humaneval_candidates.jpg}
        \caption{HumanEval+}
        \label{fig:cl_7b_num_candidates_humaneval_plus}
    \end{subfigure}\hfill
    \begin{subfigure}{0.22\textwidth}
        \centering
        \includegraphics[width=\linewidth]{figures/mbpp_candidates.jpg}
        \caption{MBPP-S+}
        \label{fig:cl_7b_num_candidates_mbpp_plus}
    \end{subfigure}

    \caption{Performance of MBR-Exec with and without filtering over different numbers of generated candidates using CodeLlama-7B-Instruct. We also provide the oracle with solid lines, denoted as \textit{Pass@k}. Results that end with $+$ mean that it's evaluated on the plus with extended test cases. Results are averaged across at least 4 runs.
    }
    \label{fig:cl_7b_num_candidates}
\end{figure*}

\begin{figure*}[htbp]
    \centering
    \begin{subfigure}{0.237\textwidth}
        \centering
        \includegraphics[width=\linewidth]{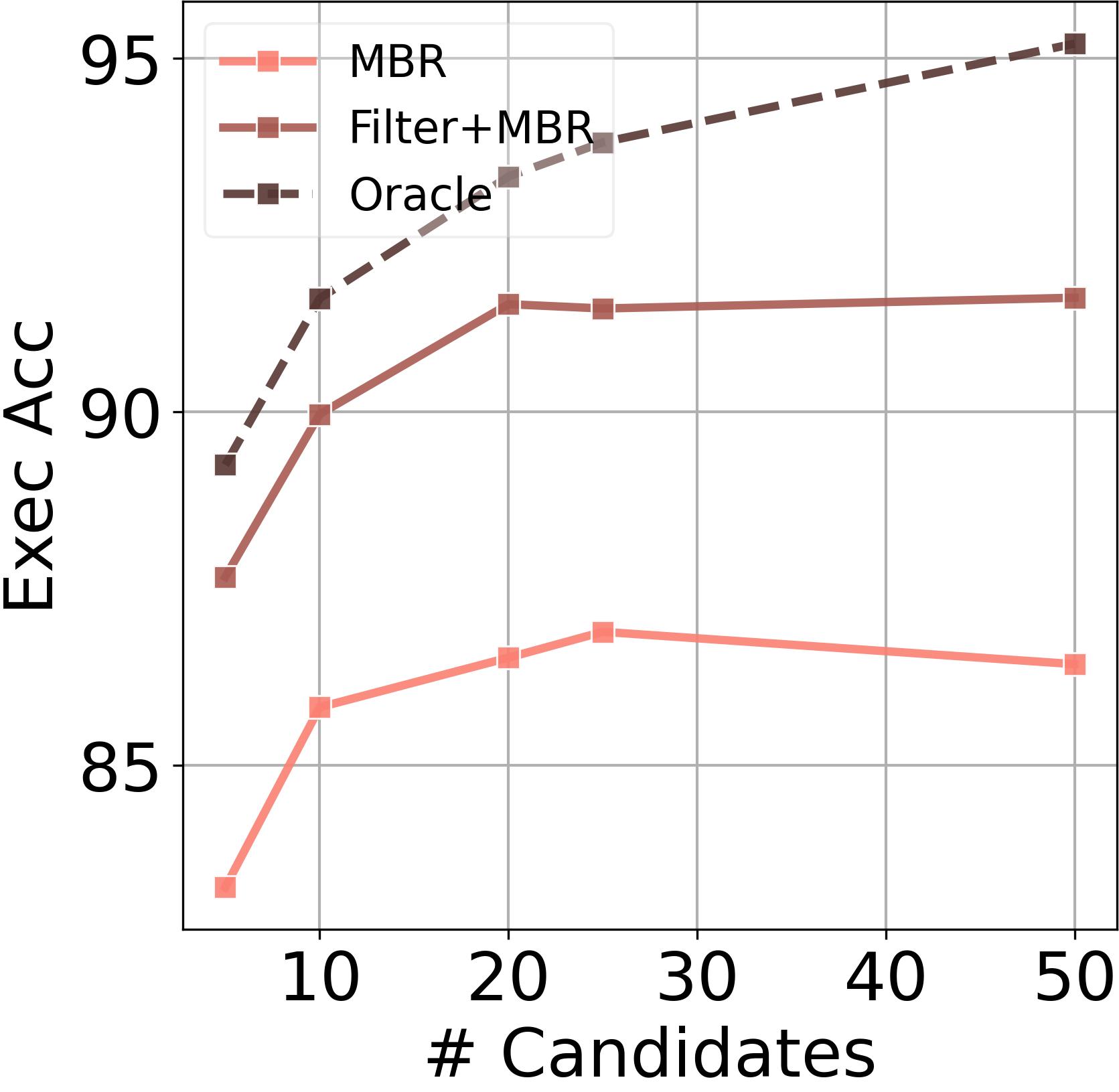}
        \caption{HumanEval}
        \label{fig:ds_6.7b_candidates_humaneval_base}
    \end{subfigure}\hfill
    \begin{subfigure}{0.22\textwidth}
        \centering
        \includegraphics[width=\linewidth]{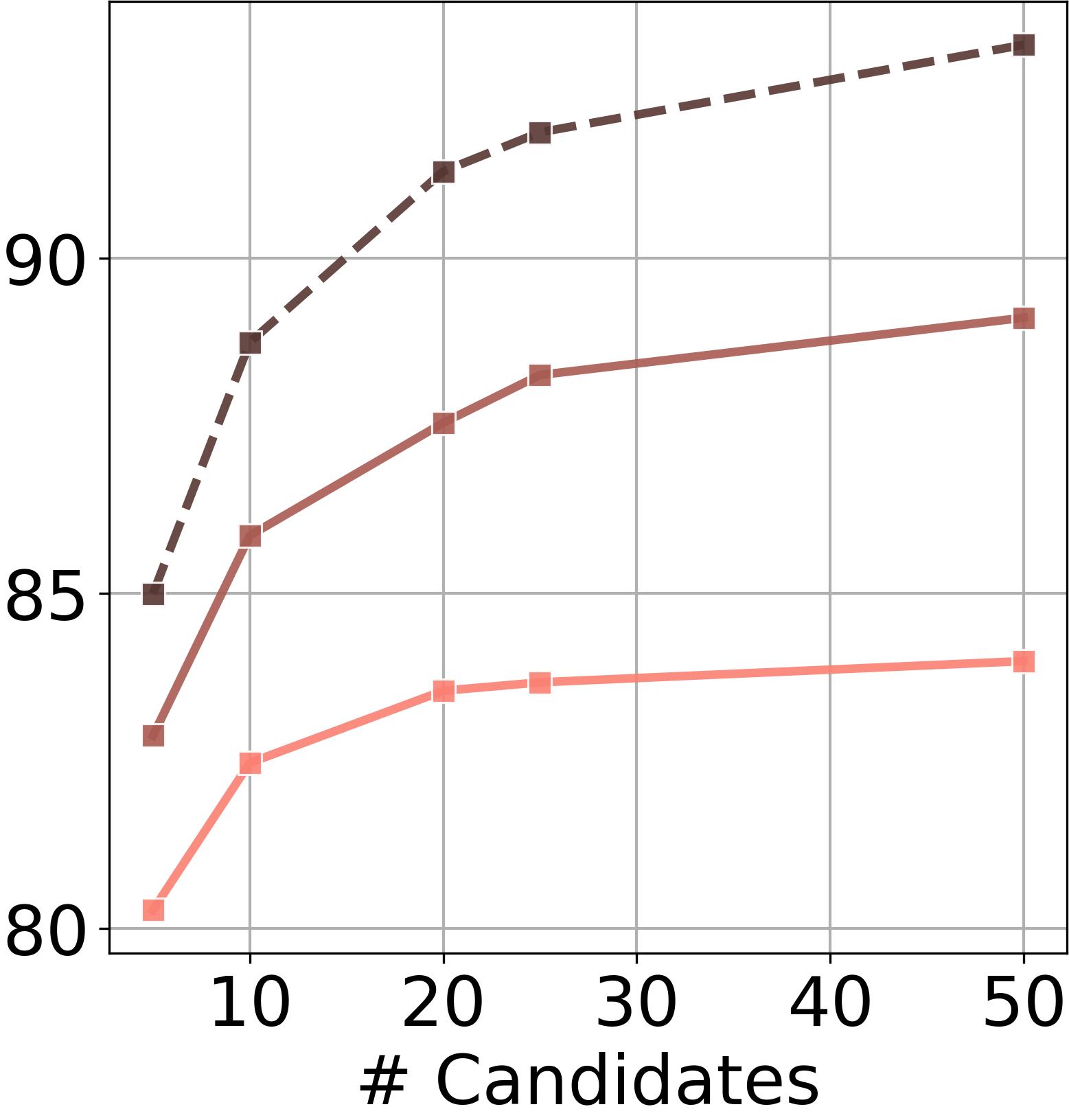}
        \caption{MBPP-S}
        \label{fig:ds_6.7b_num_candidates_mbpp_base}
    \end{subfigure}\hfill
    \begin{subfigure}{0.237\textwidth}
        \centering
        \includegraphics[width=\linewidth]{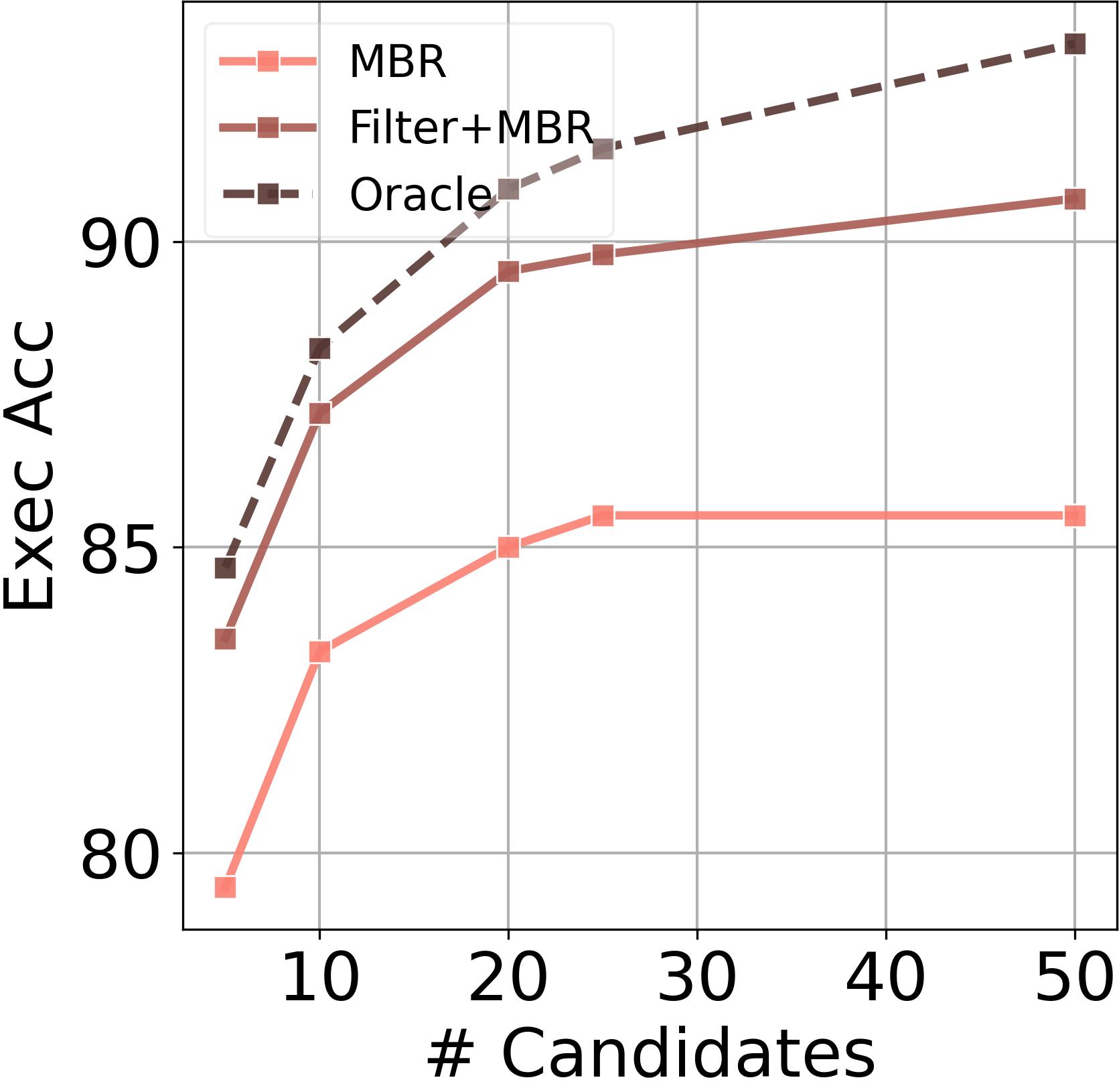}
        \caption{HumanEval+}
        \label{fig:ds_6.7b_candidates_humaneval_plus}
    \end{subfigure}\hfill
    \begin{subfigure}{0.22\textwidth}
        \centering
        \includegraphics[width=\linewidth]{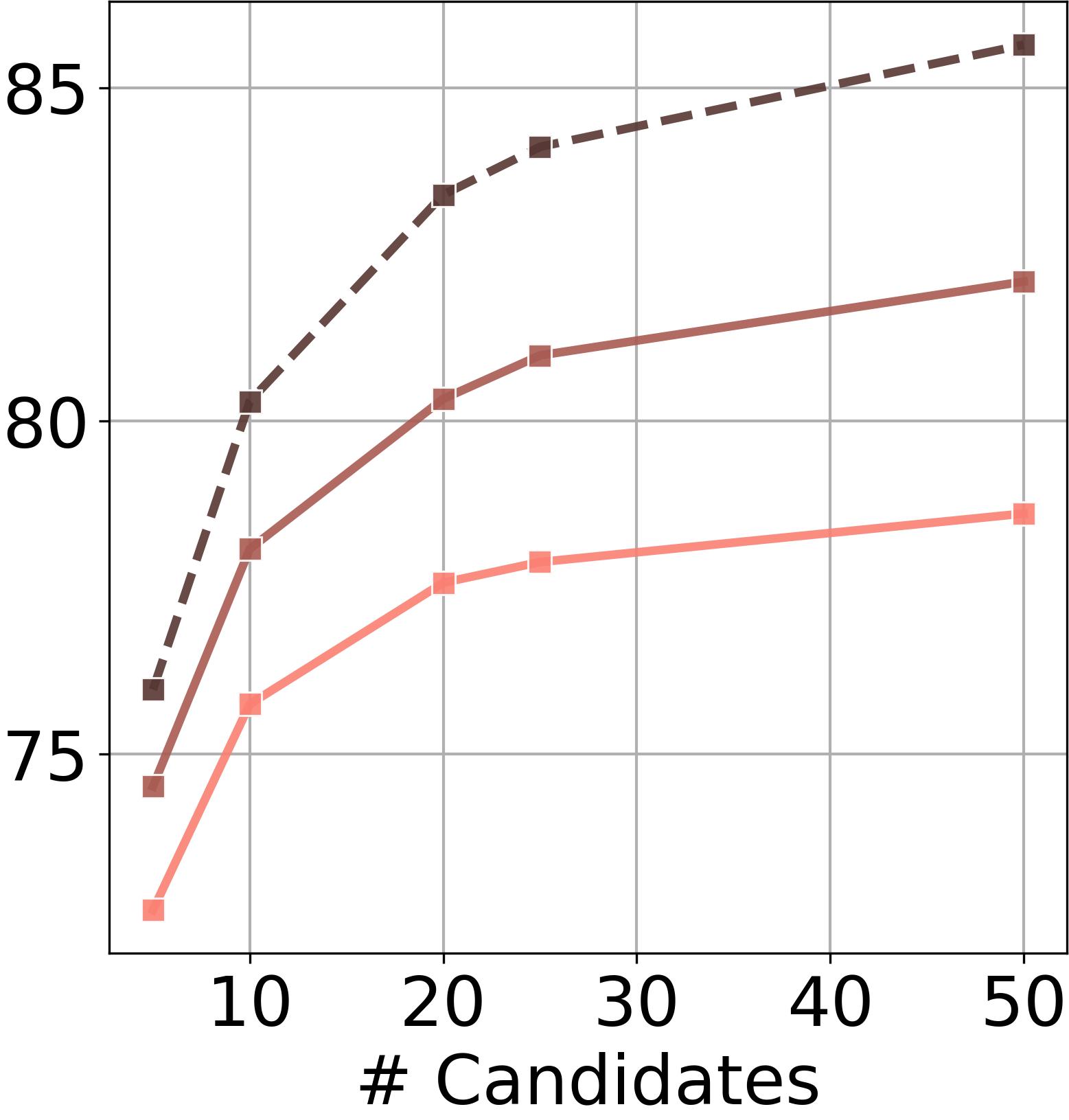}
        \caption{MBPP-S+}
        \label{ds_6.7b_candidates_humaneval_plus}
    \end{subfigure}

    \caption{Performance of MBR-Exec with and without filtering over different numbers of generated candidates using DeepSeekCoder-6.7B-Instruct. We also provide the oracle with solid lines, denoted as \textit{Pass@k}. Results that end with $+$ mean that it's evaluated on the plus with extended test cases. Results are averaged across at least 4 runs.
    }
    \label{fig:ds_6.7b_num_candidates}
\end{figure*}

\begin{figure*}[tbp]
    \centering
    \begin{subfigure}{0.237\textwidth}
        \centering
        \includegraphics[width=\linewidth]{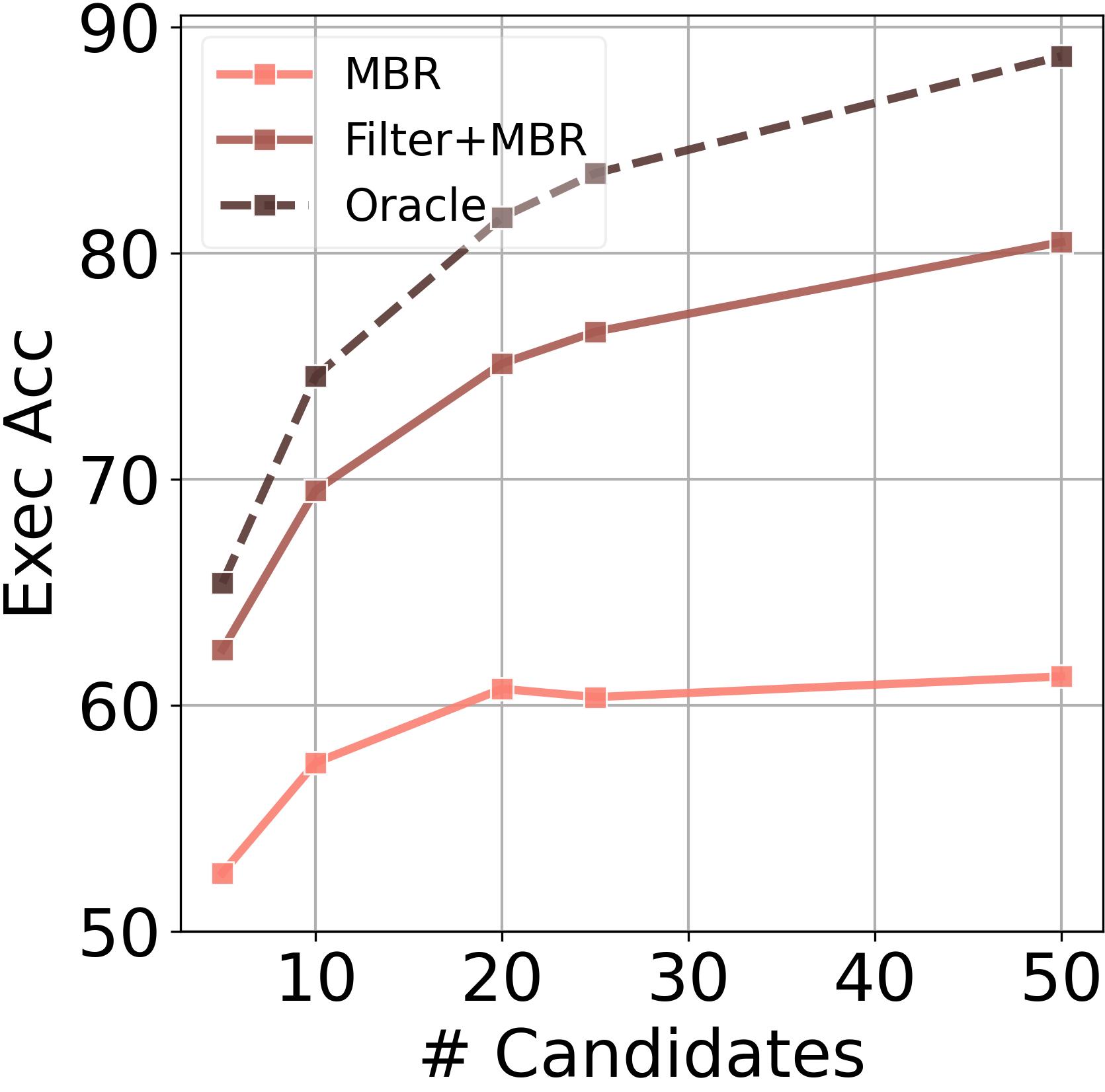}
        \caption{HumanEval}
        \label{fig:cl_13b_num_candidates_humaneval_base}
    \end{subfigure}\hfill
    \begin{subfigure}{0.22\textwidth}
        \centering
        \includegraphics[width=\linewidth]{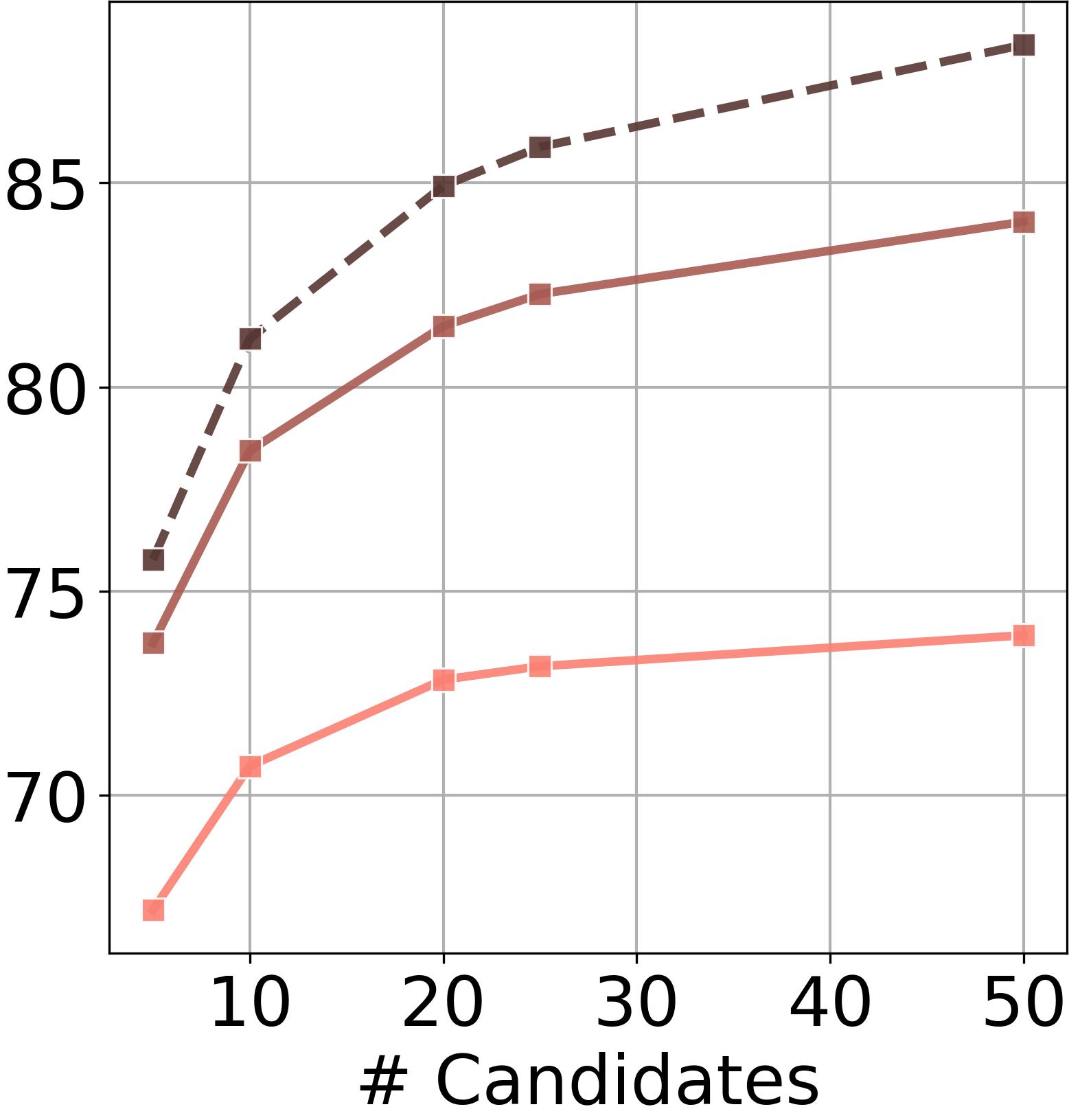}
        \caption{MBPP-S}
        \label{fig:cl_13b_num_candidates_mbpp_base}
    \end{subfigure}\hfill
    \begin{subfigure}{0.237\textwidth}
        \centering
        \includegraphics[width=\linewidth]{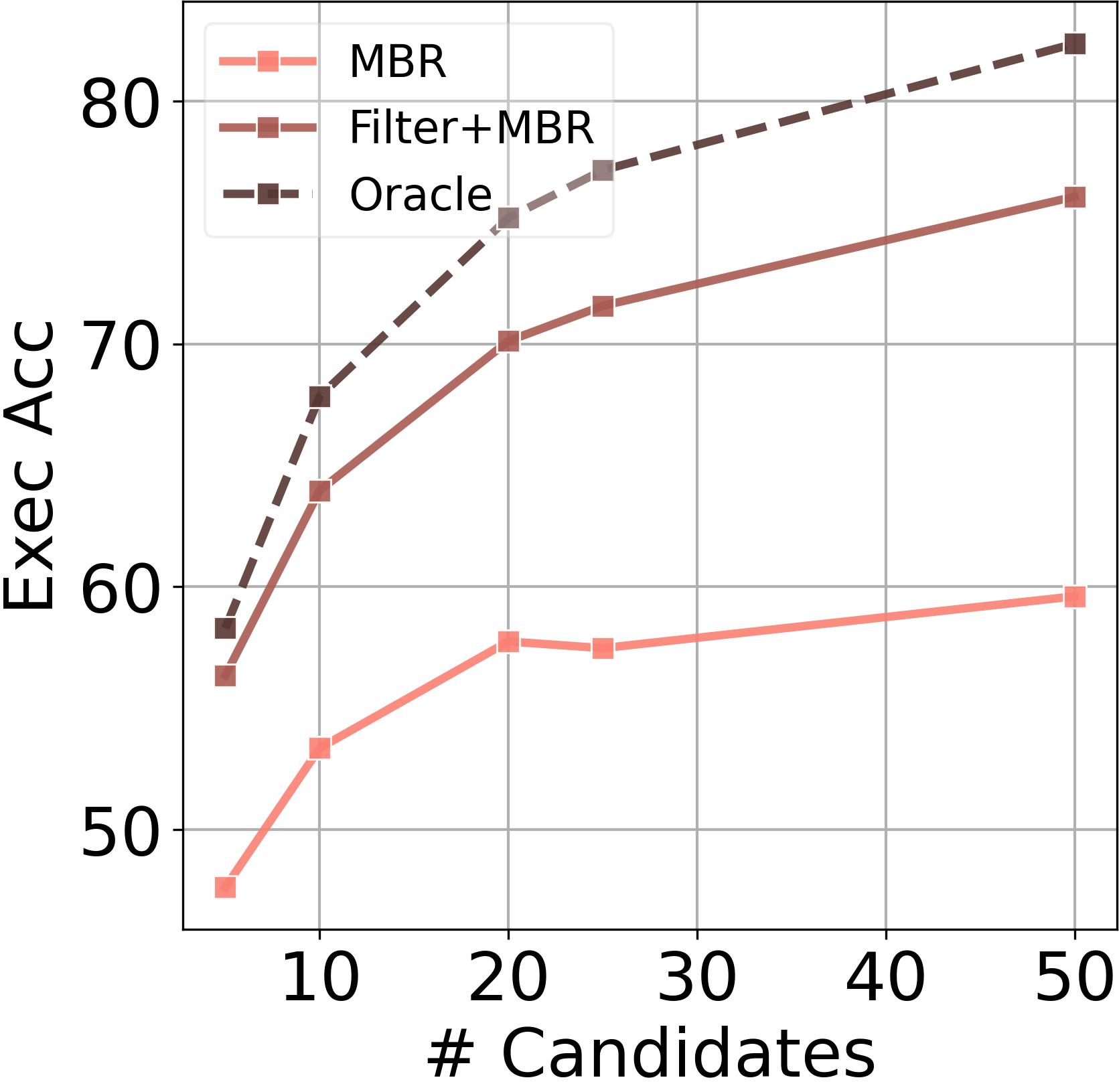}
        \caption{HumanEval+}
        \label{fig:cl_13b_num_candidates_humaneval_plus}
    \end{subfigure}\hfill
    \begin{subfigure}{0.22\textwidth}
        \centering
        \includegraphics[width=\linewidth]{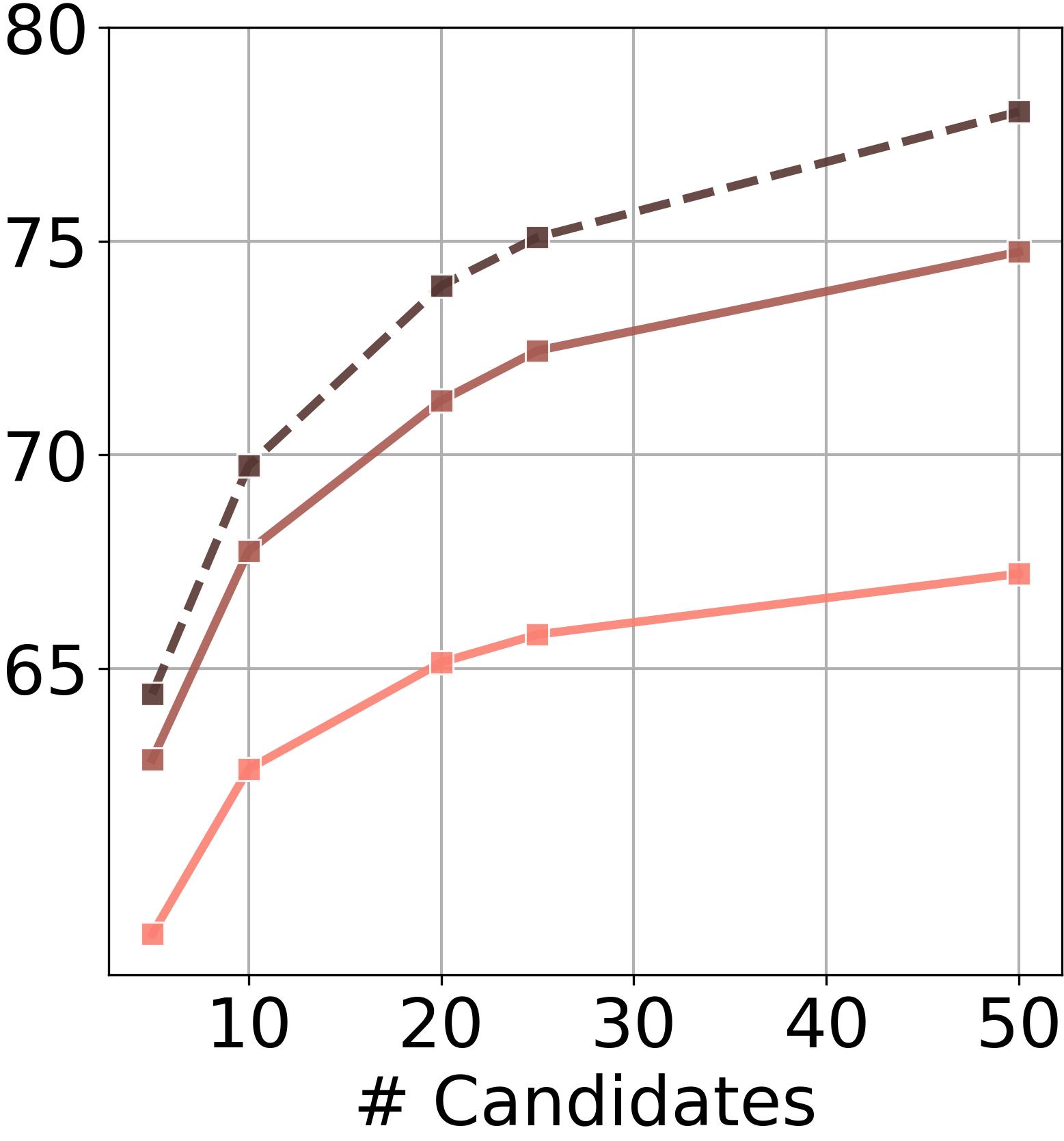}
        \caption{MBPP-S+}
        \label{fig:cl_13b_num_candidates_mbpp_plus}
    \end{subfigure}

    \caption{Performance of MBR-Exec with and without filtering over different numbers of generated candidates using CodeLlama-13B-Instruct. We also provide the oracle with solid lines, denoted as \textit{Pass@k}. Results that end with $+$ mean that it's evaluated on the plus with extended test cases. Results are averaged across at least 4 runs.
    }
    \label{fig:cl_13b_num_candidates}
\end{figure*}

\begin{figure*}[htbp]
    \centering
    \begin{subfigure}{0.237\textwidth}
        \centering
        \includegraphics[width=\linewidth]{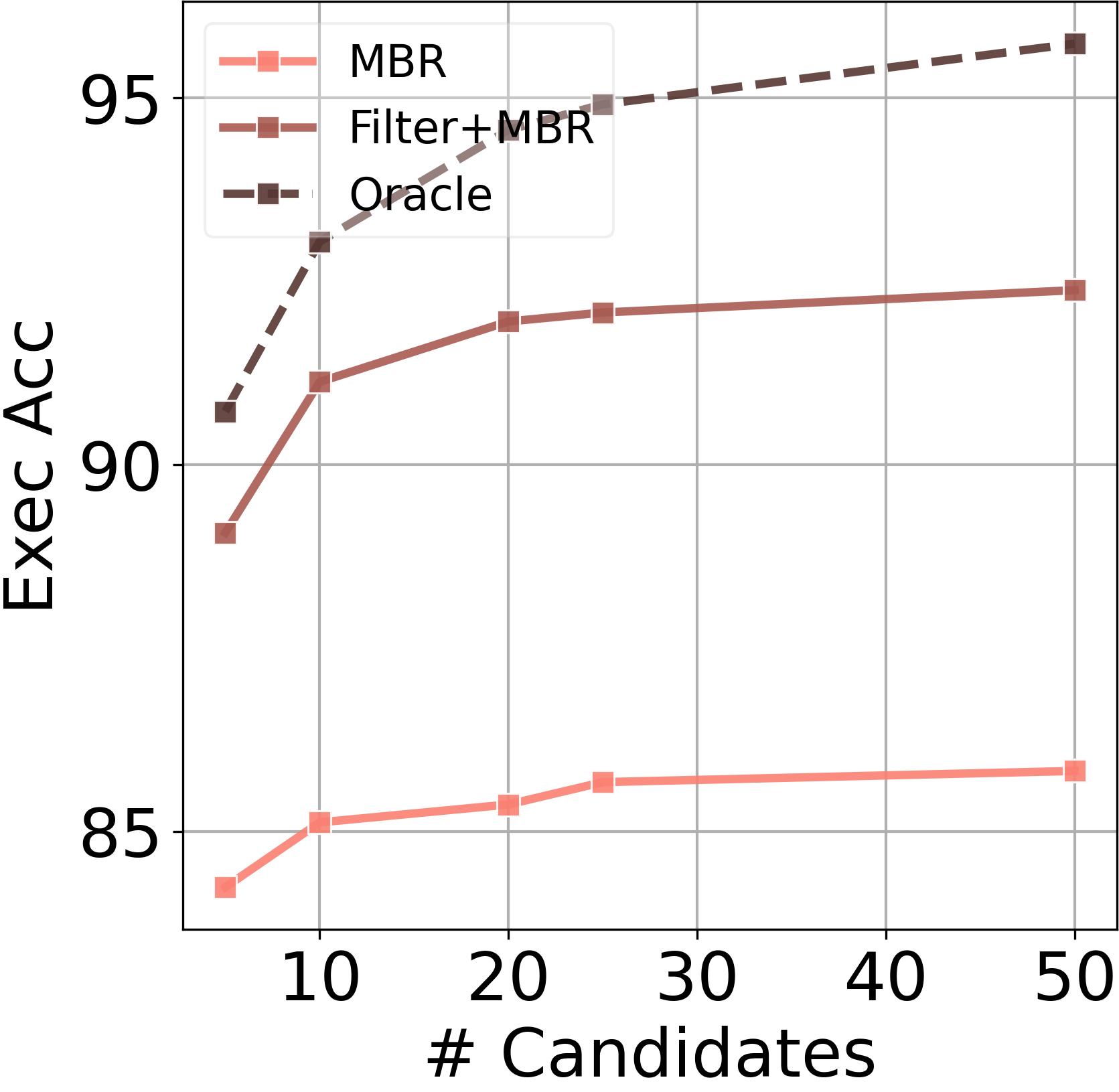}
        \caption{HumanEval}
        \label{fig:ds_16b_candidates_humaneval_base}
    \end{subfigure}\hfill
    \begin{subfigure}{0.22\textwidth}
        \centering
        \includegraphics[width=\linewidth]{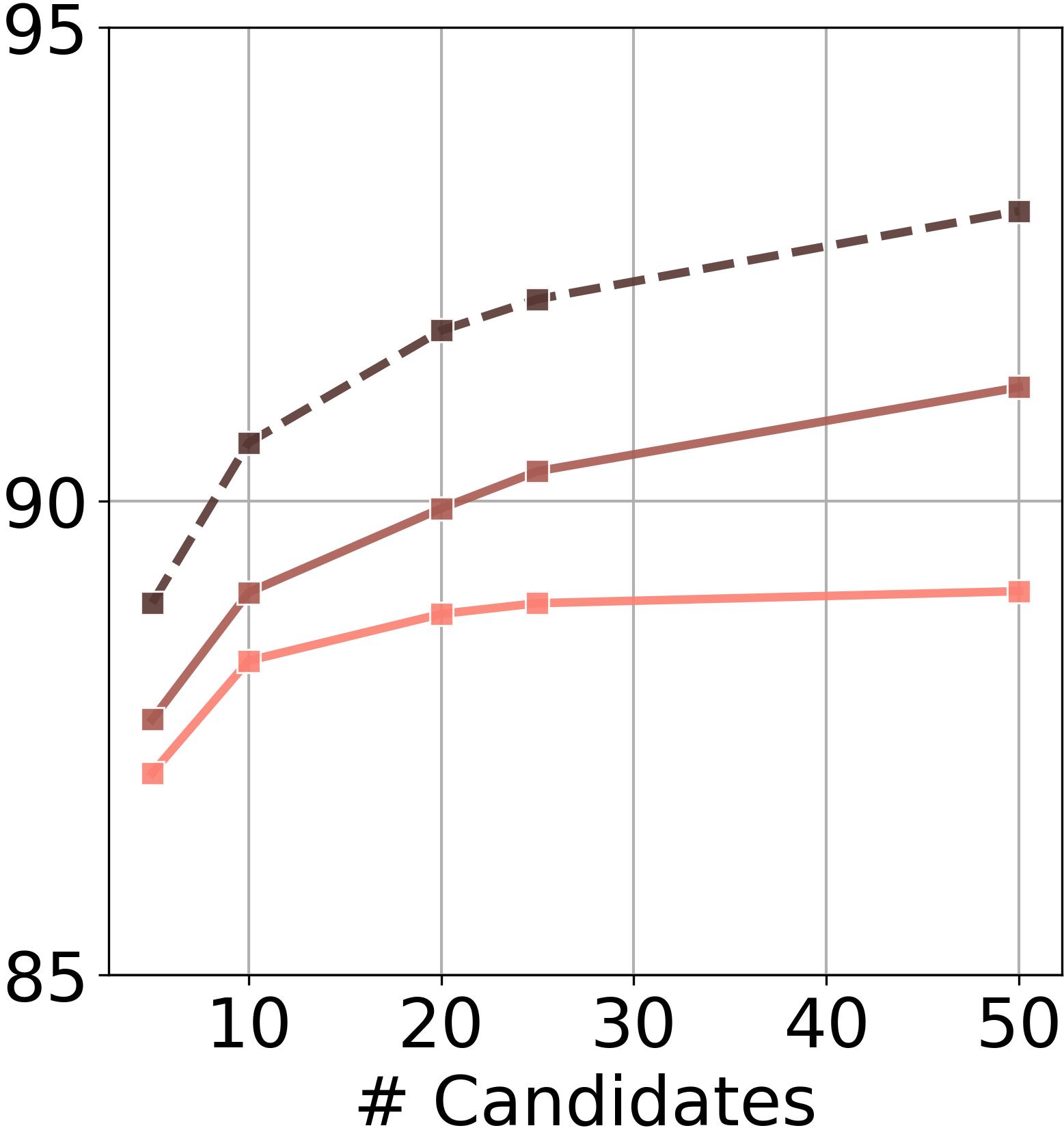}
        \caption{MBPP-S}
        \label{fig:ds_16b_num_candidates_mbpp_base}
    \end{subfigure}\hfill
    \begin{subfigure}{0.237\textwidth}
        \centering
        \includegraphics[width=\linewidth]{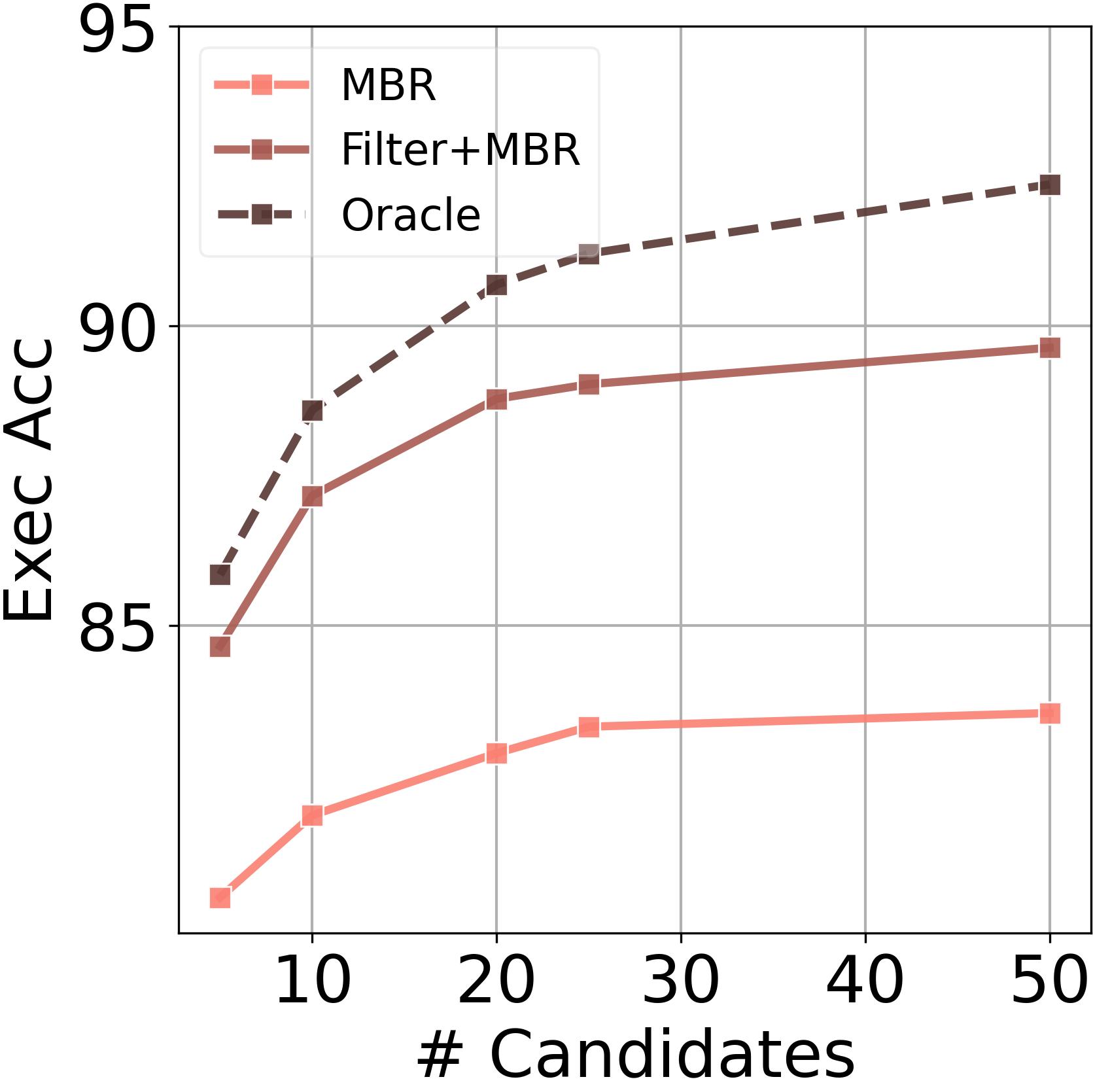}
        \caption{HumanEval+}
        \label{fig:ds_16b_candidates_humaneval_plus}
    \end{subfigure}\hfill
    \begin{subfigure}{0.22\textwidth}
        \centering
        \includegraphics[width=\linewidth]{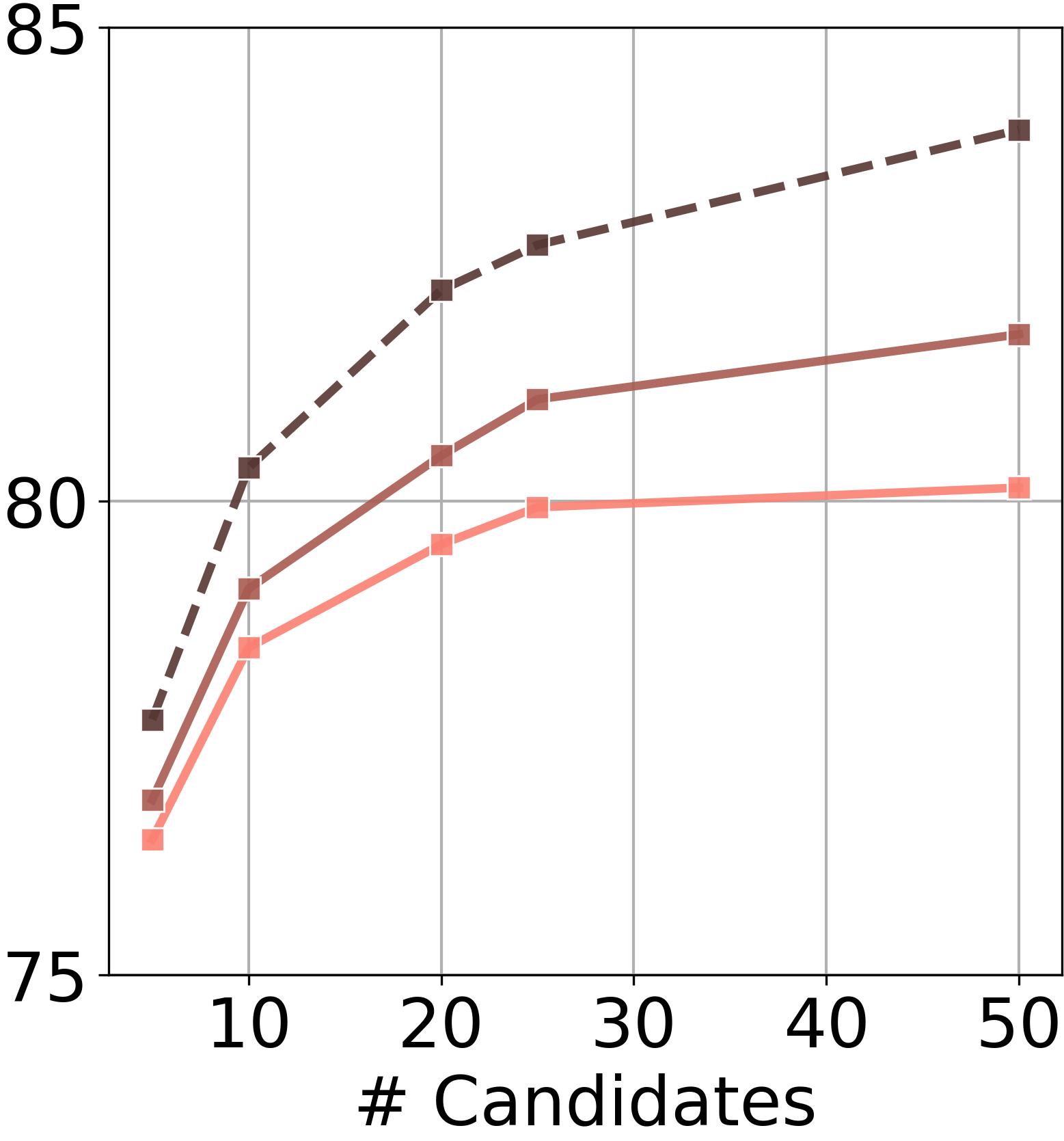}
        \caption{MBPP-S+}
        \label{ds_16b_candidates_humaneval_plus}
    \end{subfigure}

    \caption{Performance of MBR-Exec with and without filtering over different numbers of generated candidates using DeepSeekCoder-V2-Lite-Instruct. We also provide the oracle with solid lines, denoted as \textit{Pass@k}. Results that end with $+$ mean that it's evaluated on the plus with extended test cases. Results are averaged across at least 4 runs.
    }
    \label{fig:ds_16b_num_candidates}
\end{figure*}

\begin{figure*}[htbp]
    \centering
    \begin{subfigure}{0.237\textwidth}
        \centering
        \includegraphics[width=\linewidth]{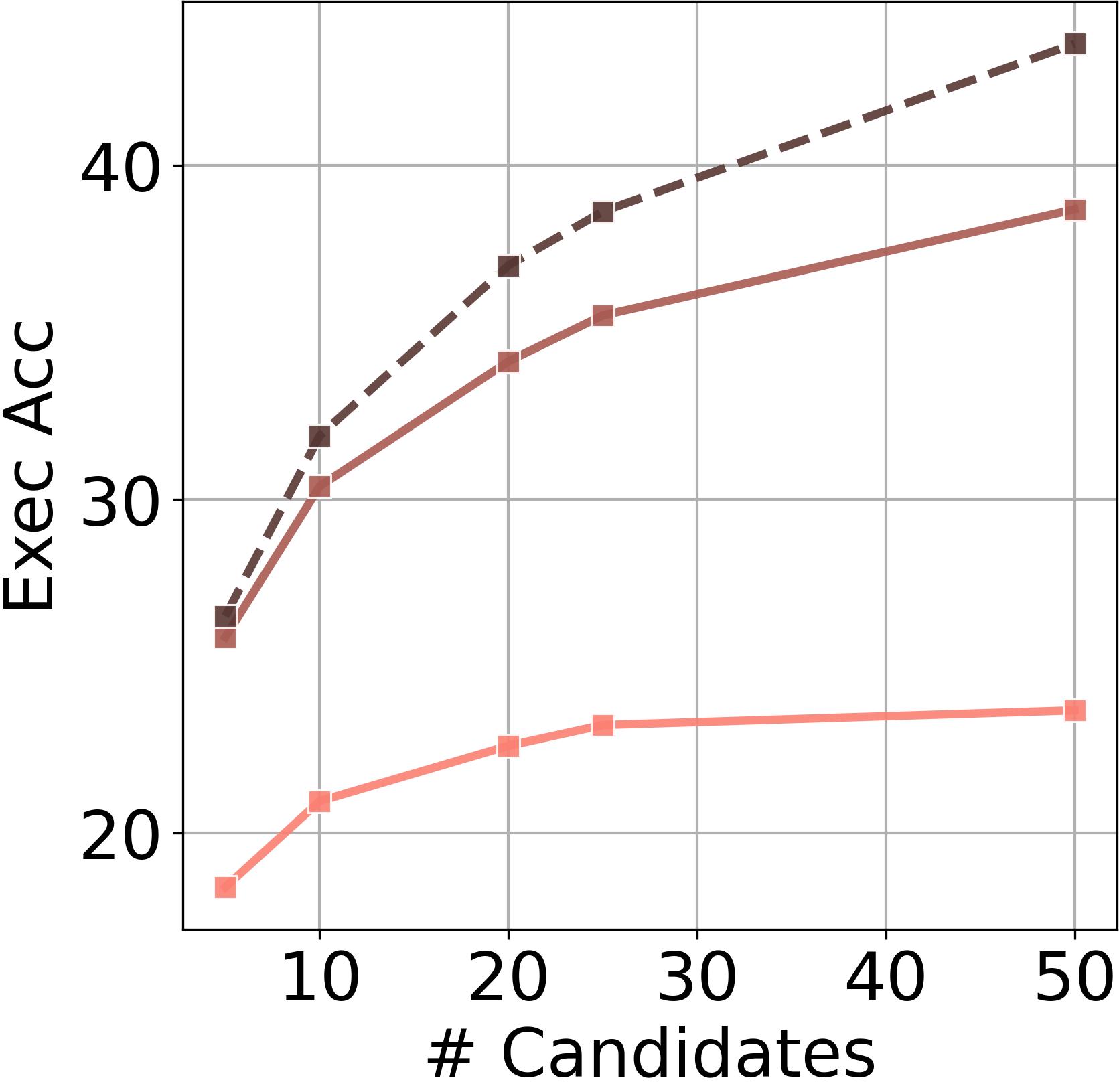}
        \caption{CL-7B}
        \label{fig:lcb_candidates_original}
    \end{subfigure}\hfill
    \begin{subfigure}{0.22\textwidth}
        \centering
        \includegraphics[width=\linewidth]{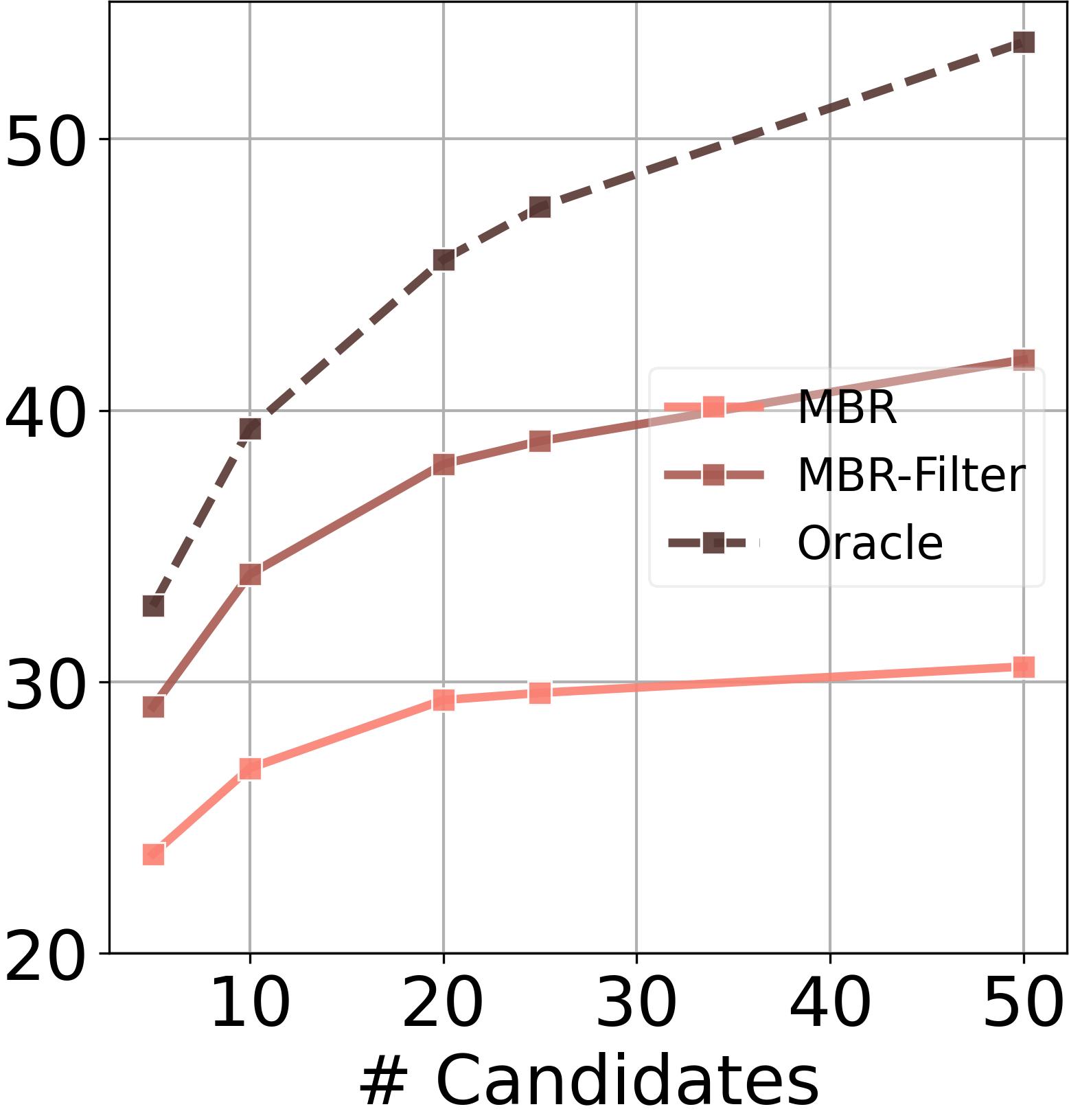}
        \caption{DS-6.7B}
        \label{fig:ds_6.7b_num_candidates_lcb}
    \end{subfigure}\hfill
    \begin{subfigure}{0.237\textwidth}
        \centering
        \includegraphics[width=\linewidth]{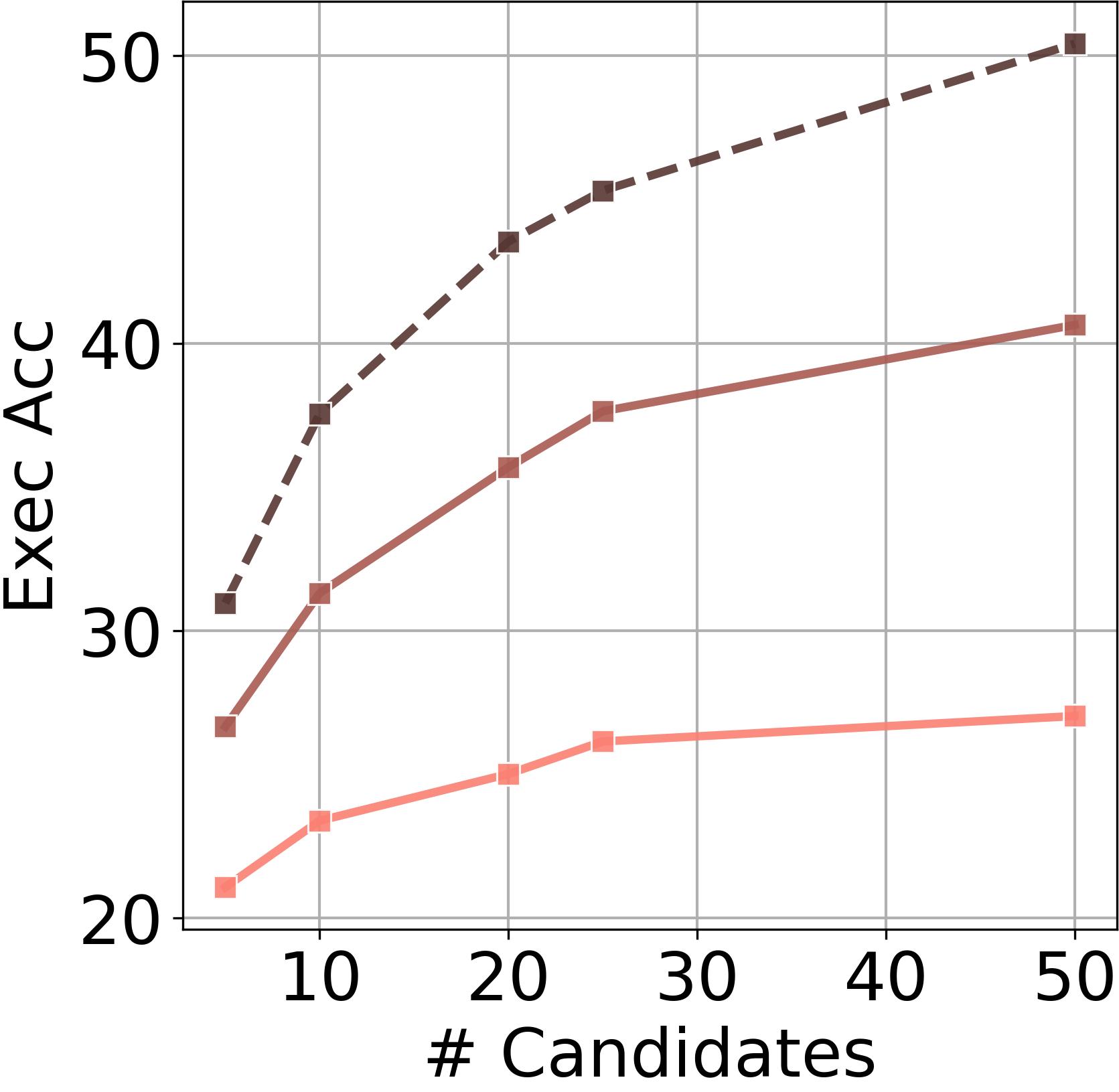}
        \caption{CL-13B}
        \label{fig:cl_13b_num_candidates_lcb}
    \end{subfigure}\hfill
    \begin{subfigure}{0.22\textwidth}
        \centering
        \includegraphics[width=\linewidth]{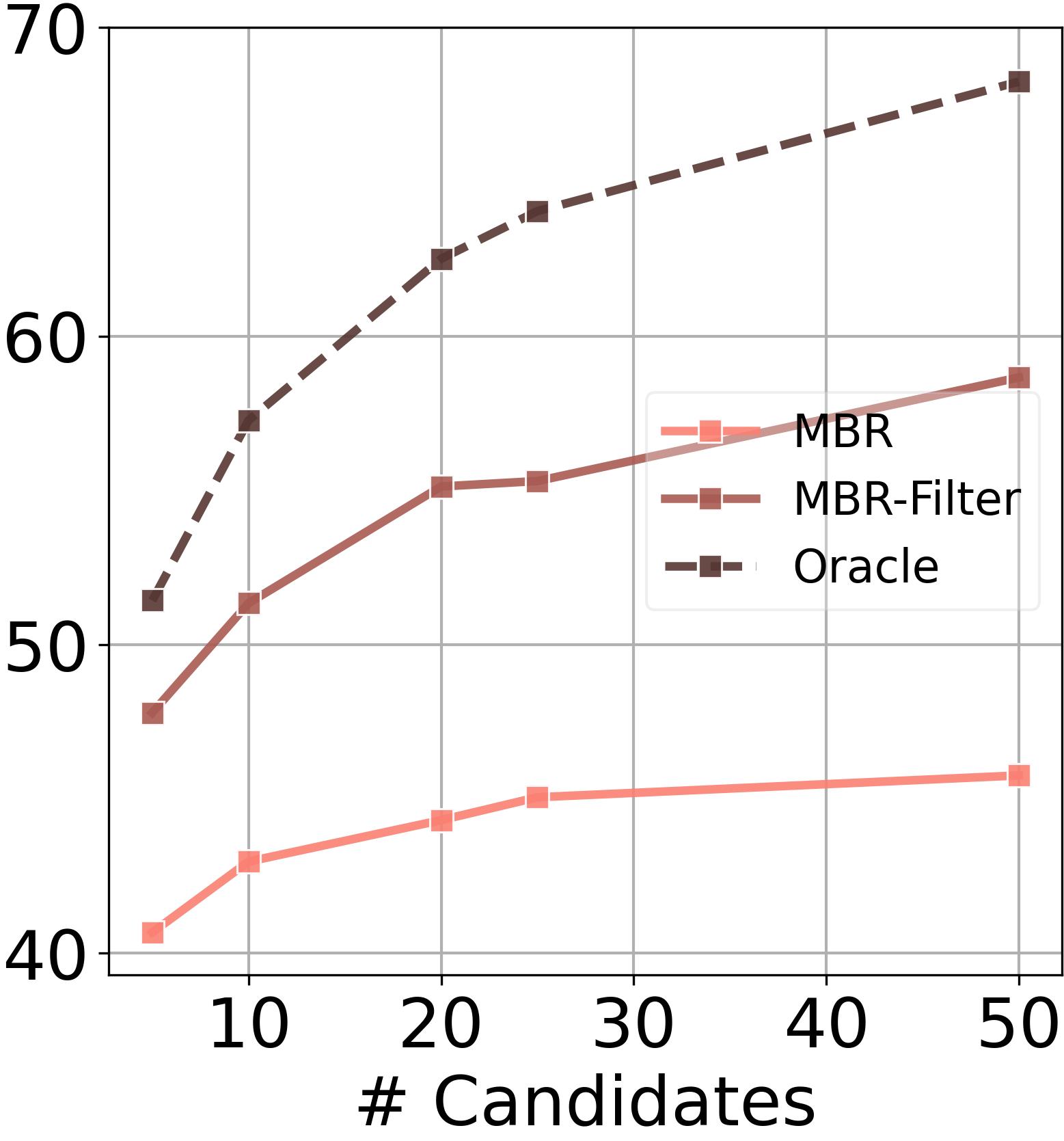}
        \caption{DS-V2-Lite}
        \label{fig:ds_16b_num_candidates_lcb}
    \end{subfigure}

    \caption{Performance of MBR-Exec with and without filtering over different numbers of generated candidates using all models and tested on LiveCodeBench. We also provide the oracle with solid lines, denoted as \textit{Pass@k}. Results are averaged across at least 2 runs.
    }
    \label{fig:lcb_num_candidates}
\end{figure*}

\begin{figure*}[tbp]
    \centering
    \begin{subfigure}{0.237\textwidth}
        \centering
        \includegraphics[width=\linewidth]{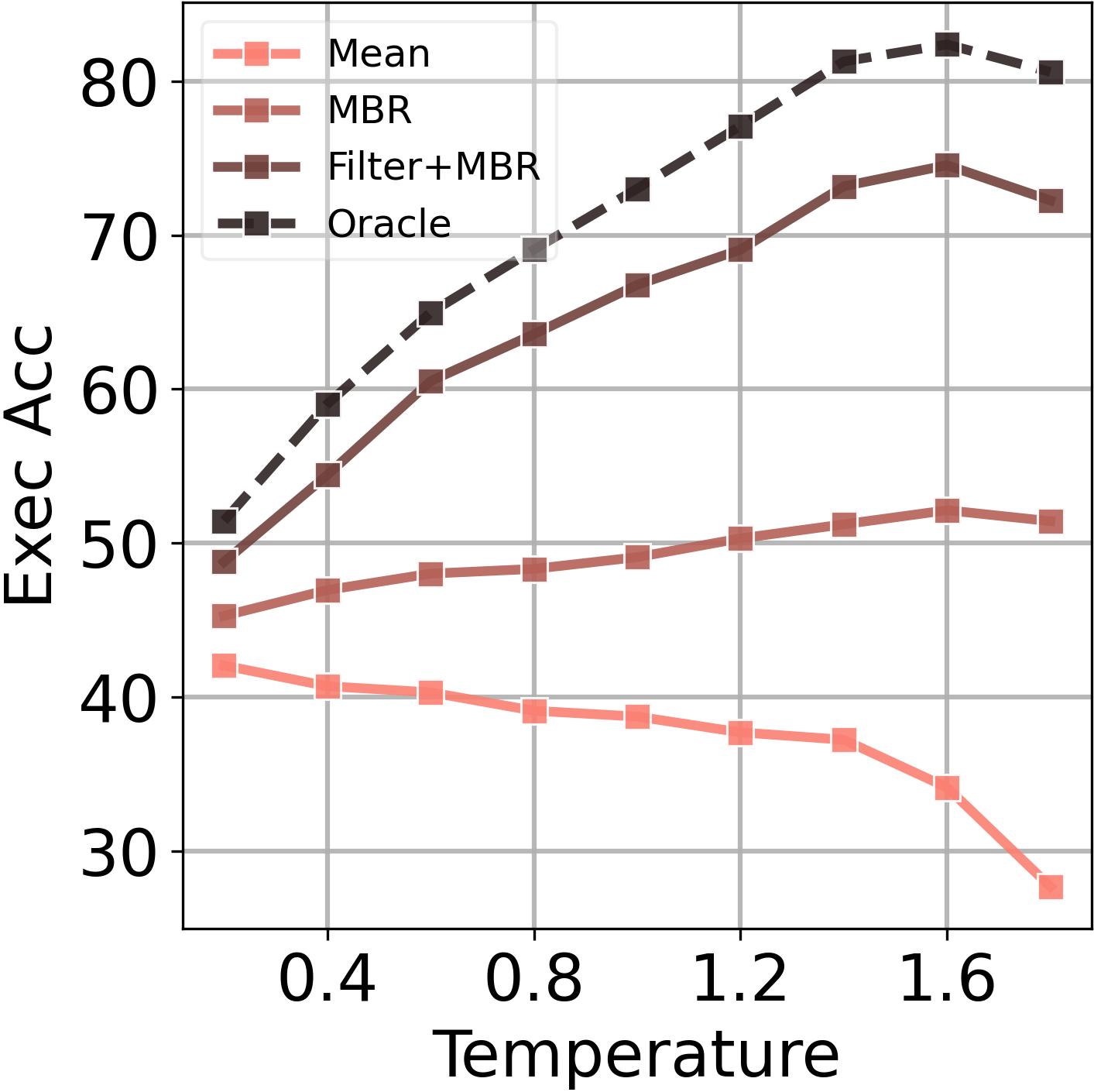}
        \caption{HumanEval}
        \label{fig:cl7b_temp_humaneval_base}
    \end{subfigure}\hfill
    \begin{subfigure}{0.237\textwidth}
        \centering
        \includegraphics[width=\linewidth]{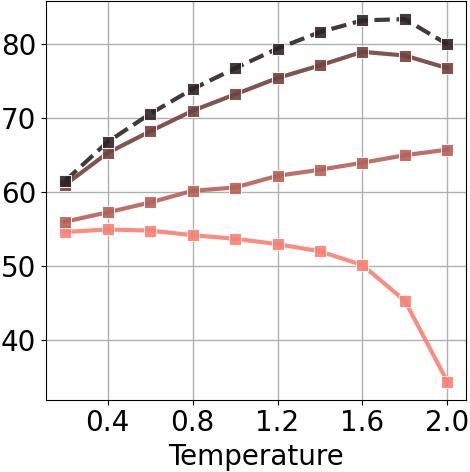}
        \caption{MBPP-S}
        \label{fig:cl7b_temp_mbpp_base}
    \end{subfigure}
    \begin{subfigure}{0.237\textwidth}
        \centering
        \includegraphics[width=\linewidth]{figures/humaneval_temp_all.jpg}
        \caption{HumanEval+}
        \label{fig:cl7b_temp_humaneval_plus}
    \end{subfigure}\hfill
    \begin{subfigure}{0.237\textwidth}
        \centering
        \includegraphics[width=\linewidth]{figures/mbpp_temp_all.jpg}
        \caption{MBPP-S+}
        \label{fig:cl7b_temp_mbpp_plus}
    \end{subfigure}
    \caption{Performance of MBR-Exec with and without filtering over sampling temperatures using CodeLlama-7B-Instruct with 50 generated candidates. We also provide the oracle of reranking with dashed lines. Results that end with $+$ mean that it's evaluated on the plus with extended test cases, otherwise it's evaluated on the original test cases. Results are averaged across 4 runs.}
    \label{fig:cl7b_temperature}
\end{figure*}

\begin{table*}[!tb]
\small
\centering
\scalebox{0.93}{
\begin{tabular}{lcccccc@{}}
\toprule
\multirow{2}{*}{Model} & \multicolumn{2}{c}{DS-6.7B-Instruct} & \multicolumn{2}{c}{CL-13B-Instruct} & \multicolumn{2}{c}{DS-V2-Lite-Instruct} \\
 &  temp = 0.8 & temp = 1.2 & temp = 0.8 & temp = 1.6 (1.2 for LCB) & temp = 0.8 & temp = 1.2 \\ \midrule
\multicolumn{7}{c}{HumanEval(+)} \\ \midrule
Mean 
& 77.3 (70.6) 
& 72.5 \textcolor{red}{\scriptsize -4.8} (65.1 \textcolor{red}{\scriptsize -5.5}) 
& 45.7 (40.0) 
& 35.8 \textcolor{red}{\scriptsize -9.9} (31.1 \textcolor{red}{\scriptsize -8.9}) 
& 81.7 (76.6) 
& 81.7  \textcolor{red}{\scriptsize -0} (76.6 \textcolor{red}{\scriptsize -0}) 
\\
MBR 
&  87.0 (85.8) 
& 86.4 \textcolor{red}{\scriptsize -0.6} (85.5 \textcolor{red}{\scriptsize -0.3}) 
& 59.0 (54.4) 
& 61.3 \textcolor{plusgreen}{\scriptsize +2.3} (59.6 \textcolor{plusgreen}{\scriptsize +5.2}) 
& 84.8 (81.3) 
& 85.8 \textcolor{plusgreen}{\scriptsize +1.0} (83.5 \textcolor{plusgreen}{\scriptsize +2.2})\\ 
Filter + MBR 
& 90.7 (89.8) 
& 91.6 \textcolor{plusgreen}{\scriptsize +0.9} (90.7 \textcolor{plusgreen}{\scriptsize +0.9}) 
& 76.5 (69.2) 
& 80.5 \textcolor{plusgreen}{\scriptsize +4.0} (76.1 \textcolor{plusgreen}{\scriptsize +6.9}) 
& 91.5 (87.2) 
& 92.4 \textcolor{plusgreen}{\scriptsize +1.0} (89.6 \textcolor{plusgreen}{\scriptsize +2.4}) \\
Oracle 
& 92.9 (91.5) 
& 95.2 \textcolor{plusgreen}{\scriptsize +2.3} (93.2 \textcolor{plusgreen}{\scriptsize +1.7}) 
& 82.8 (75.1) 
& 88.7 \textcolor{plusgreen}{\scriptsize +5.9} (82.4 \textcolor{plusgreen}{\scriptsize +7.3}) 
& 94.5 (89.7) 
& 95.7 \textcolor{plusgreen}{\scriptsize +1.2} (92.3 \textcolor{plusgreen}{\scriptsize +2.6}) 
\\
\midrule
\multicolumn{7}{c}{MBPP(+)} 
\\ \midrule
Mean 
& 71.4 (61.8)
& 68.6 \textcolor{red}{\scriptsize -2.8} (58.6 \textcolor{red}{\scriptsize -3.2}) 
& 60.7 (51.0)
& 52.5 \textcolor{red}{\scriptsize -8.2} (43.5 \textcolor{red}{\scriptsize -7.5}) 
& 79.7 (67.2) 
& 79.1 \textcolor{red}{\scriptsize -0.6} (66.8 \textcolor{red}{\scriptsize -0.4})
\\
MBR 
& 83.7 (78.2)
& 84.0 \textcolor{plusgreen}{\scriptsize +0.3} (78.6 \textcolor{plusgreen}{\scriptsize +0.4}) 
& 67.5 (60.7)
& 73.9 \textcolor{plusgreen}{\scriptsize +6.4} (67.2 \textcolor{plusgreen}{\scriptsize +6.5}) 
& 88.6 (78.5)
& 89.1 \textcolor{plusgreen}{\scriptsize +0.5} (80.1 \textcolor{plusgreen}{\scriptsize +1.6})
\\
Filter + MBR 
& 87.3 (80.5)
& 89.1 \textcolor{plusgreen}{\scriptsize +1.8} (82.1 \textcolor{plusgreen}{\scriptsize +1.6}) 
& 78.0 (67.8)
& 84.1 \textcolor{plusgreen}{\scriptsize +6.1} (74.7 \textcolor{plusgreen}{\scriptsize +6.9}) 
& 89.7 (79.6) 
& 91.2 \textcolor{plusgreen}{\scriptsize +1.5} (81.8 \textcolor{plusgreen}{\scriptsize +2.2})
\\
Oracle 
& 91.2 (82.8)
& 93.2 \textcolor{plusgreen}{\scriptsize +2.0} (85.6 \textcolor{plusgreen}{\scriptsize +2.8}) 
& 82.0 (70.4)
& 88.4 \textcolor{plusgreen}{\scriptsize +6.4} (78.0 \textcolor{plusgreen}{\scriptsize +7.4}) 
& 91.3 (80.9) 
& 93.1 \textcolor{plusgreen}{\scriptsize +1.8} (83.9 \textcolor{plusgreen}{\scriptsize +3.0})
\\
\midrule
\multicolumn{7}{c}{LiveCodeBench} 
\\ \midrule
Mean 
& 19.9 
& 17.5 \textcolor{red}{\scriptsize -2.4}
& 17.4 
& 15.9 \textcolor{red}{\scriptsize -1.5} 
& 36.1 
& 35.4 \textcolor{red}{\scriptsize -0.7} 
\\
MBR 
& 30.0 
& 30.6 \textcolor{plusgreen}{\scriptsize +0.6}
& 24.9 
& 27.0 \textcolor{plusgreen}{\scriptsize +2.1}
& 42.6 
& 45.8 \textcolor{plusgreen}{\scriptsize +3.2} 
\\
Filter + MBR 
& 43.1 
& 41.9 \textcolor{red}{\scriptsize -1.2}
& 35.7 
& 40.6 \textcolor{plusgreen}{\scriptsize +4.9} 
& 57.8 
& 58.7 \textcolor{plusgreen}{\scriptsize +0.9}
\\
Oracle 
& 52.5 
& 53.6 \textcolor{plusgreen}{\scriptsize +1.1}
& 46.8 
& 50.4 \textcolor{plusgreen}{\scriptsize +3.6} 
& 63.1 
& 68.2 \textcolor{plusgreen}{\scriptsize +5.1}
\\
\bottomrule
\end{tabular}}
\caption{Comparison of performance of sampling and reranking using temperature 0.8 and those chosen for further experiments. We report mean execution accuracies, MBR-Exec results, and oracle performances of 50 candidates generated by DeepSeekCoder-\{6.7B, V2-Lite\}-Instruct or CodeLlama-13B-Instruct. Results that end with + mean that it is evaluated on the plus with extended test cases, otherwise, it is evaluated on the original test cases. We also show \textcolor{red}{decreases} and \textcolor{plusgreen}{improvements} of results in our choice of temperature over 0.8. Results are averaged across 4 runs for HumanEval(+) and MBPP-S(+), and 2 runs for LiveCodeBench.}
\label{tab:compare_temp}
\end{table*}

\begin{table}[!tb]
\small
\centering
\scalebox{1}{
\begin{tabular}{lcc@{}}
\toprule
& \multirow{2}{*}{HE (+)} & \multirow{2}{*}{MBPP-S (+)} \\
 &  &  \\ \midrule
Random & 34.1 (30.2) & 50.1 (41.9) \\ 
Greedy & 43.3 (39.0) & 53.2 (44.8) \\ 
Oracle & 82.4 (76.2) & 83.2 (72.0) \\
\midrule
\multicolumn{3}{c}{N-Best Reranking} \\ \midrule
LL & 45.1 (38.7) & 54.2 (42.6) \\
CR & 46.0 (40.5) & 53.1 (43.1) \\
CS & 33.1 (30.2) & - \\

\rowcolor{softblue}
Filter & 70.0 (59.7) & 77.4 (\textbf{60.8}) \\
\rowcolor{softblue}
Filter + LL & 71.0 (60.4) & \underline{78.1} (59.7) \\
\rowcolor{softblue}
Filter + CR & \textbf{71.3} (\textbf{61.3}) & \textbf{78.5} (\underline{60.2}) \\
\rowcolor{softblue}
Filter + CS & \underline{71.2} (\underline{61.0}) & - \\ 
\midrule

\multicolumn{3}{c}{MBR} \\ \midrule
MBR-CBS & 40.5 (35.5) & 55.3 (45.4) \\
MBR-CS & 37.5 (31.9) & - \\
MBR-Exec & 52.1 (49.5) & 64.0 (57.8) \\
\midrule
\multicolumn{3}{c}{N-Best Reranking + MBR} \\ \midrule
LL + MBR-Exec & 51.8 (49.4) & 64.4 (58.1) \\
CR + MBR-Exec & 51.4 (48.9) & 64.8 (58.7) \\
\rowcolor{softblue}
Filter + MBR-CBS & 69.8 (60.1) & 77.3 (61.3) \\
\rowcolor{softblue}
Filter + MBR-CS & 71.5 (59.9) & - \\
\rowcolor{softblue}
Filter + MBR-Exec & \textbf{74.5} (\underline{70.1}) & \underline{79.0} (\underline{69.3}) \\
\rowcolor{softblue}
Filter + LL + MBR-Exec & \textbf{74.5} (\textbf{70.6}) & 78.9 (69.2) \\
\rowcolor{softblue}
Filter + CR + MBR-Exec & 74.0 (70.0) & \textbf{79.3} (\textbf{69.4}) \\
\bottomrule
\end{tabular}}
\caption{Comparison of reranking methods using CodeLlama-7B-Instruct on HumanEval(+) and MBPP(+). For N-Best Reranking, we compare log-likelihood (LL), Coder-Reviewer (CR), and CodeScore (CS). For MBR, we compare CodeScore, CodeBertScore (CBS), and MBR-Exec. \textbf{HE} stands for the HumanEval benchmark, $+$ refers to evaluation on extended test cases on EvalPlus. We also highlighted \textbf{best} and \underline{second best} reranking results of the class of reranking methods.}
\label{tab:compare_reranking}
\end{table}

\begin{table}[htbp]
\small
\centering
\scalebox{1}{
\begin{tabular}{lcc@{}}
\toprule
& \multirow{2}{*}{HE (+)} & \multirow{2}{*}{MBPP-S (+)} \\
 &  &  \\ \midrule
Random & 72.5 (65.1) & 68.6 (58.6) \\ 
Greedy & 82.3 (76.8) & 72.2 (64.3) \\ 
Oracle & 95.2 (93.2) & 93.2 (85.6) \\
\midrule
\multicolumn{3}{c}{N-Best Reranking} \\ \midrule
LL & 81.4 (71.8) & 70.1 (58.9) \\
CR & 78.7 (70.3) & 76.1 (64.0) \\
CS & 70.1 (63.6) & - \\

\rowcolor{softblue}
Filter & 88.4 (78.7) & 87.4 (72.6) \\
\rowcolor{softblue}
Filter + LL & \textbf{90.5} (\underline{79.1}) & \textbf{88.5} (\textbf{72.8}) \\
\rowcolor{softblue}
Filter + CR & \underline{89.5} (\textbf{79.6}) & \underline{88.2} (\textbf{72.8}) \\
\rowcolor{softblue}
Filter + CS & 87.7 (78.4) & - \\ 
\midrule

\multicolumn{3}{c}{MBR} \\ \midrule
MBR-CBS & 81.1 (73.0) & 72.6 (63.4) \\
MBR-CS & 74.2 (67.5) & - \\
MBR-Exec & 86.4 (85.5) & 84.0 (78.6) \\
\midrule
\multicolumn{3}{c}{N-Best Reranking + MBR} \\ \midrule
LL + MBR-Exec & 86.6 (85.7) & 84.4 (79.0) \\
CR + MBR-Exec & 86.6 (85.7) & 85.4 (79.7) \\
\rowcolor{softblue}
Filter + MBR-CBS & 90.7 (81.6) & 88.1 (75.3)  \\
\rowcolor{softblue}
Filter + MBR-CS & 89.6 (80.3) & - \\
\rowcolor{softblue}
Filter + MBR-Exec & 91.6 (90.7) & 89.1 (82.1) \\
\rowcolor{softblue}
Filter + LL + MBR-Exec & \textbf{91.9} (\textbf{90.9}) & \underline{89.5} (\underline{82.4}) \\
\rowcolor{softblue}
Filter + CR + MBR-Exec & \textbf{91.9} (\textbf{90.9}) & \textbf{89.6} (\textbf{82.5}) \\
\bottomrule
\end{tabular}}
\caption{Comparison of reranking methods using DeepSeekCoder-6.7B-Instruct on HumanEval(+) and MBPP(+). For N-Best Reranking, we compare log-likelihood (LL), Coder-Reviewer (CR), and CodeScore (CS). For MBR, we compare CodeScore, CodeBertScore (CBS), and MBR-Exec. \textbf{HE} stands for the HumanEval benchmark, $+$ refers to evaluation on extended test cases on EvalPlus. We also highlighted \textbf{best} and \underline{second best} reranking results of the class of reranking methods.}
\label{tab:ds_6.7b_compare_reranking}
\end{table}

\begin{table}[htbp]
\small
\centering
\scalebox{1}{
\begin{tabular}{lcc@{}}
\toprule
& \multirow{2}{*}{HE (+)} & \multirow{2}{*}{MBPP-S (+)} \\
 &  &  \\ \midrule
Random & 35.8 (31.1) & 52.5 (43.5) \\ 
Greedy & 45.7 (39.6) & 62.5 (52.7) \\ 
Oracle & 88.7 (82.4) & 88.4 (78.0) \\
\midrule
\multicolumn{3}{c}{N-Best Reranking} \\ \midrule
LL & 54.0 (45.6) & 57.9 (47.2) \\
CR & 51.1 (43.8) & 60.4 (49.5) \\
CS & 35.8 (31.7) & - \\

\rowcolor{softblue}
Filter & 75.4 (63.4) & 81.9 (\underline{64.7}) \\
\rowcolor{softblue}
Filter + LL & \textbf{77.4} (\textbf{65.7}) & \textbf{82.8} (64.1) \\
\rowcolor{softblue}
Filter + CR & \underline{76.8} (\textbf{65.7}) & \underline{82.7} (\textbf{65.0}) \\
\rowcolor{softblue}
Filter + CS & 76.7 (\textbf{65.7}) & - \\ 
\midrule

\multicolumn{3}{c}{MBR} \\ \midrule
MBR-CBS & 47.3 (40.2) & 64.4 (53.5) \\
MBR-CS & 37.5 (33.4) & - \\
MBR-Exec & 61.3 (59.6) & 73.9 (67.2) \\
\midrule
\multicolumn{3}{c}{N-Best Reranking + MBR} \\ \midrule
LL + MBR-Exec & 61.6 (59.8) & 73.8 (67.0) \\
CR + MBR-Exec & 62.5 (60.5) & 75.1 (68.2) \\
\rowcolor{softblue}
Filter + MBR-CBS & 75.9 (64.5) & 82.5 (65.8) \\
\rowcolor{softblue}
Filter + MBR-CS & 77.0 (65.7) & - \\
\rowcolor{softblue}
Filter + MBR-Exec & \textbf{80.5} (\textbf{76.1}) & \underline{84.1} (\underline{74.7}) \\
\rowcolor{softblue}
Filter + LL + MBR-Exec & \underline{79.9} (\underline{75.5}) & 83.9 (74.5) \\
\rowcolor{softblue}
Filter + CR + MBR-Exec & 79.6 (75.3) & \textbf{84.2} (\textbf{74.8}) \\
\bottomrule
\end{tabular}}
\caption{Comparison of reranking methods using CodeLlama-13B-Instruct on HumanEval(+) and MBPP(+). For N-Best Reranking, we compare log-likelihood (LL), Coder-Reviewer (CR), and CodeScore (CS). For MBR, we compare CodeScore, CodeBertScore (CBS), and MBR-Exec. \textbf{HE} stands for the HumanEval benchmark, $+$ refers to evaluation on extended test cases on EvalPlus. We also highlighted \textbf{best} and \underline{second best} reranking results of the class of reranking methods.}
\label{tab:cl_13b_compare_reranking}
\end{table}

\begin{figure}[htbp]
    \centering
    \begin{subfigure}{0.243\textwidth}
        \centering
        \includegraphics[width=\linewidth]{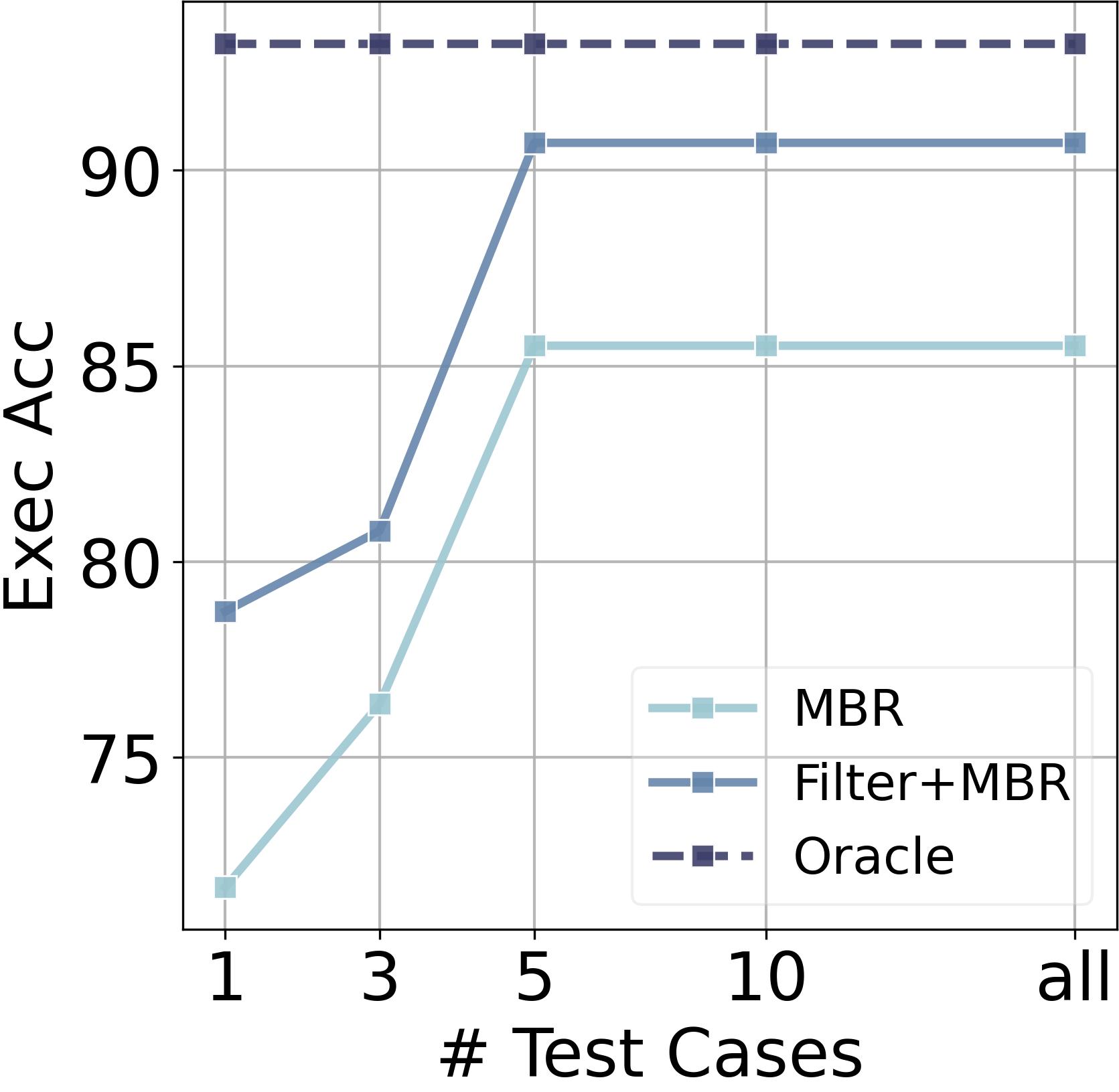}
        \caption{HumanEval+}
        \label{fig:ds_6.7b_num_uts_humaneval_plus}
    \end{subfigure}\hfill
    \begin{subfigure}{0.226\textwidth}
        \centering
        \includegraphics[width=\linewidth]{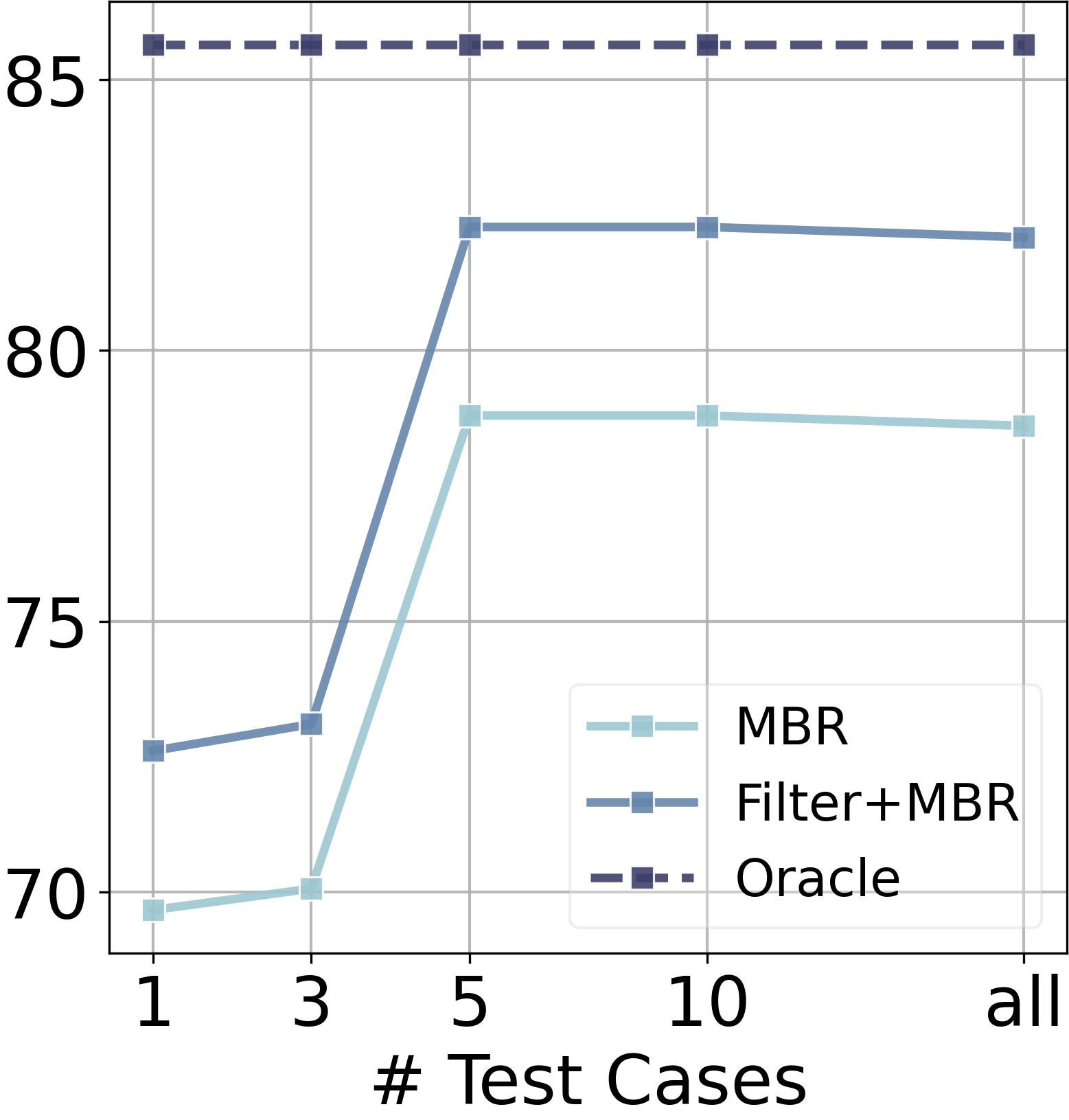}
        \caption{MBPP-S+}
        \label{fig:ds_6.7b_num_uts_mbpp_plus}
    \end{subfigure}
    \caption{Performance of MBR-Exec with fewer unit tests using candidates generated from DeepSeekCoder-6.7B-Instruct.}
    \label{fig:ds_6.7b_num_uts}
\end{figure}

\begin{figure}[tbp]
    \centering
    \begin{subfigure}{0.243\textwidth}
        \centering
        \includegraphics[width=\linewidth]{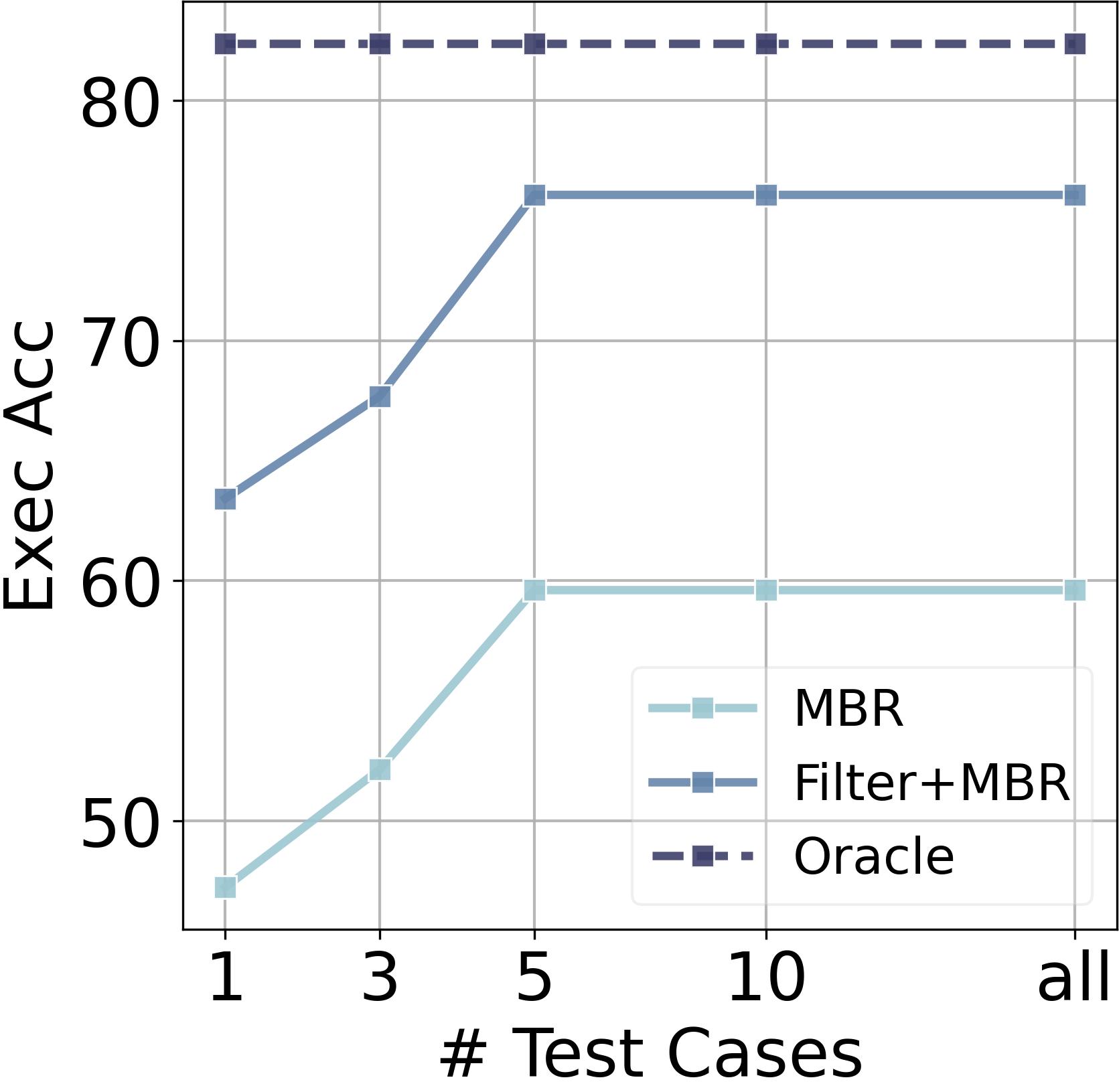}
        \caption{HumanEval+}
        \label{fig:cl_13b_num_uts_humaneval_plus}
    \end{subfigure}\hfill
    \begin{subfigure}{0.226\textwidth}
        \centering
        \includegraphics[width=\linewidth]{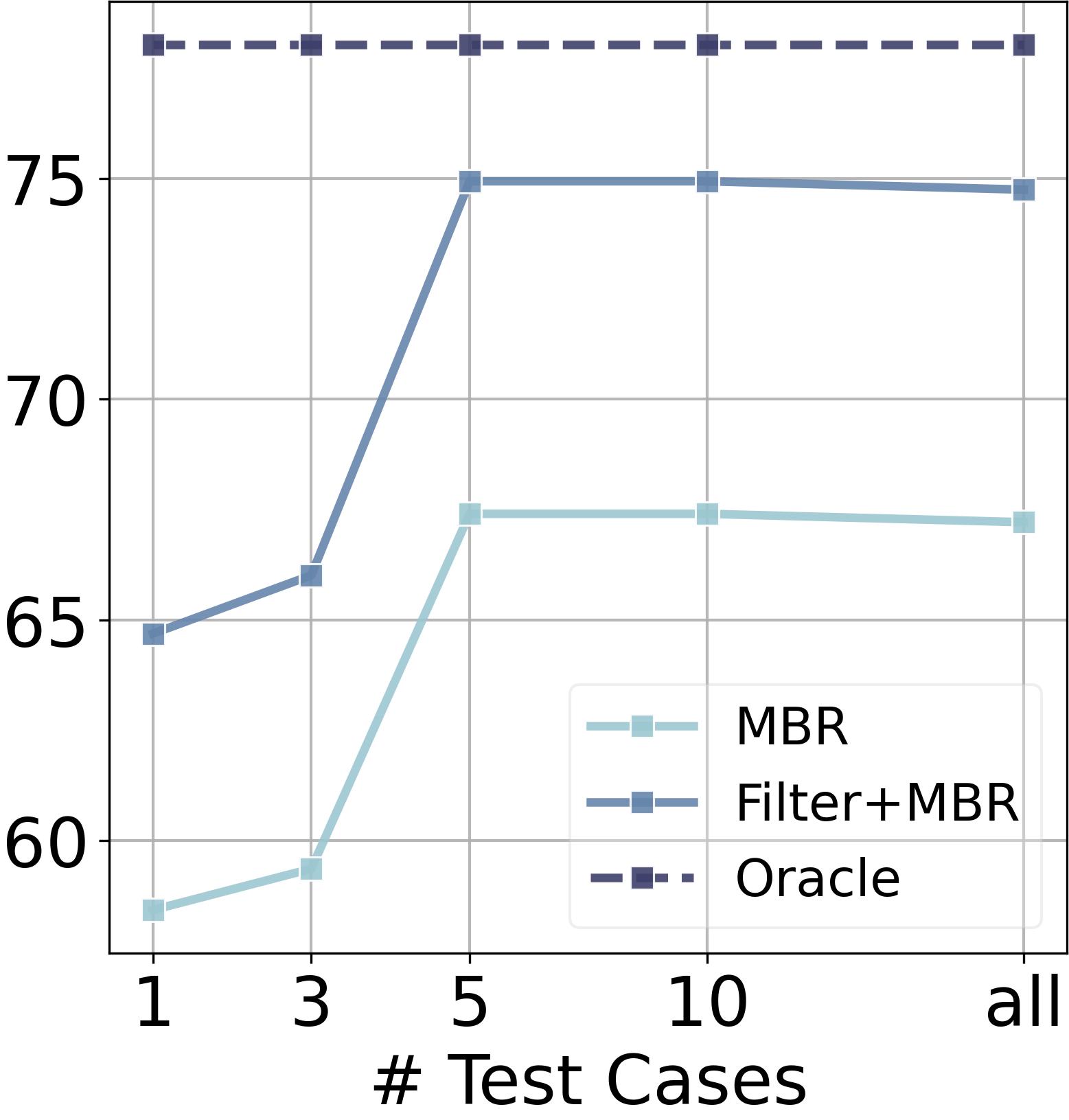}
        \caption{MBPP-S+}
        \label{fig:cl_13b_num_uts_mbpp_plus}
    \end{subfigure}
    \caption{Performance of MBR-Exec with fewer unit tests using candidates generated from CodeLlama-13B-Instruct.}
    \label{fig:cl_13b_num_uts}
\end{figure}

\begin{figure*}[tbp]
    \centering
    \begin{subfigure}{0.24\textwidth}
        \centering
        \includegraphics[width=\linewidth]{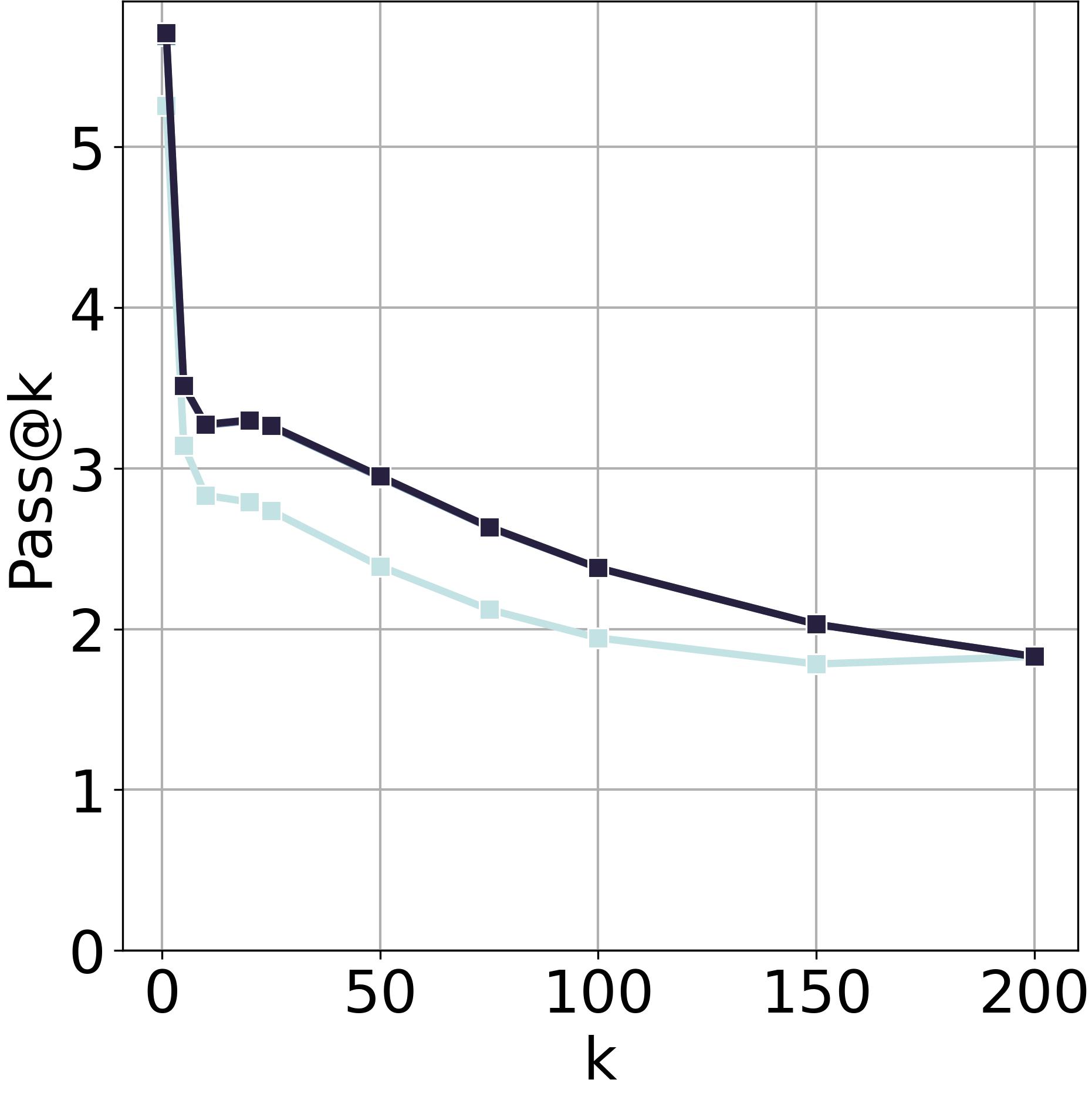}
        \caption{HumanEval}
        \label{fig:cl_7b_sd_upperbound_humaneval_base}
    \end{subfigure}\hfill
    \begin{subfigure}{0.226\textwidth}
        \centering
        \includegraphics[width=\linewidth]{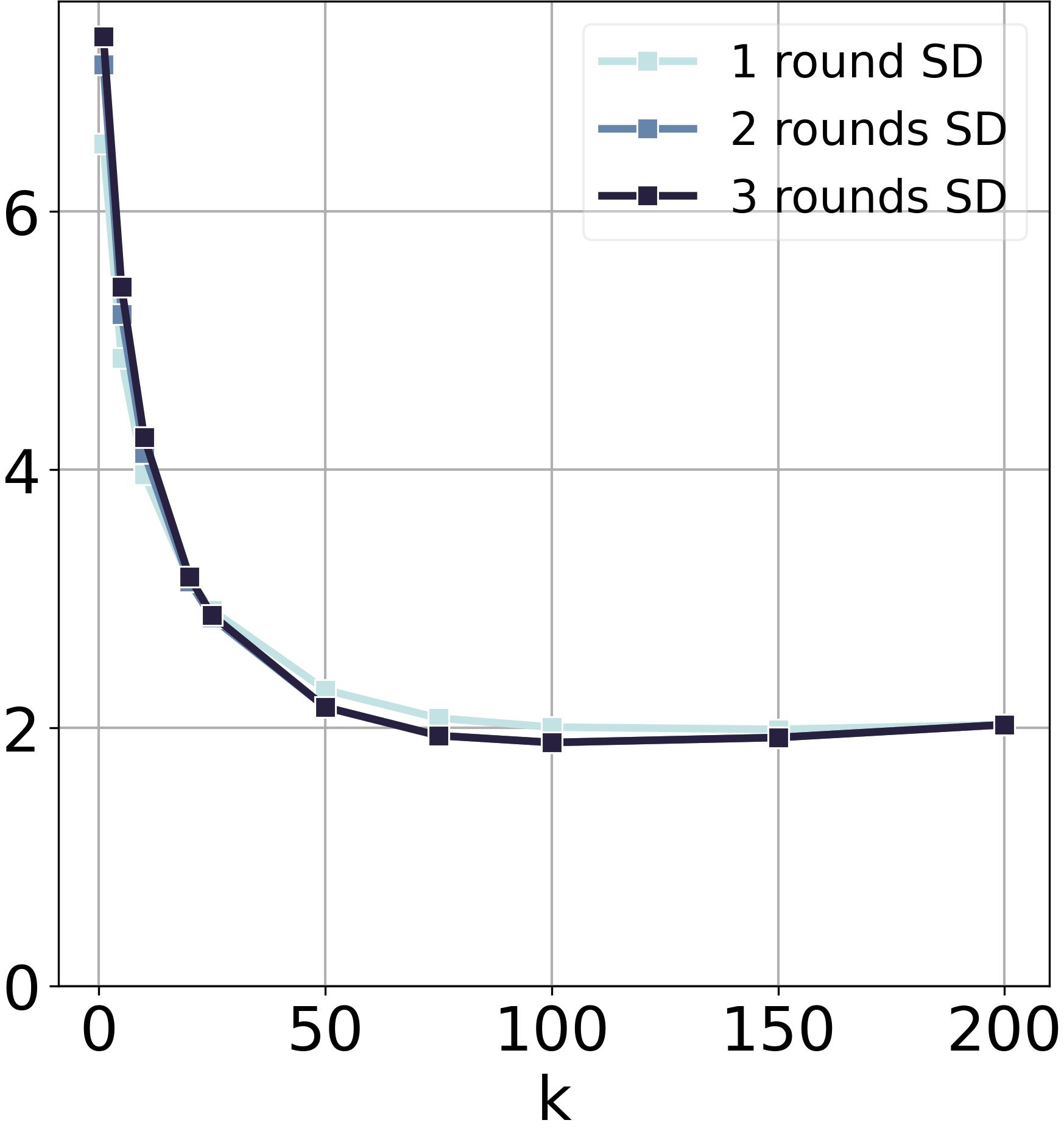}
        \caption{MBPP-S}
        \label{fig:cl_7b_sd_upperbound_mbpp_base}
    \end{subfigure}
    \begin{subfigure}{0.24\textwidth}
        \centering
        \includegraphics[width=\linewidth]{figures/humaneval_pass_at_k_performance.jpg}
        \caption{HumanEval+}
        \label{fig:cl_7b_sd_upperbound_humaneval_plus}
    \end{subfigure}\hfill
    \begin{subfigure}{0.226\textwidth}
        \centering
        \includegraphics[width=\linewidth]{figures/mbpp_pass_at_k_performance.jpg}
        \caption{MBPP-S+}
        \label{fig:cl_7b_sd_upperbound_mbpp_plus}
    \end{subfigure}
    \caption
{Improvement in Pass@k of CodeLlama-7B-Instruct after self-debugging compared to no self-debugging applied.}
    \label
{fig:cl_7b_sd_upperbound}
\end{figure*}

\begin{figure*}[tbp]
    \centering
    \begin{subfigure}{0.24\textwidth}
        \centering
        \includegraphics[width=\linewidth]{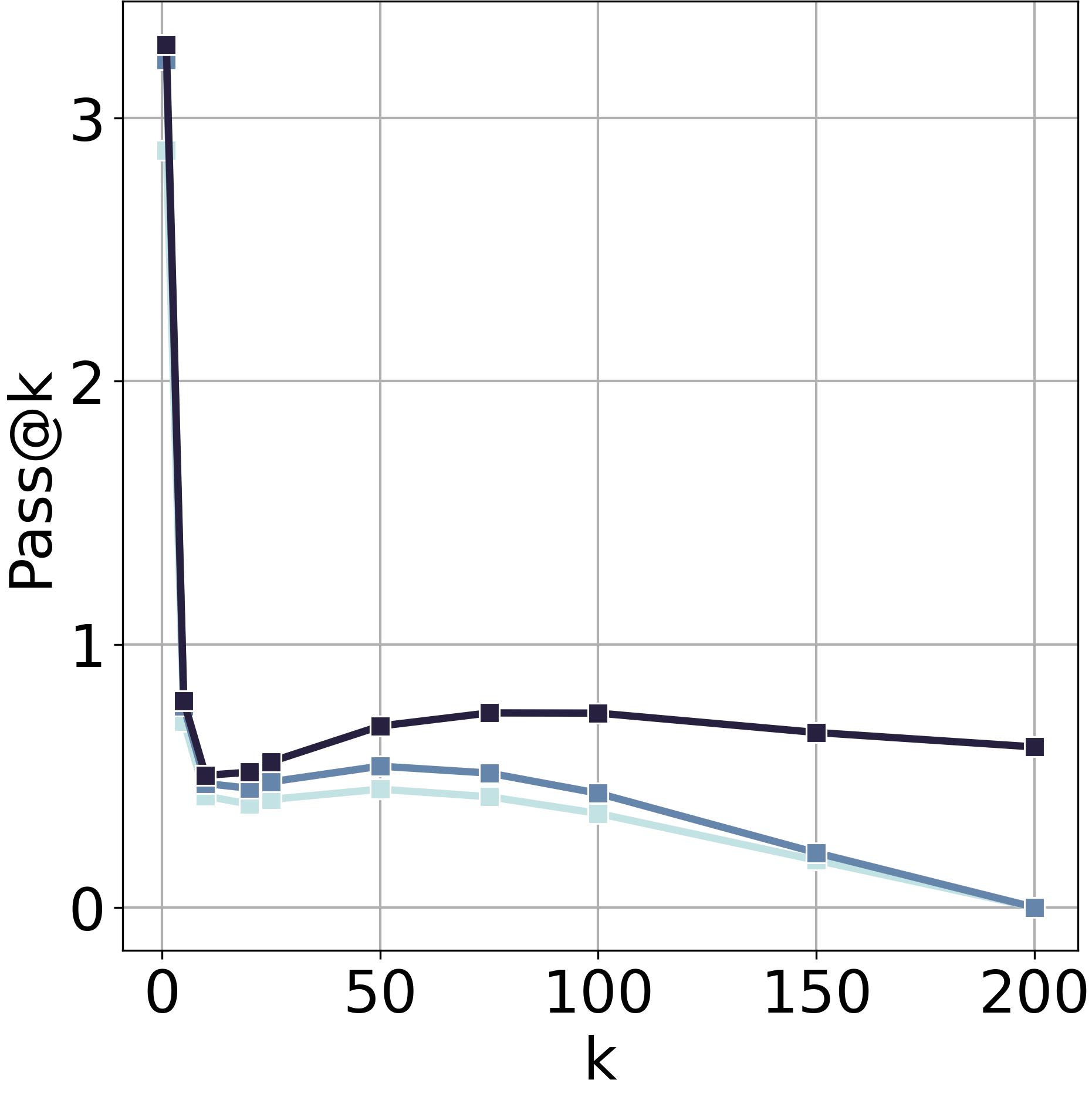}
        \caption{HumanEval}
        \label{fig:ds_6.7b_sd_upperbound_humaneval_base}
    \end{subfigure}\hfill
    \begin{subfigure}{0.226\textwidth}
        \centering
        \includegraphics[width=\linewidth]{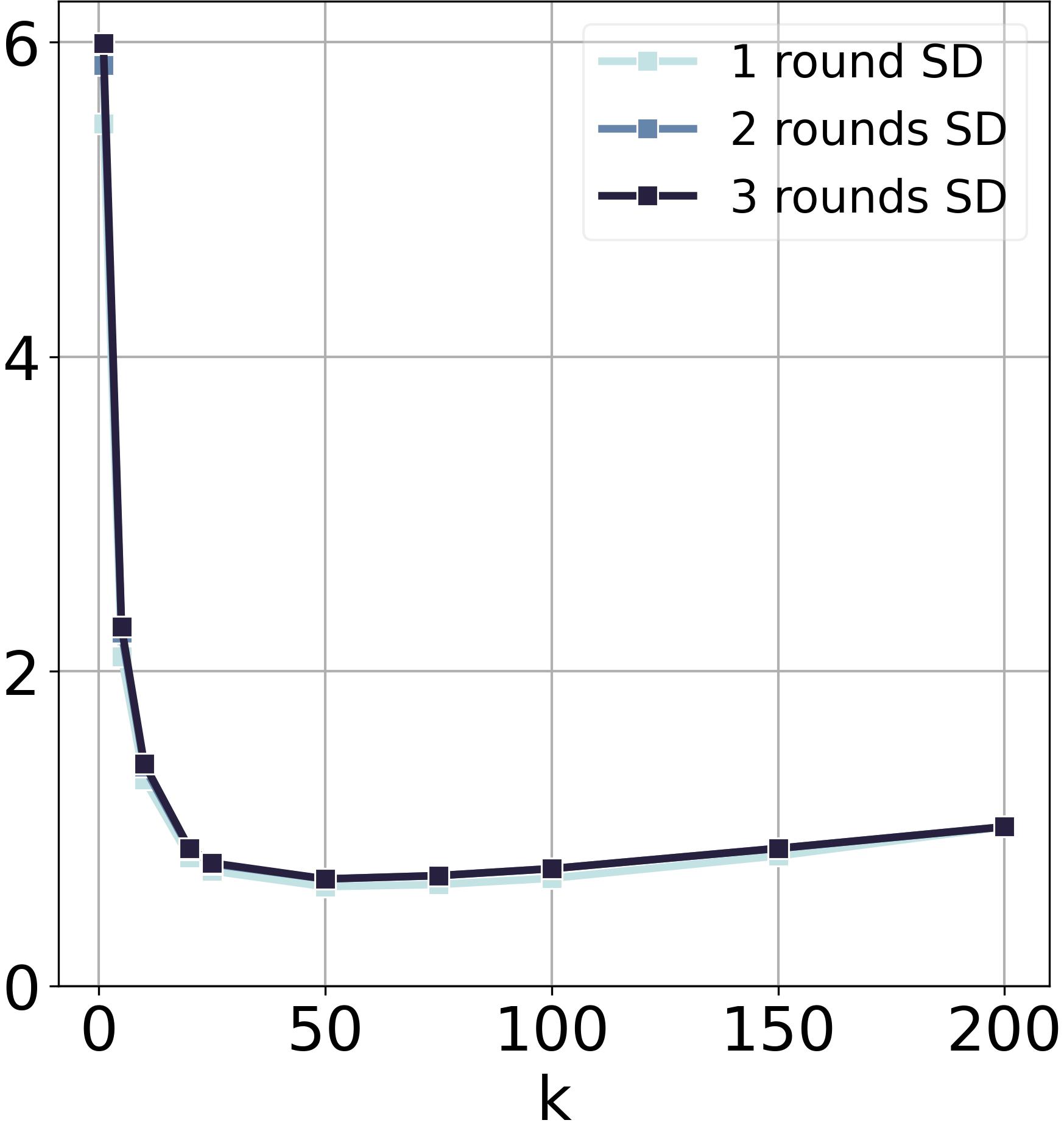}
        \caption{MBPP-S}
        \label{fig:ds_6.7b_sd_upperbound_mbpp_base}
    \end{subfigure}
    \begin{subfigure}{0.24\textwidth}
        \centering
        \includegraphics[width=\linewidth]{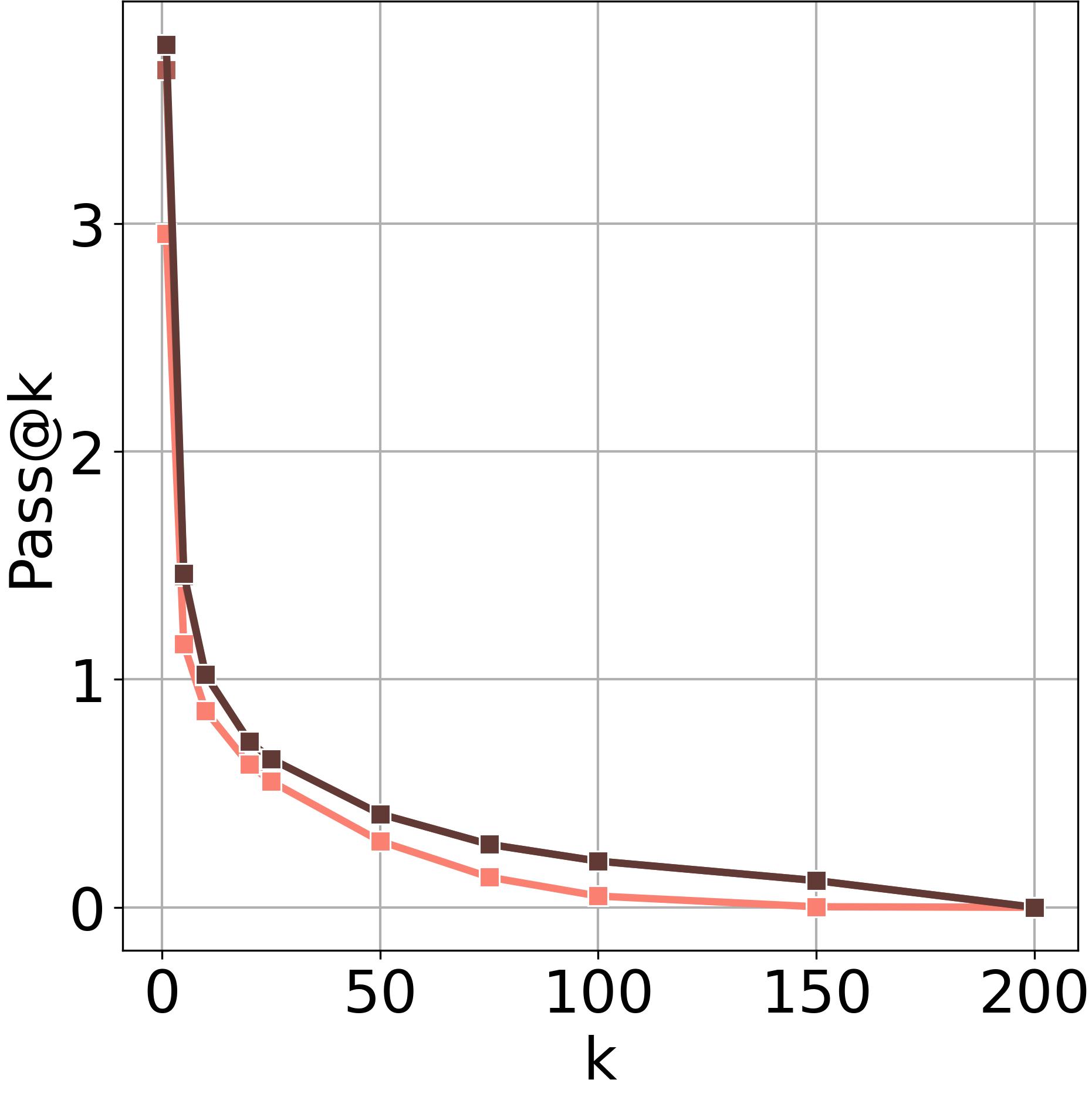}
        \caption{HumanEval+}
        \label{fig:ds_6.7b_sd_upperbound_humaneval_plus}
    \end{subfigure}\hfill
    \begin{subfigure}{0.226\textwidth}
        \centering
        \includegraphics[width=\linewidth]{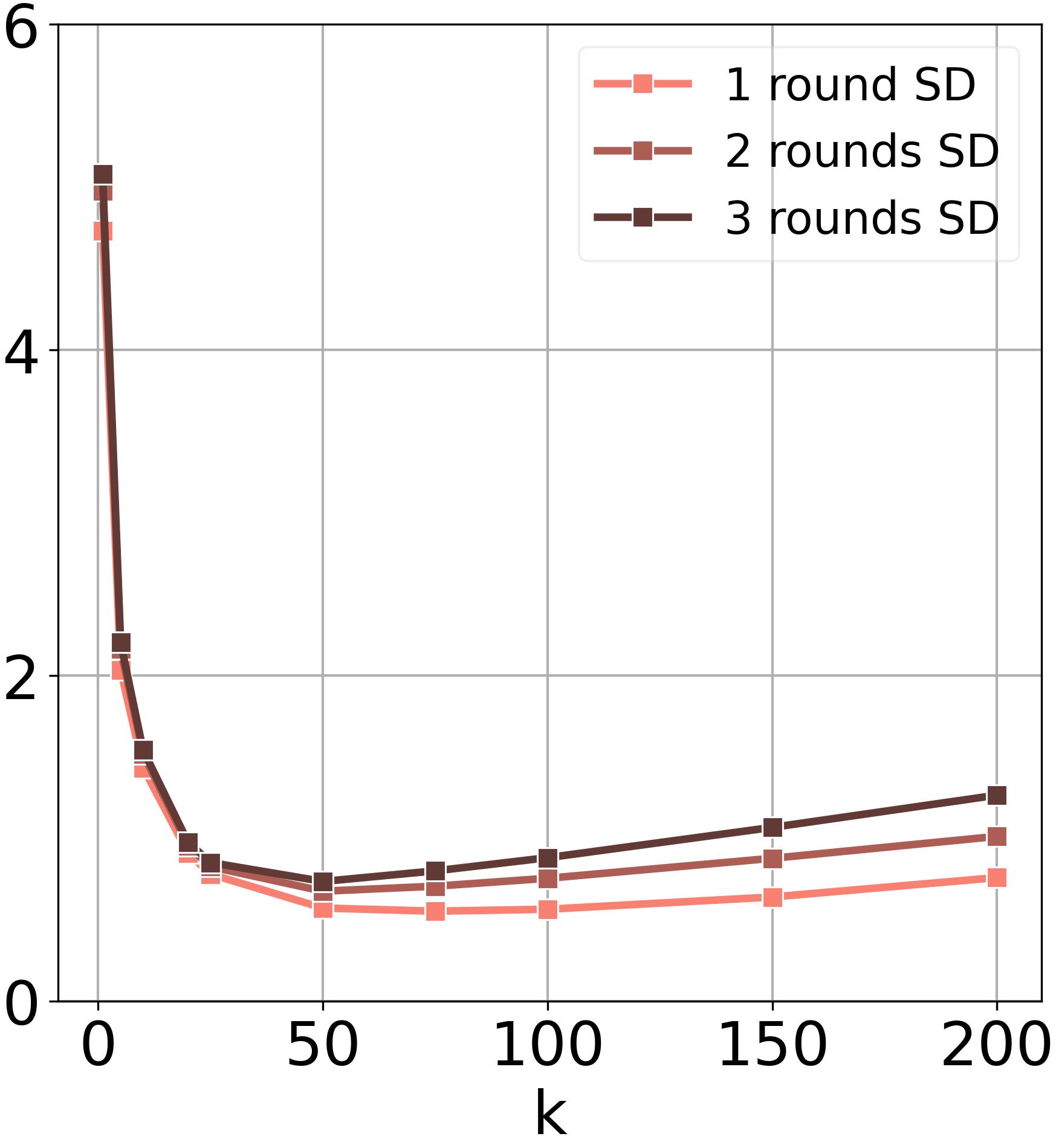}
        \caption{MBPP-S+}
        \label{fig:ds_6.7b_sd_upperbound_mbpp_plus}
    \end{subfigure}
    \caption
{Improvement in Pass@k of DeepSeekCoder-6.7B-Instruct after self-debugging compared to no self-debugging applied.}
    \label
{fig:ds_6.7b_sd_upperbound}
\end{figure*}

\begin{figure*}[tbp]
    \centering
    \begin{subfigure}{0.24\textwidth}
        \centering
        \includegraphics[width=\linewidth]{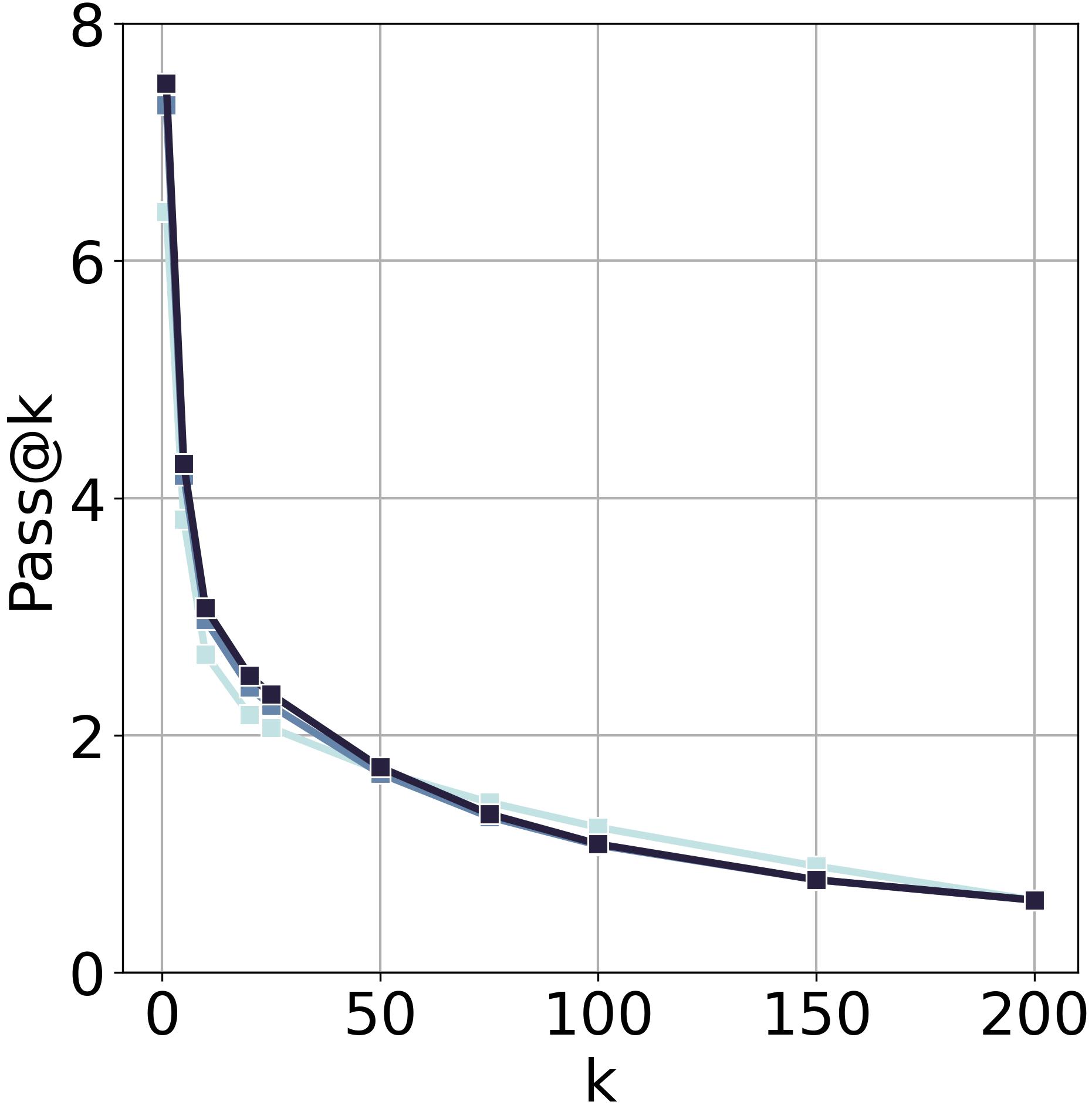}
        \caption{HumanEval}
        \label{fig:cl_13b_sd_upperbound_humaneval_base}
    \end{subfigure}\hfill
    \begin{subfigure}{0.226\textwidth}
        \centering
        \includegraphics[width=\linewidth]{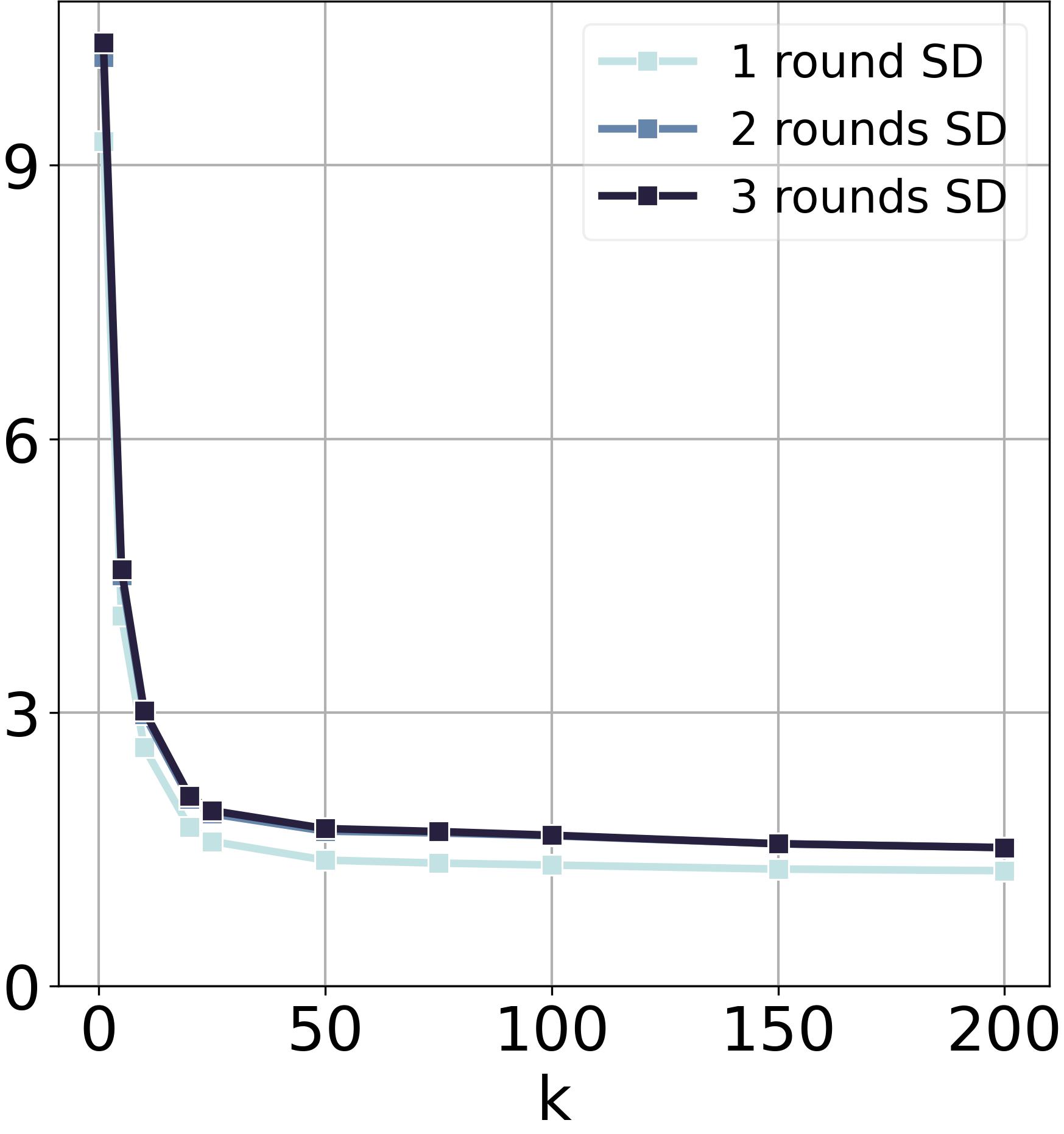}
        \caption{MBPP-S}
        \label{fig:cl_13b_sd_upperbound_mbpp_base}
    \end{subfigure}
    \begin{subfigure}{0.24\textwidth}
        \centering
        \includegraphics[width=\linewidth]{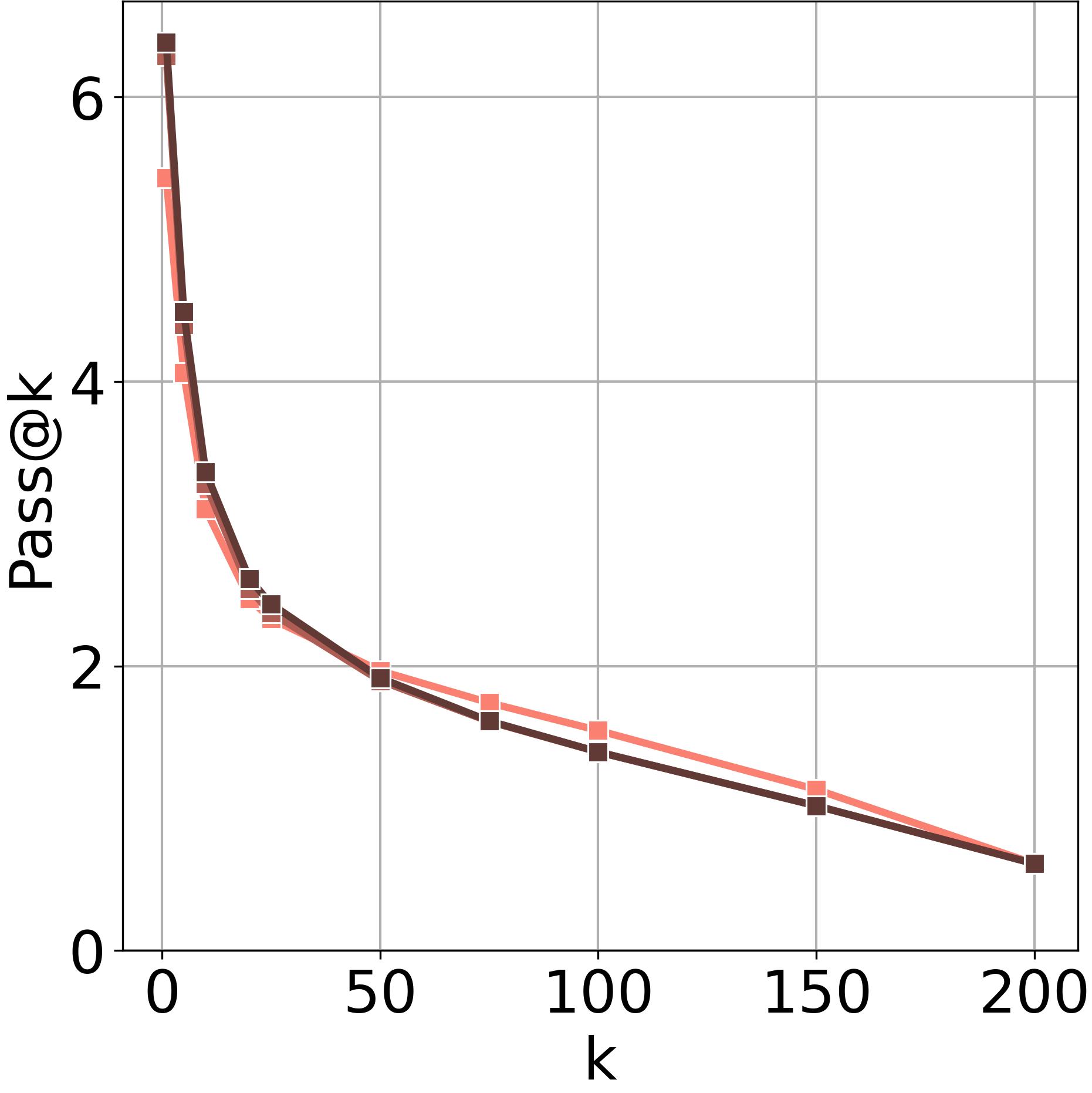}
        \caption{HumanEval+}
        \label{fig:cl_13b_sd_upperbound_humaneval_plus}
    \end{subfigure}\hfill
    \begin{subfigure}{0.226\textwidth}
        \centering
        \includegraphics[width=\linewidth]{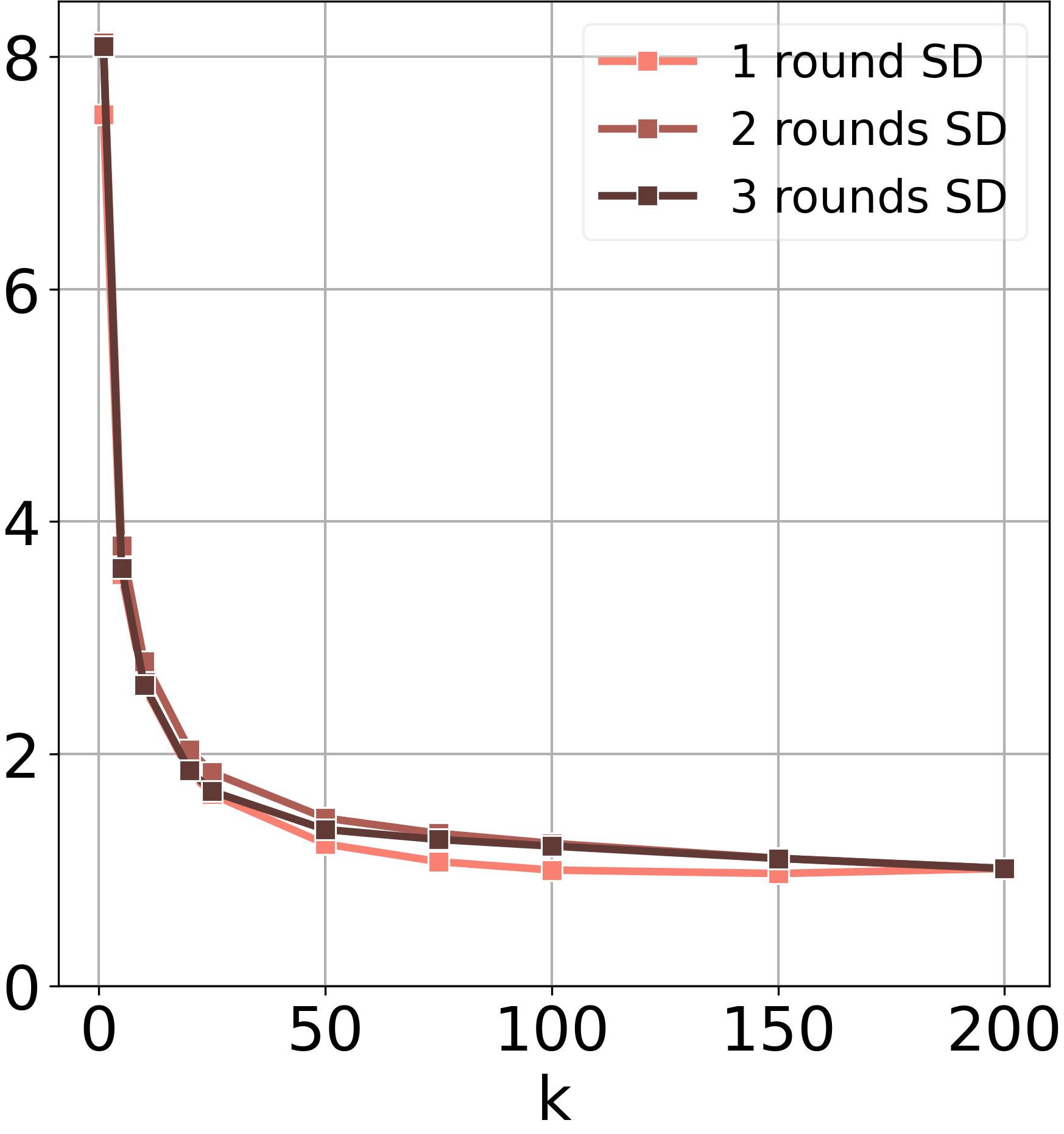}
        \caption{MBPP-S+}
        \label{fig:cl_13b_sd_upperbound_mbpp_plus}
    \end{subfigure}
    \caption
{Improvement in Pass@k of CodeLlama-13B-Instruct after self-debugging compared to no self-debugging applied.}
    \label
{fig:cl_13b_sd_upperbound}
\end{figure*}

\begin{figure*}[tbp]
    \centering
    \begin{subfigure}{0.24\textwidth}
        \centering
        \includegraphics[width=\linewidth]{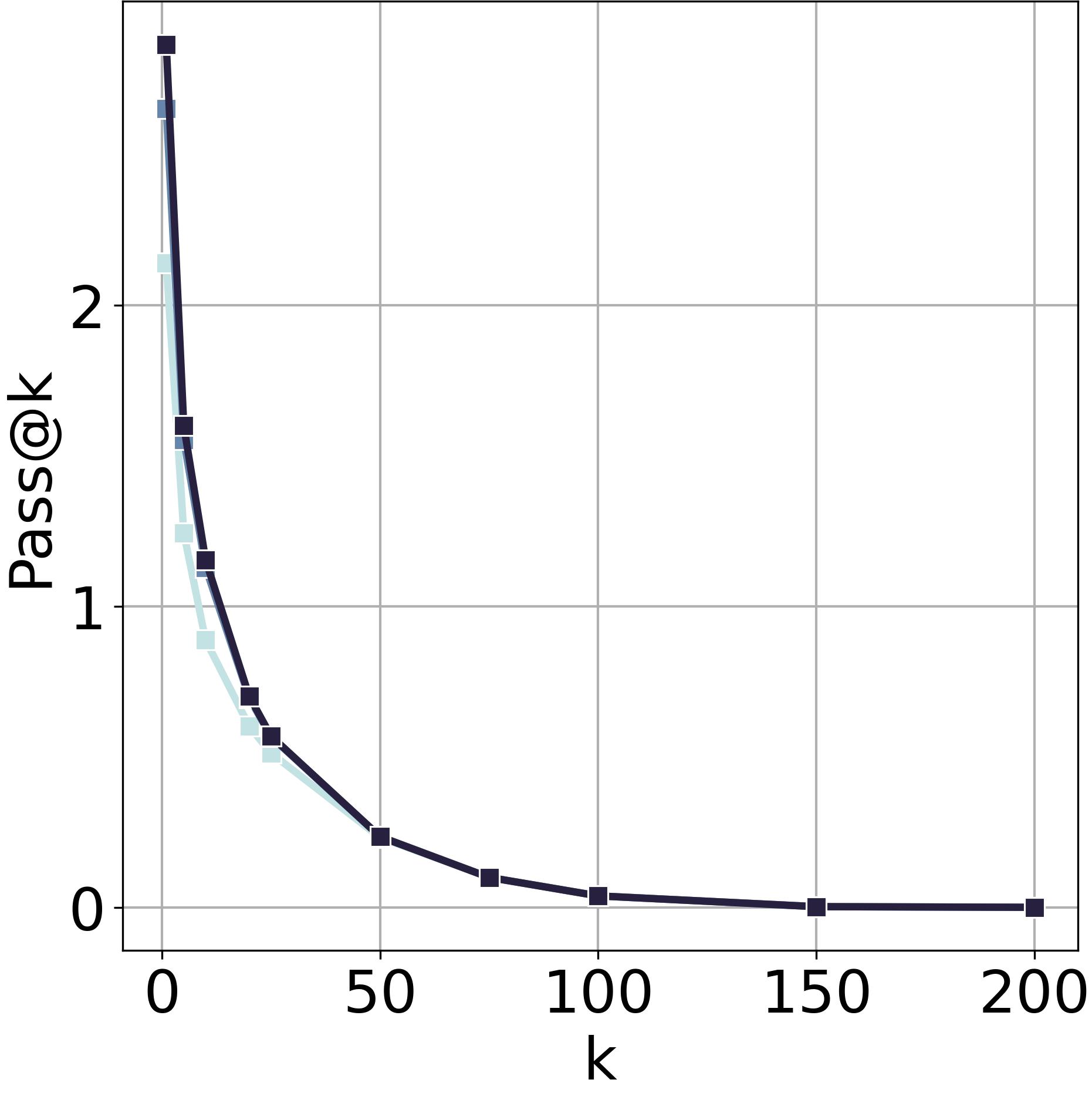}
        \caption{HumanEval}
        \label{fig:ds_16b_sd_upperbound_humaneval_base}
    \end{subfigure}\hfill
    \begin{subfigure}{0.226\textwidth}
        \centering
        \includegraphics[width=\linewidth]{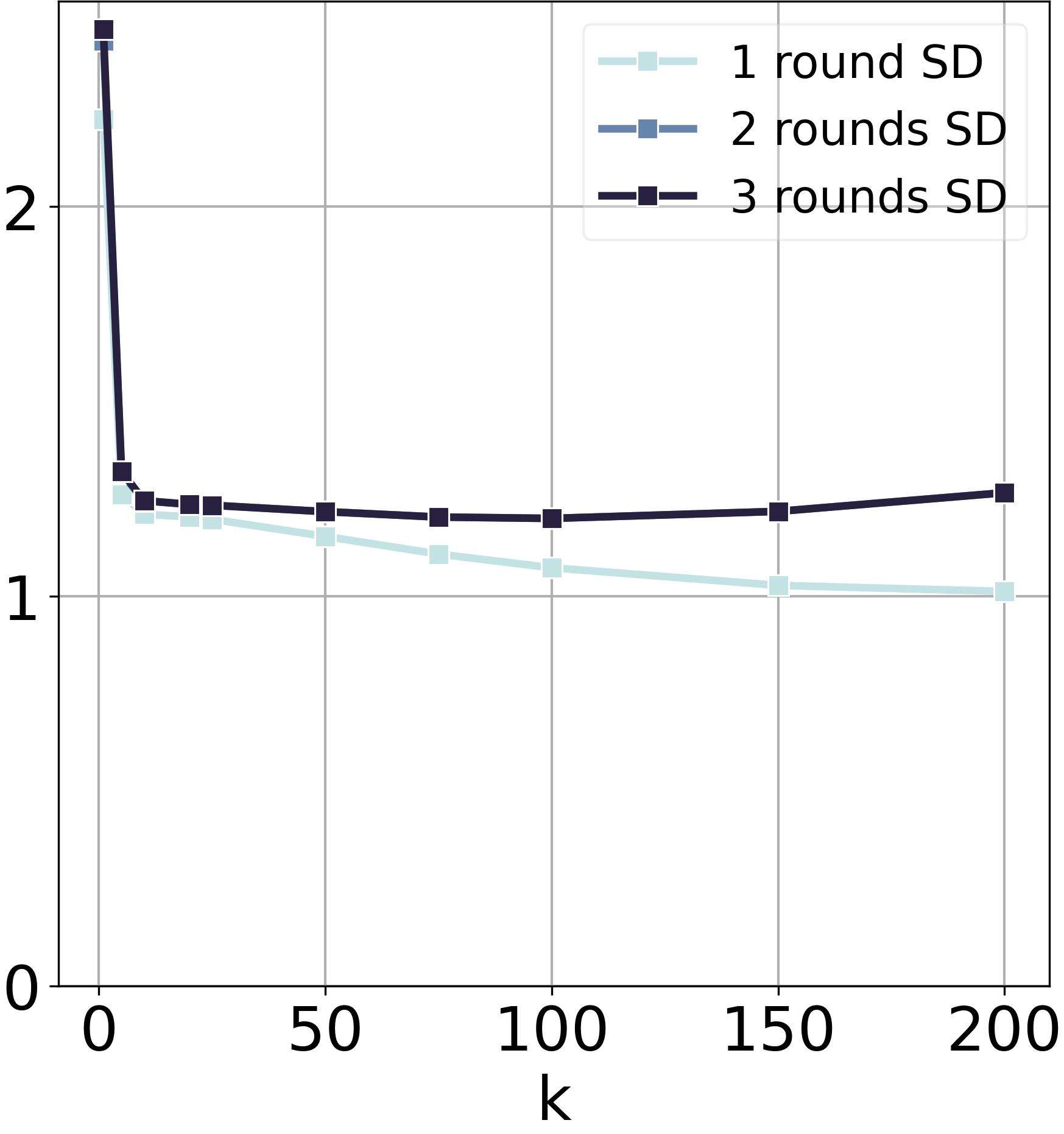}
        \caption{MBPP-S}
        \label{fig:ds_16b_sd_upperbound_mbpp_base}
    \end{subfigure}
    \begin{subfigure}{0.24\textwidth}
        \centering
        \includegraphics[width=\linewidth]{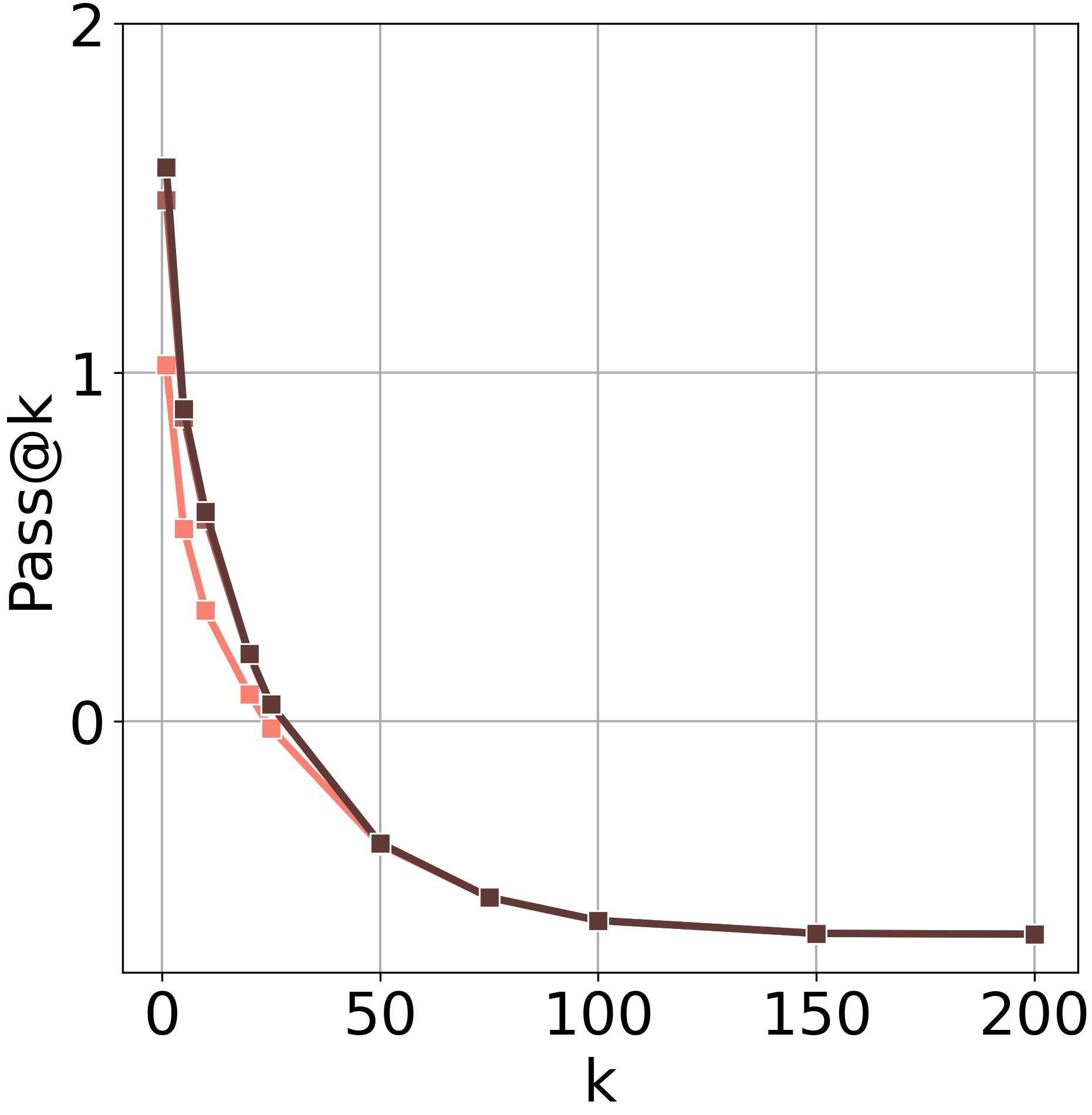}
        \caption{HumanEval+}
        \label{fig:ds_16b_sd_upperbound_humaneval_plus}
    \end{subfigure}\hfill
    \begin{subfigure}{0.226\textwidth}
        \centering
        \includegraphics[width=\linewidth]{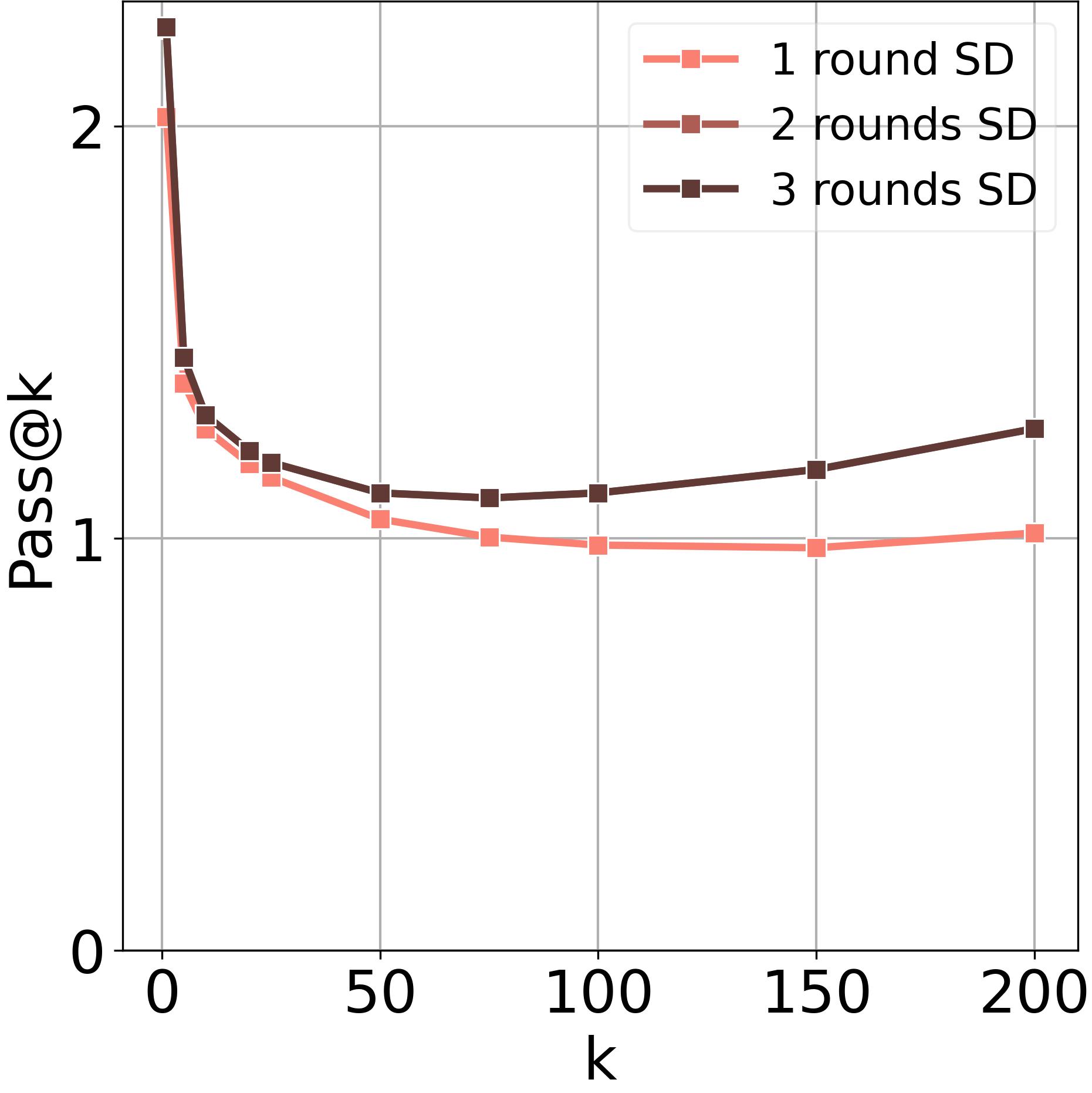}
        \caption{MBPP-S+}
        \label{fig:ds_16b_sd_upperbound_mbpp_plus}
    \end{subfigure}
    \caption
{Improvement in Pass@k of DeepSeekCoder-V2-Lite-Instruct after self-debugging compared to no self-debugging applied.}
    \label
{fig:ds_16b_sd_upperbound}
\end{figure*}

\begin{figure*}[tbp]
    \centering
    \begin{subfigure}{0.24\textwidth}
        \centering
        \includegraphics[width=\linewidth]{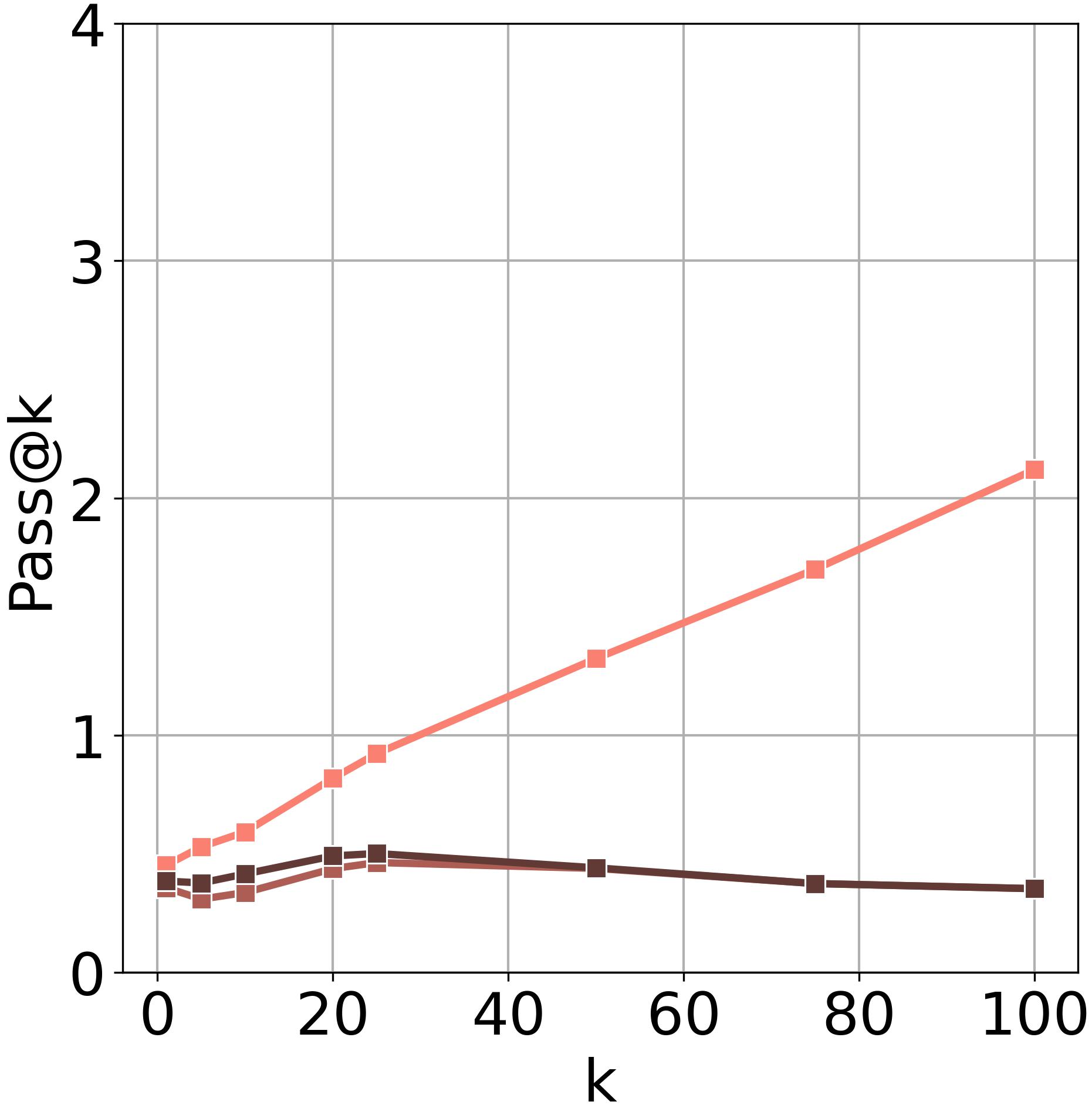}
        \caption{CL-7B}
        \label{fig:cl_7b_sd_upperbound_lcb_base}
    \end{subfigure}\hfill
    \begin{subfigure}{0.226\textwidth}
        \centering
        \includegraphics[width=\linewidth]{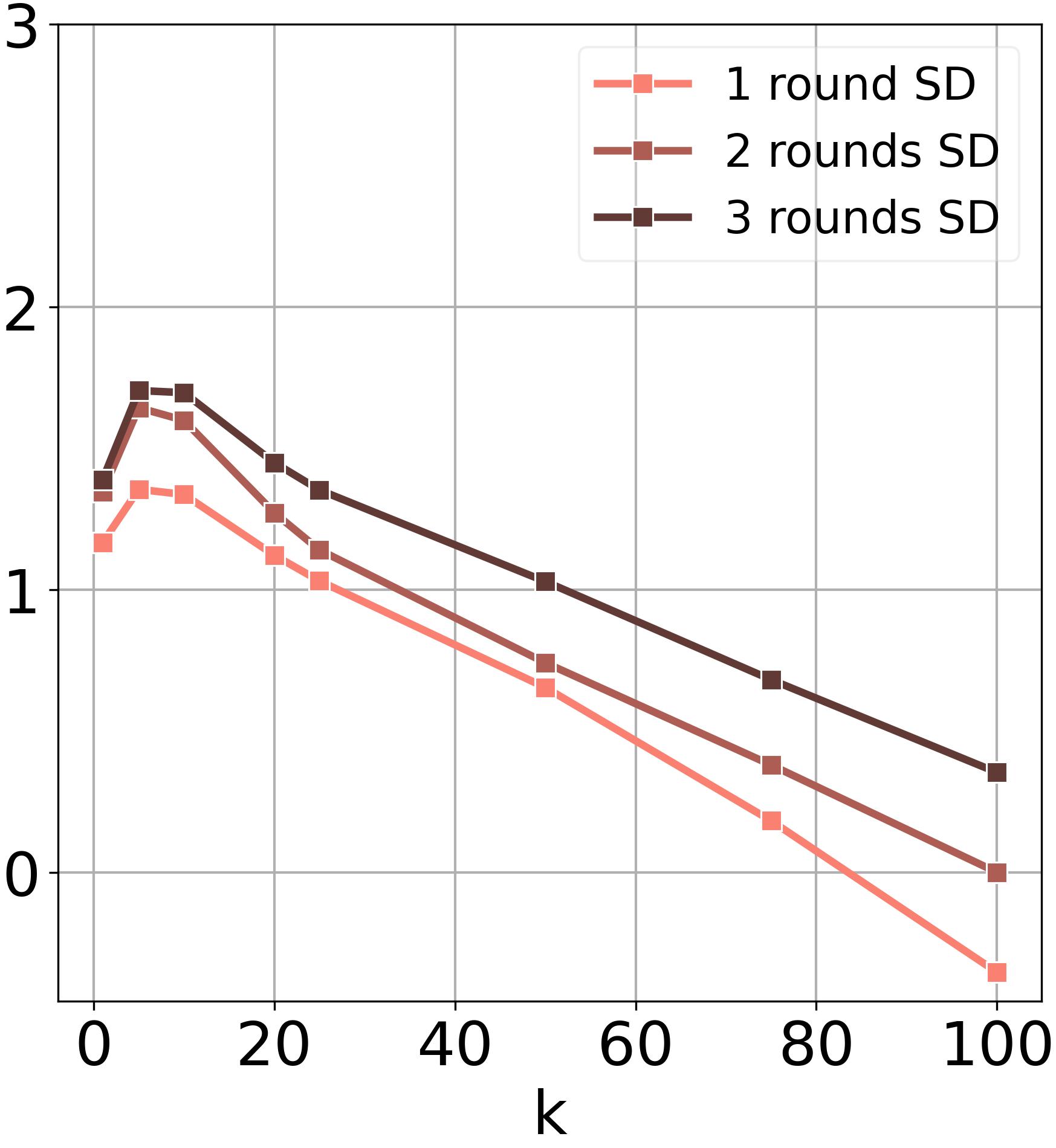}
        \caption{DS-6.7B}
        \label{fig:ds_6.7b_sd_upperbound_lcb_base}
    \end{subfigure}
    \begin{subfigure}{0.24\textwidth}
        \centering
        \includegraphics[width=\linewidth]{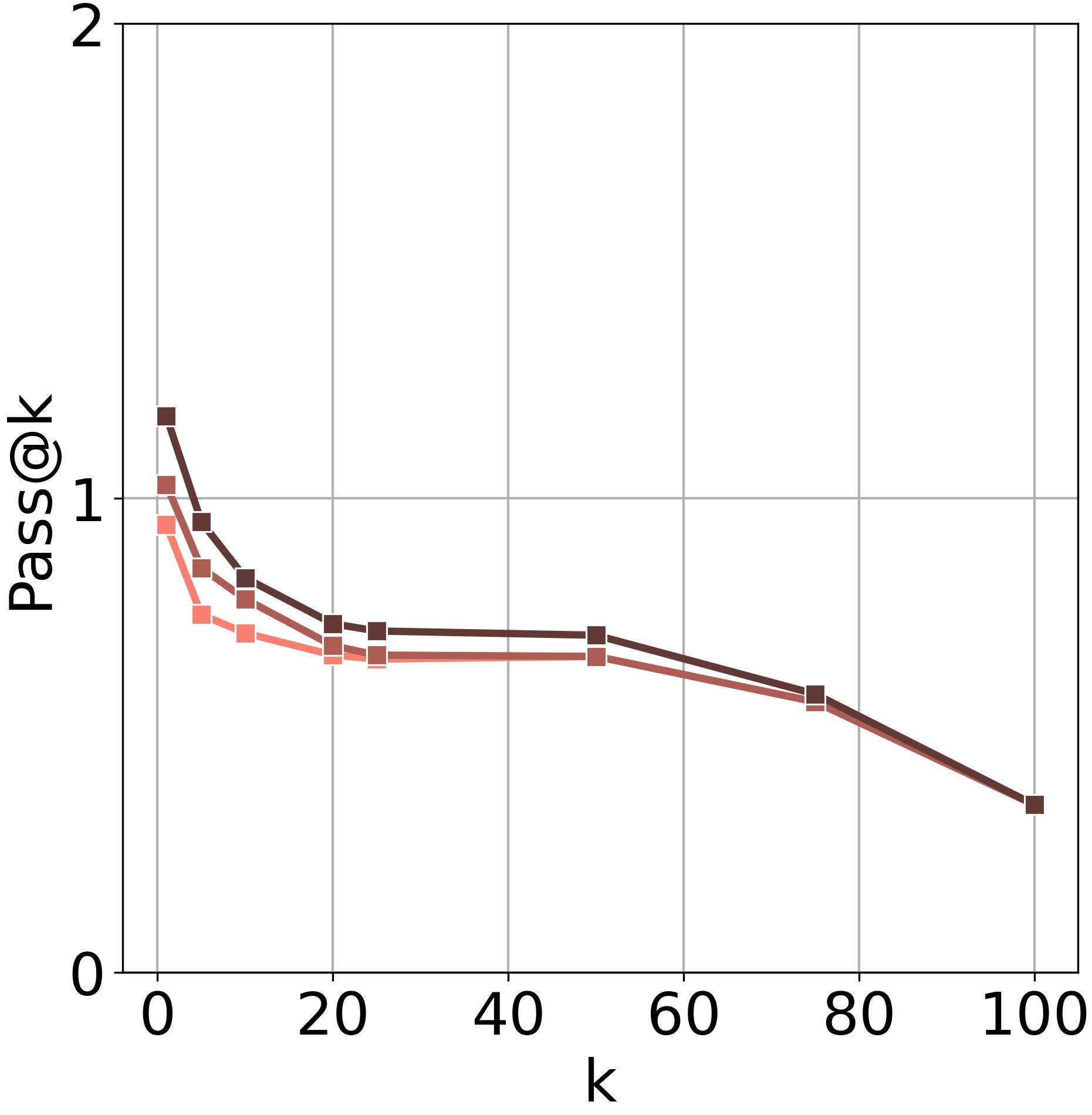}
        \caption{CL-13B}
        \label{fig:cl_13b_sd_upperbound_lcb_base}
    \end{subfigure}\hfill
    \begin{subfigure}{0.226\textwidth}
        \centering
        \includegraphics[width=\linewidth]{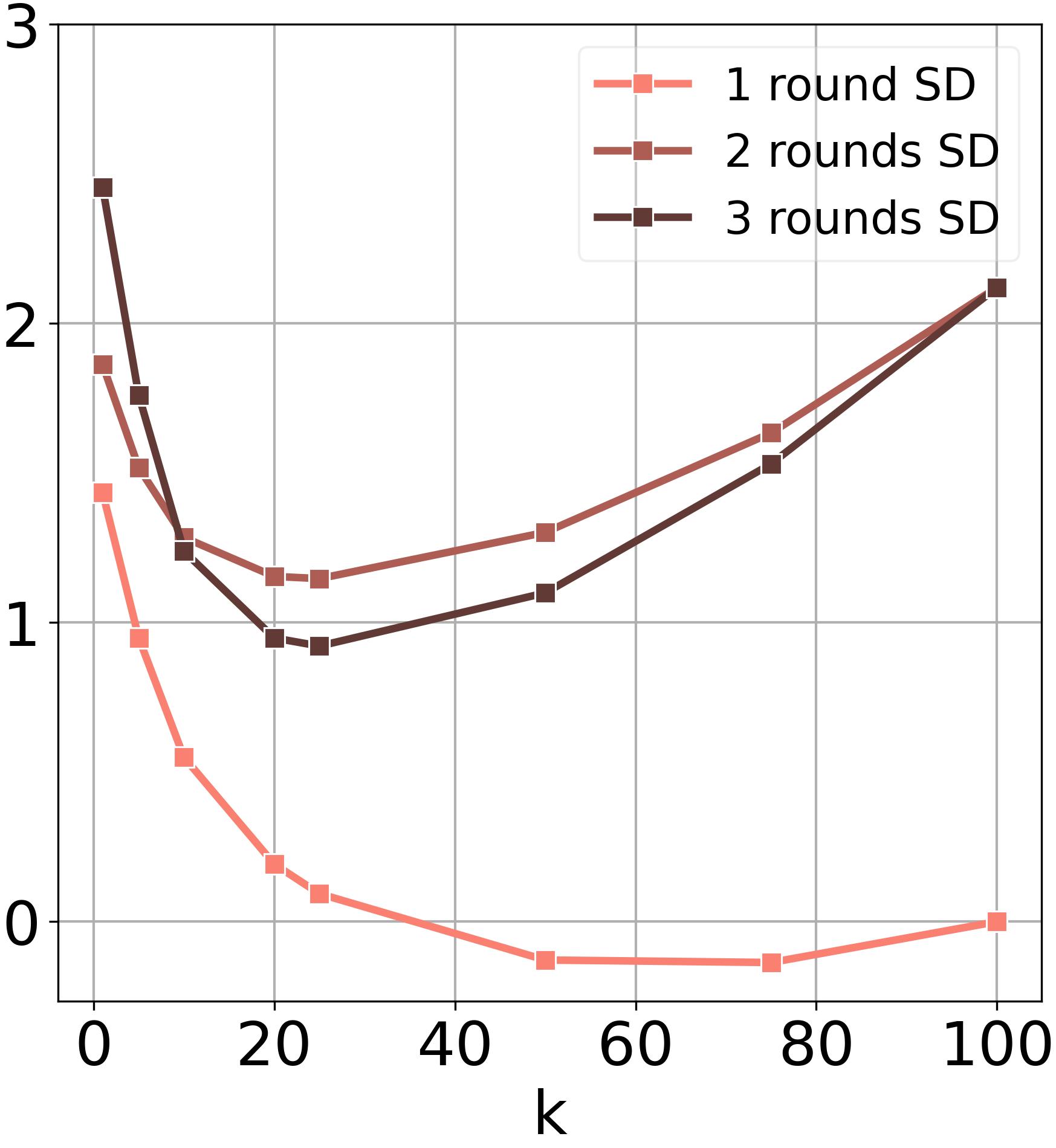}
        \caption{DS-V2-Lite}
        \label{fig:ds_16b_sd_upperbound_lcb_base}
    \end{subfigure}
    \caption
{Improvement in Pass@k on LiveCodeBench.}
    \label
{fig:lcb_sd_upperbound}
\end{figure*}

\begin{figure*}[tbp]
    \centering
    \begin{subfigure}{0.243\textwidth}
        \centering
        \includegraphics[width=\linewidth]{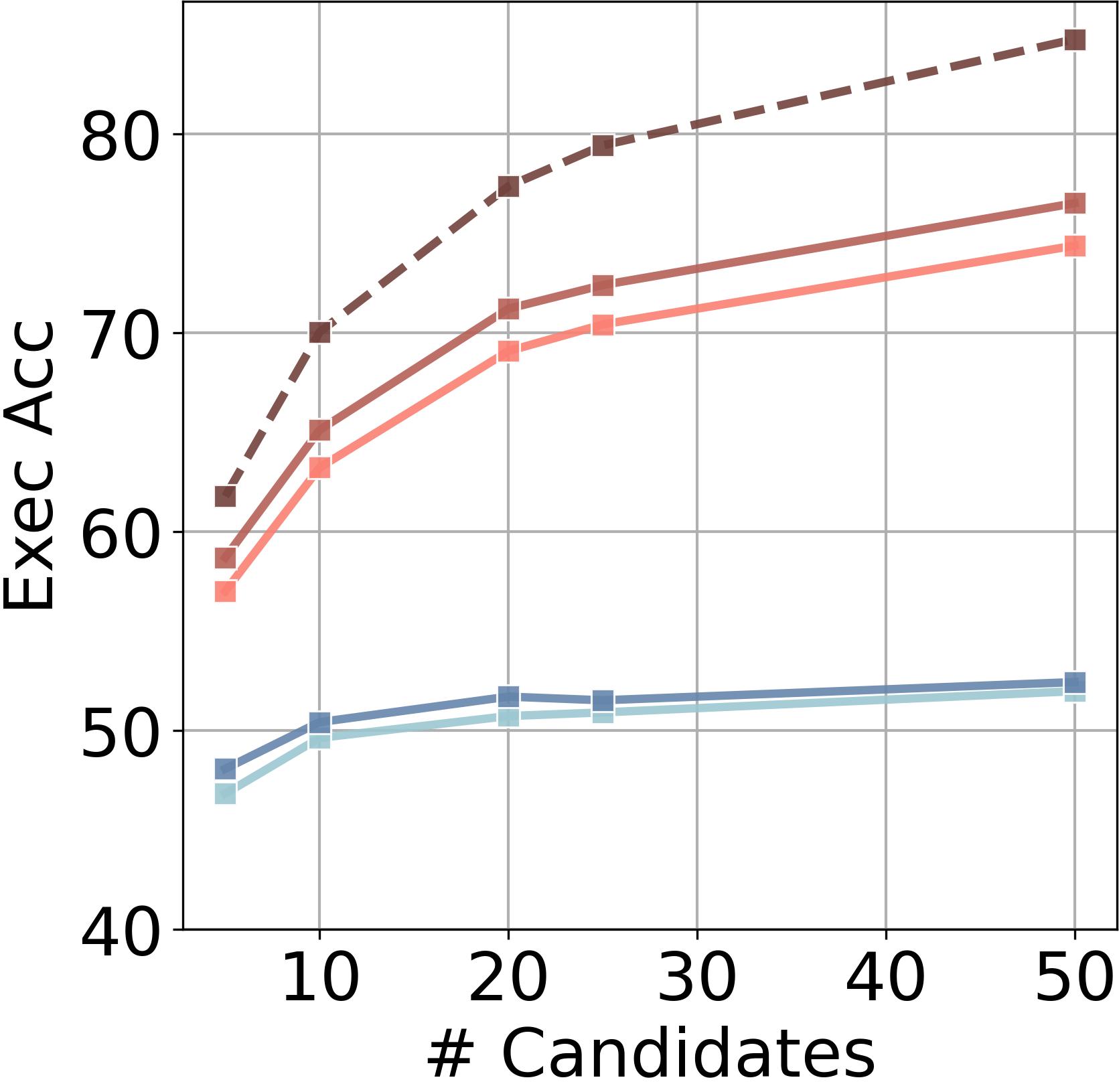}
        \caption{HumanEval}
        \label{fig:cl_7b_sd_candidates_humaneval_base}
    \end{subfigure}\hfill
    \begin{subfigure}{0.227\textwidth}
        \centering
        \includegraphics[width=\linewidth]{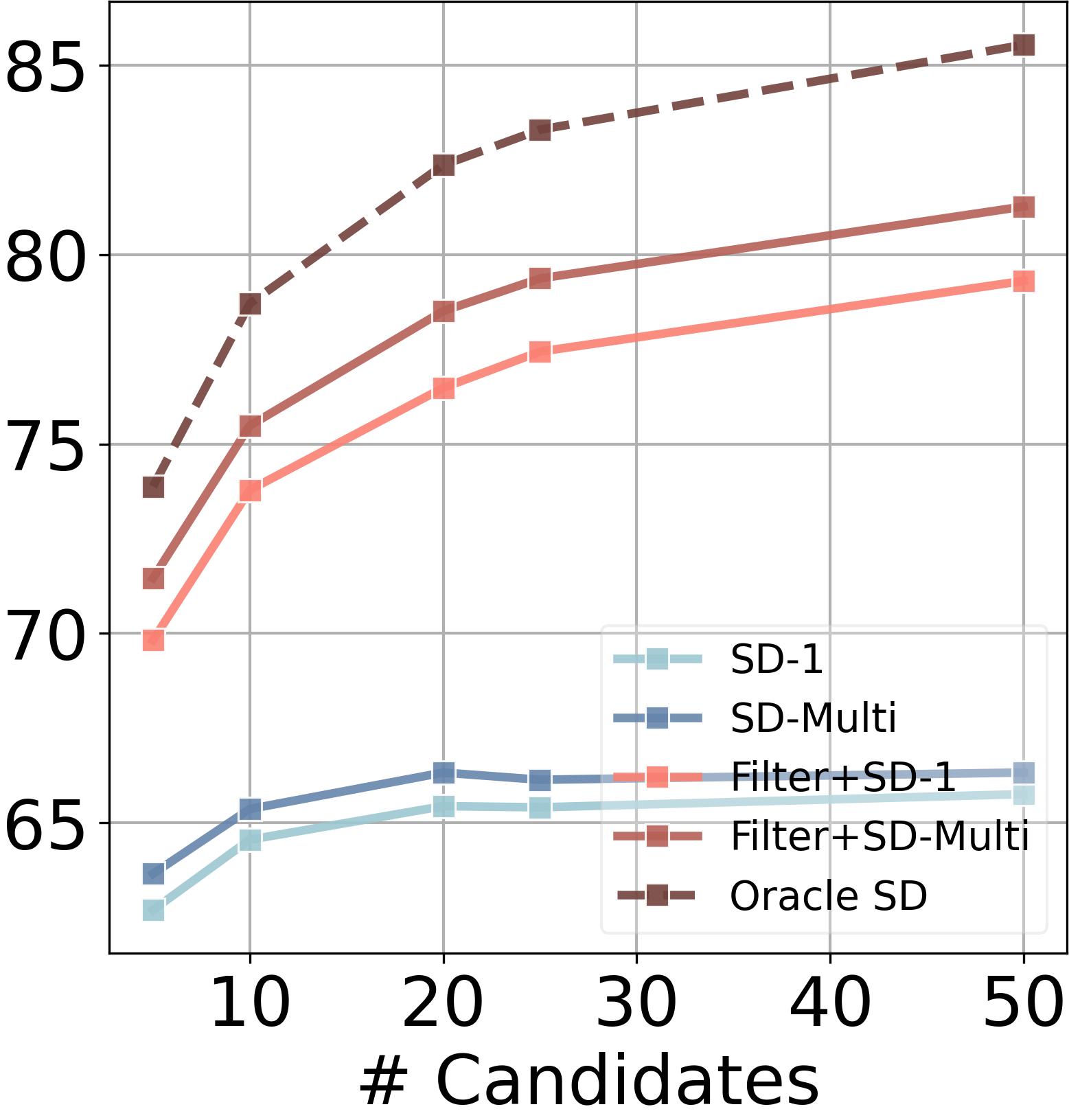}
        \caption{MBPP-S}
        \label{fig:cl_7b_sd_candidates_mbpp_base}
    \end{subfigure}
    \begin{subfigure}{0.243\textwidth}
        \centering
        \includegraphics[width=\linewidth]{figures/humaneval_sd_candidates.jpg}
        \caption{HumanEval+}
        \label{fig:cl_7b_sd_candidates_humaneval_plus}
    \end{subfigure}\hfill
    \begin{subfigure}{0.227\textwidth}
        \centering
        \includegraphics[width=\linewidth]{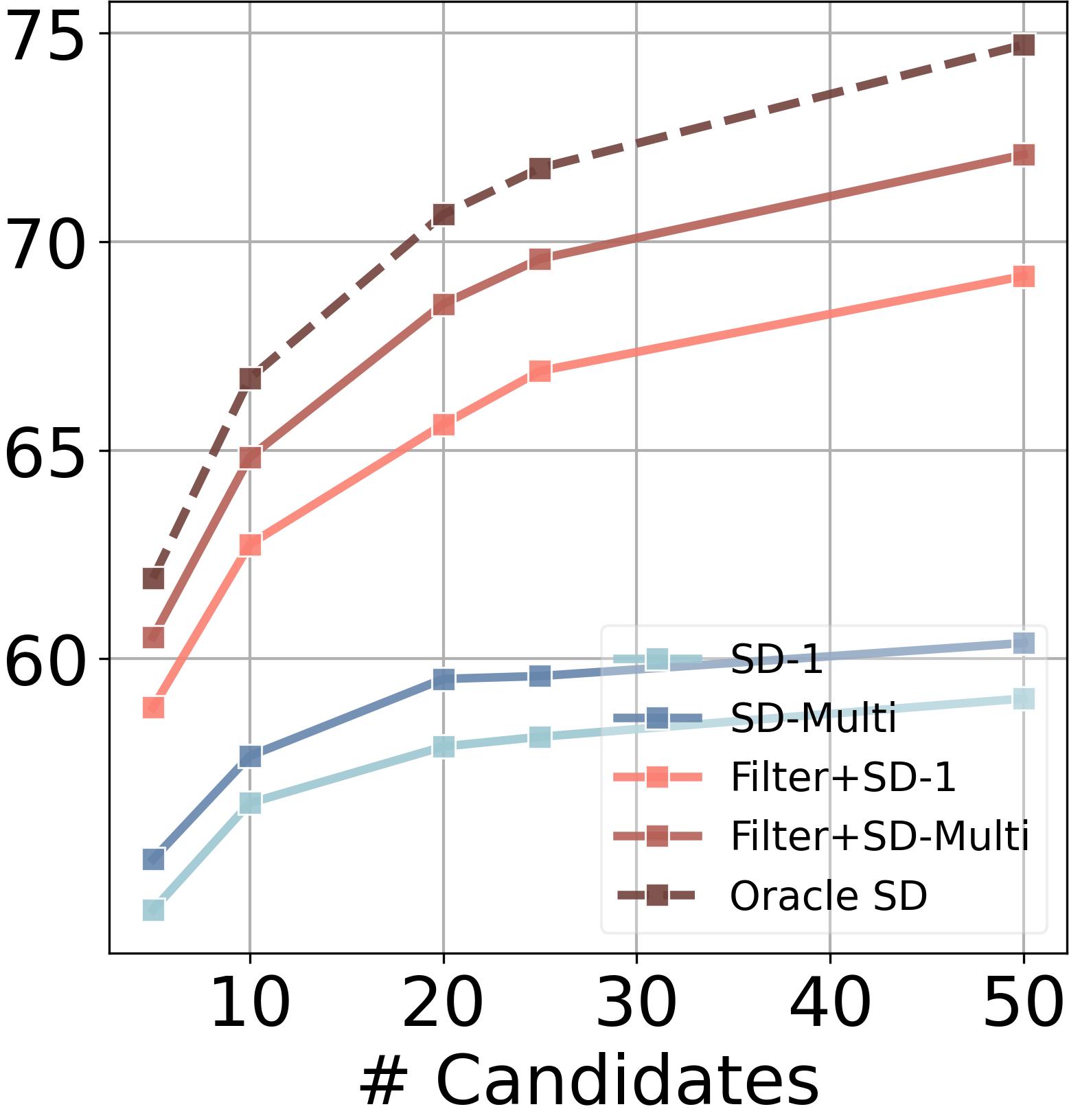}
        \caption{MBPP-S+}
        \label{fig:cl_7b_sd_candidates_mbpp_plus}
    \end{subfigure}
    \caption{Comparison of self-debugging methods over different numbers of candidates generated by CodeLlama-7B-Instruct, and debugged over $\{1,\textbf{Multi}\}$ candidates using the same LLM. We also provide the oracle after self-debugging for one round. Note that results ending with $+$ mean it is evaluated on the plus with extended test cases. Results are averaged across at least 4 runs.}
    \label{fig:cl_7b_sd_candidates}
\end{figure*}

\begin{figure*}[tbp]
    \centering
    \begin{subfigure}{0.243\textwidth}
        \centering
        \includegraphics[width=\linewidth]{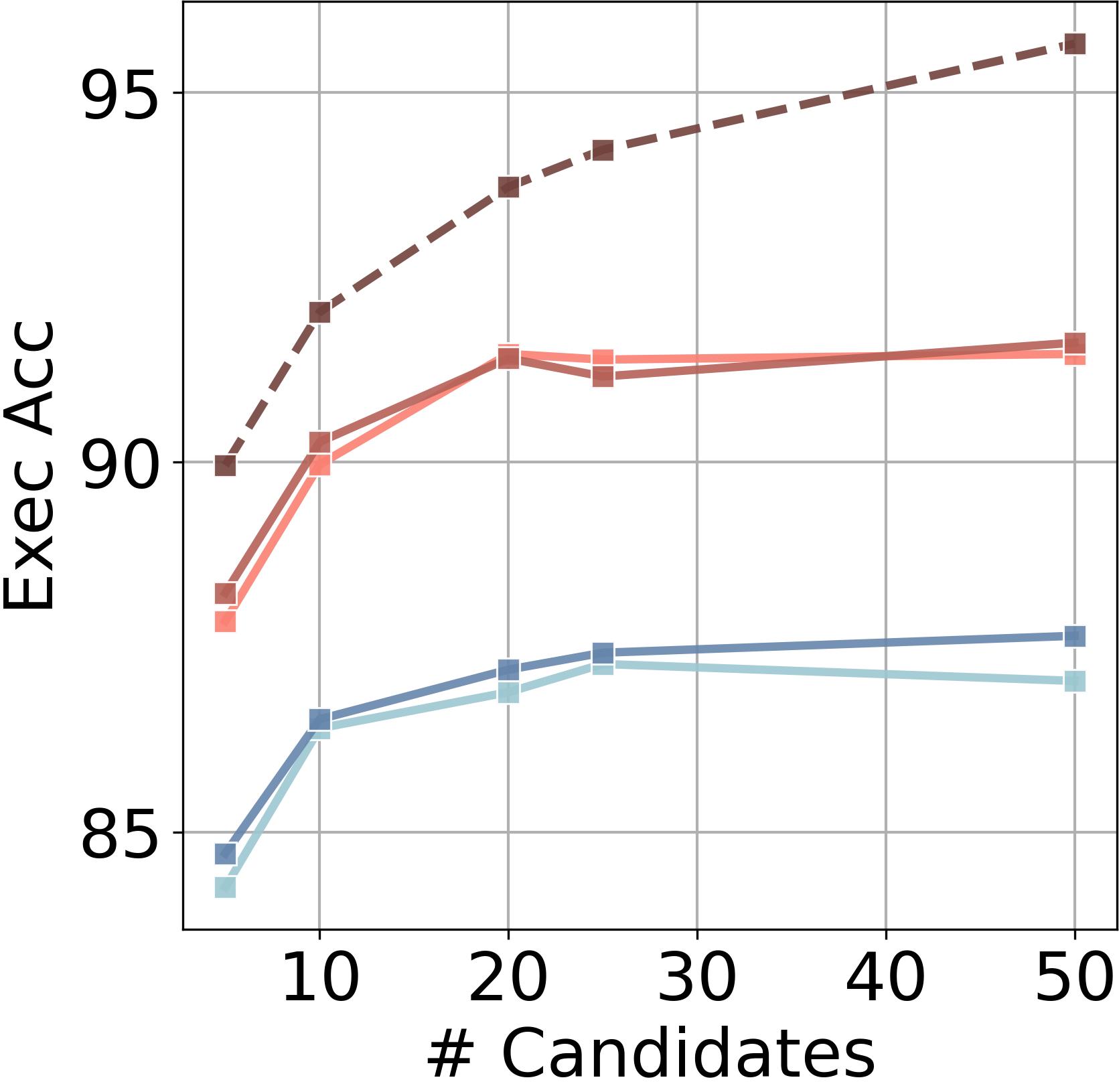}
        \caption{HumanEval}
        \label{fig:ds_6.7b_sd_candidates_humaneval_base}
    \end{subfigure}\hfill
    \begin{subfigure}{0.227\textwidth}
        \centering
        \includegraphics[width=\linewidth]{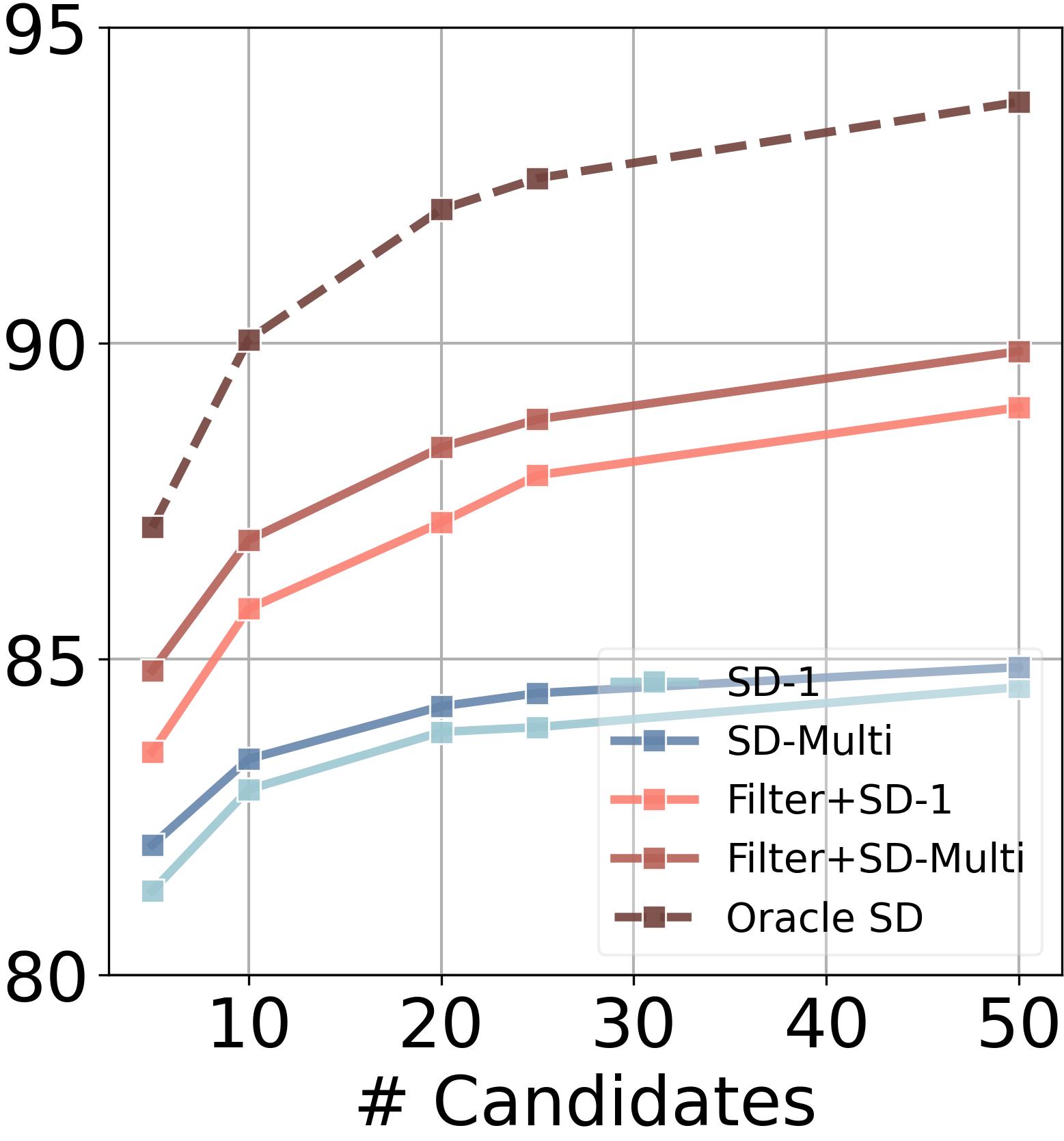}
        \caption{MBPP-S}
        \label{fig:ds_6.7b_sd_candidates_mbpp_base}
    \end{subfigure}
    \begin{subfigure}{0.243\textwidth}
        \centering
        \includegraphics[width=\linewidth]{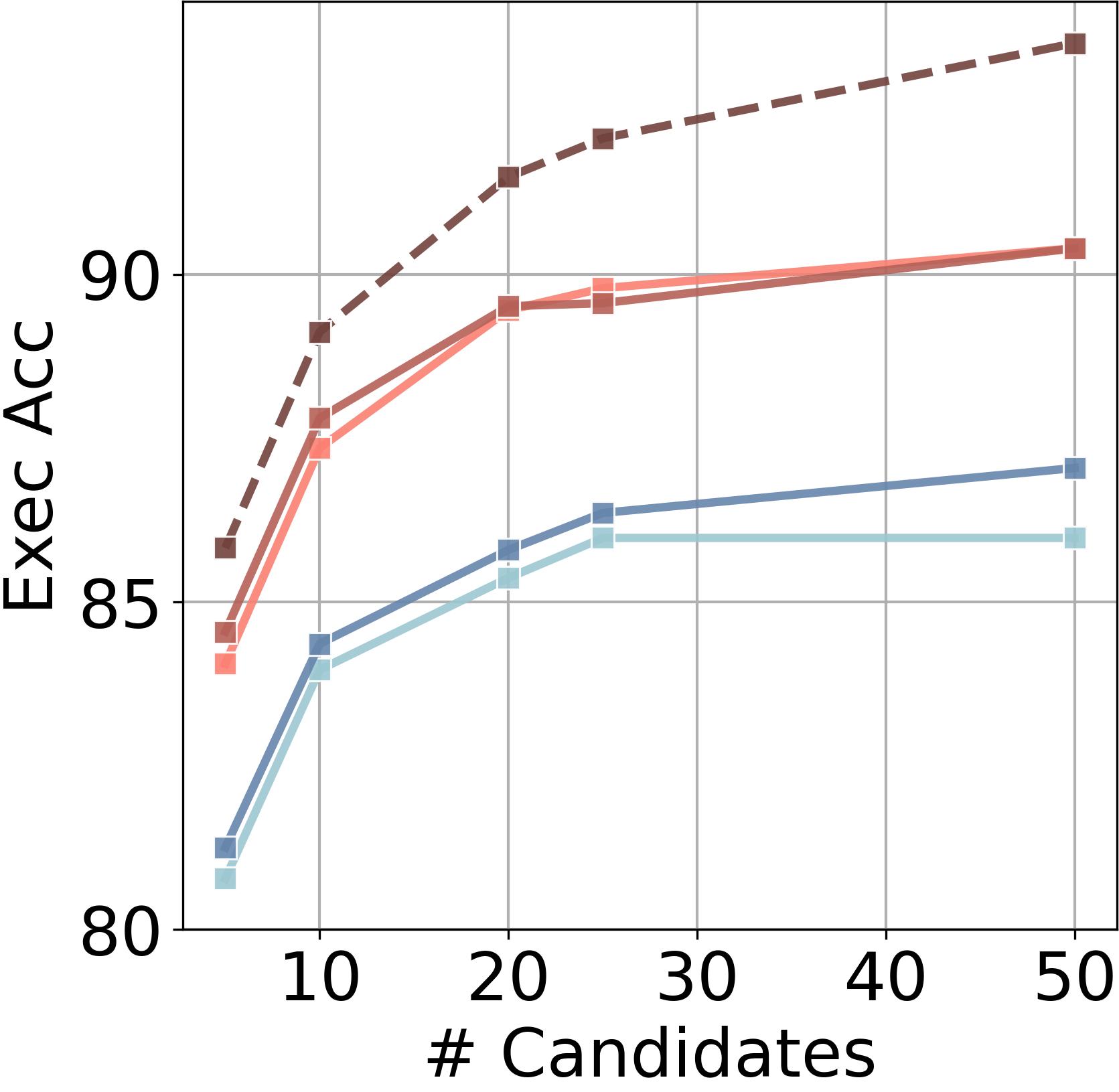}
        \caption{HumanEval+}
        \label{fig:ds_6.7b_sd_candidates_humaneval_plus}
    \end{subfigure}\hfill
    \begin{subfigure}{0.227\textwidth}
        \centering
        \includegraphics[width=\linewidth]{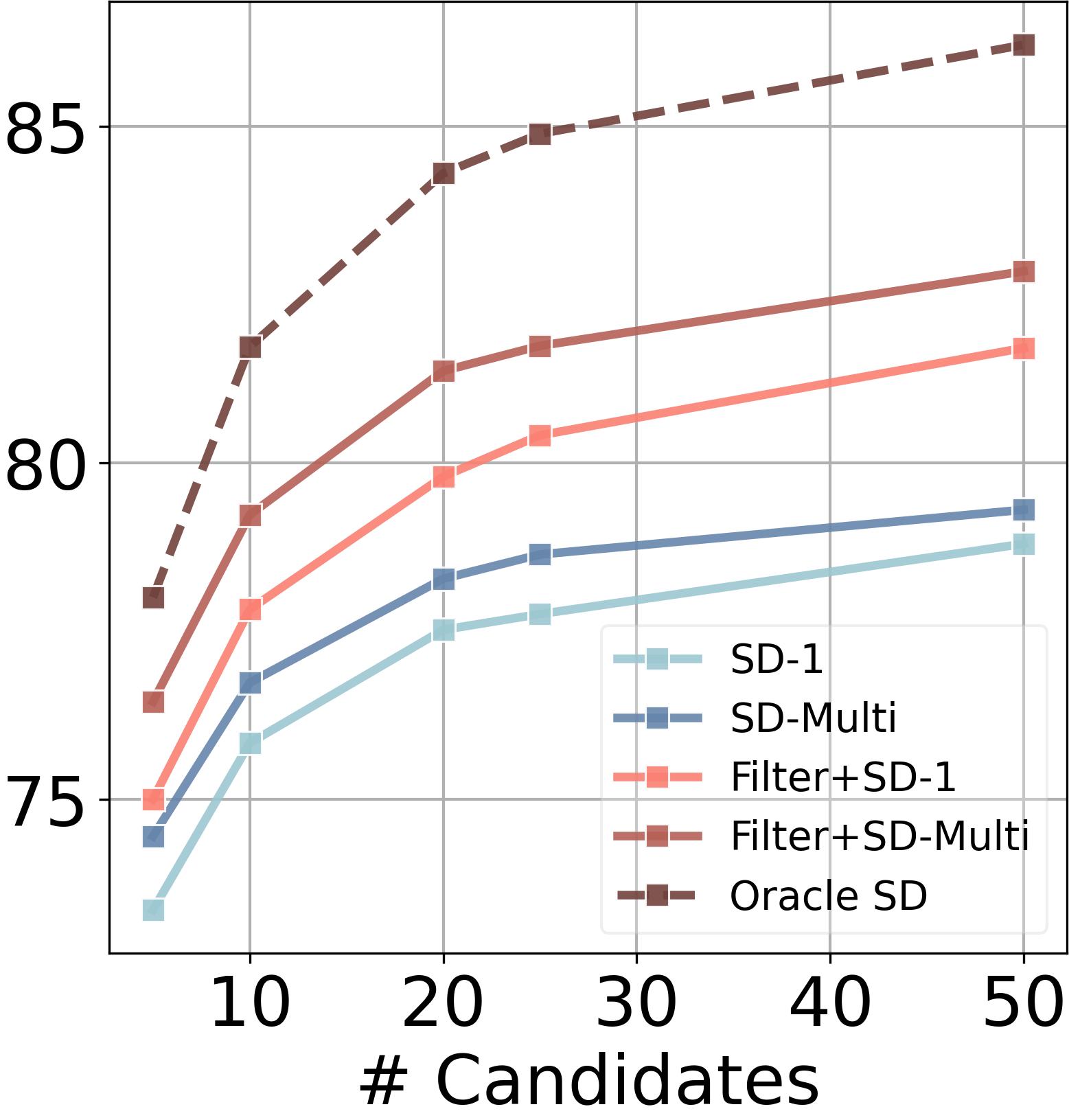}
        \caption{MBPP-S+}
        \label{fig:ds_6.7b_sd_candidates_mbpp_plus}
    \end{subfigure}
    \caption{Comparison of self-debugging methods over different numbers of candidates generated by DeepSeekCoder-6.7B-Instruct, and debugged over $\{1,\textbf{Multi}\}$ candidates using the same LLM. We also provide the oracle after self-debugging for one round. Note that results ending with $+$ mean it is evaluated on the plus with extended test cases. Results are averaged across at least 4 runs.}
    \label{fig:ds_6.7b_sd_candidates}
\end{figure*}

\begin{figure*}[tbp]
    \centering
    \begin{subfigure}{0.243\textwidth}
        \centering
        \includegraphics[width=\linewidth]{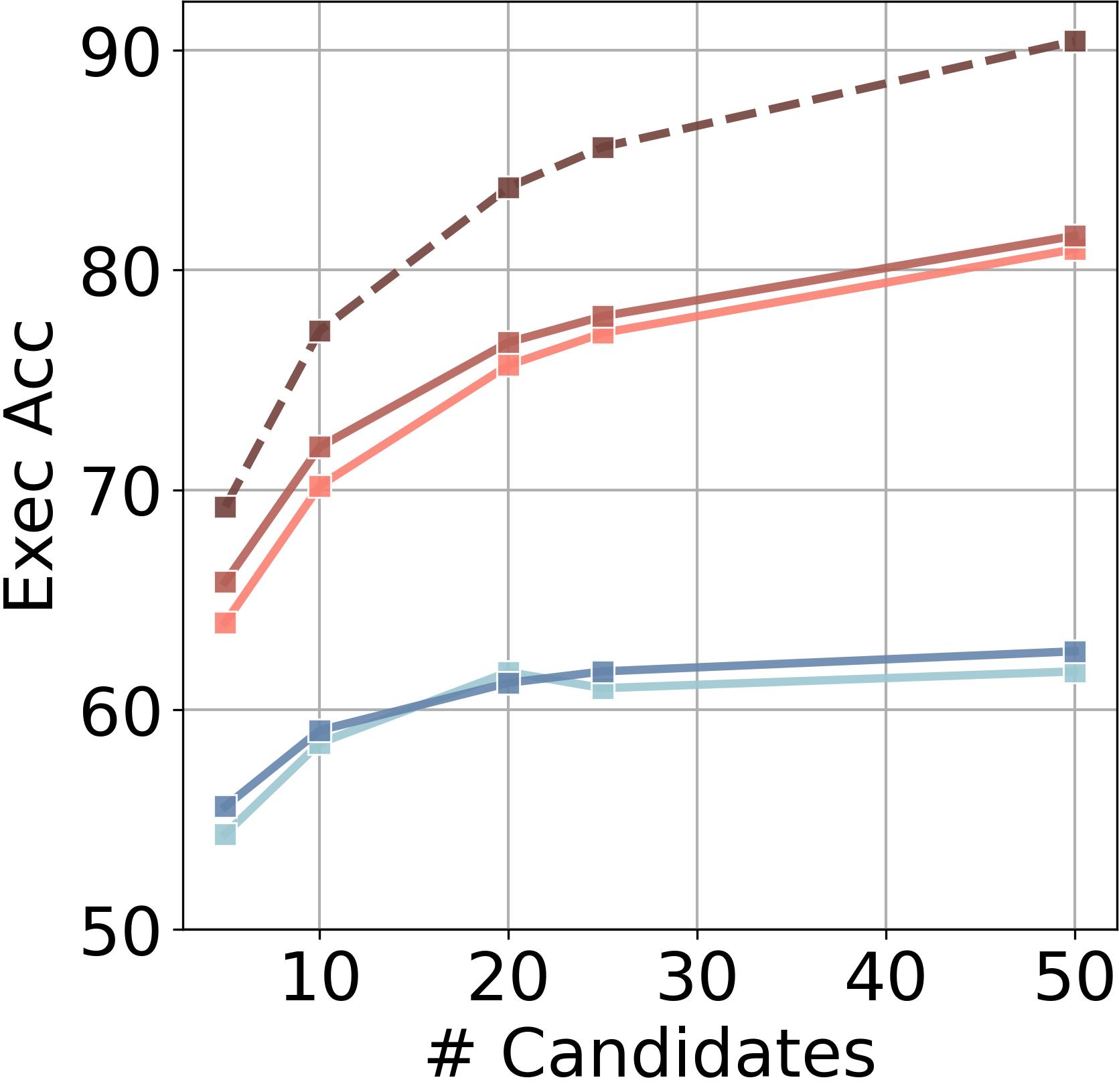}
        \caption{HumanEval}
        \label{fig:cl_13b_sd_candidates_humaneval_base}
    \end{subfigure}\hfill
    \begin{subfigure}{0.227\textwidth}
        \centering
        \includegraphics[width=\linewidth]{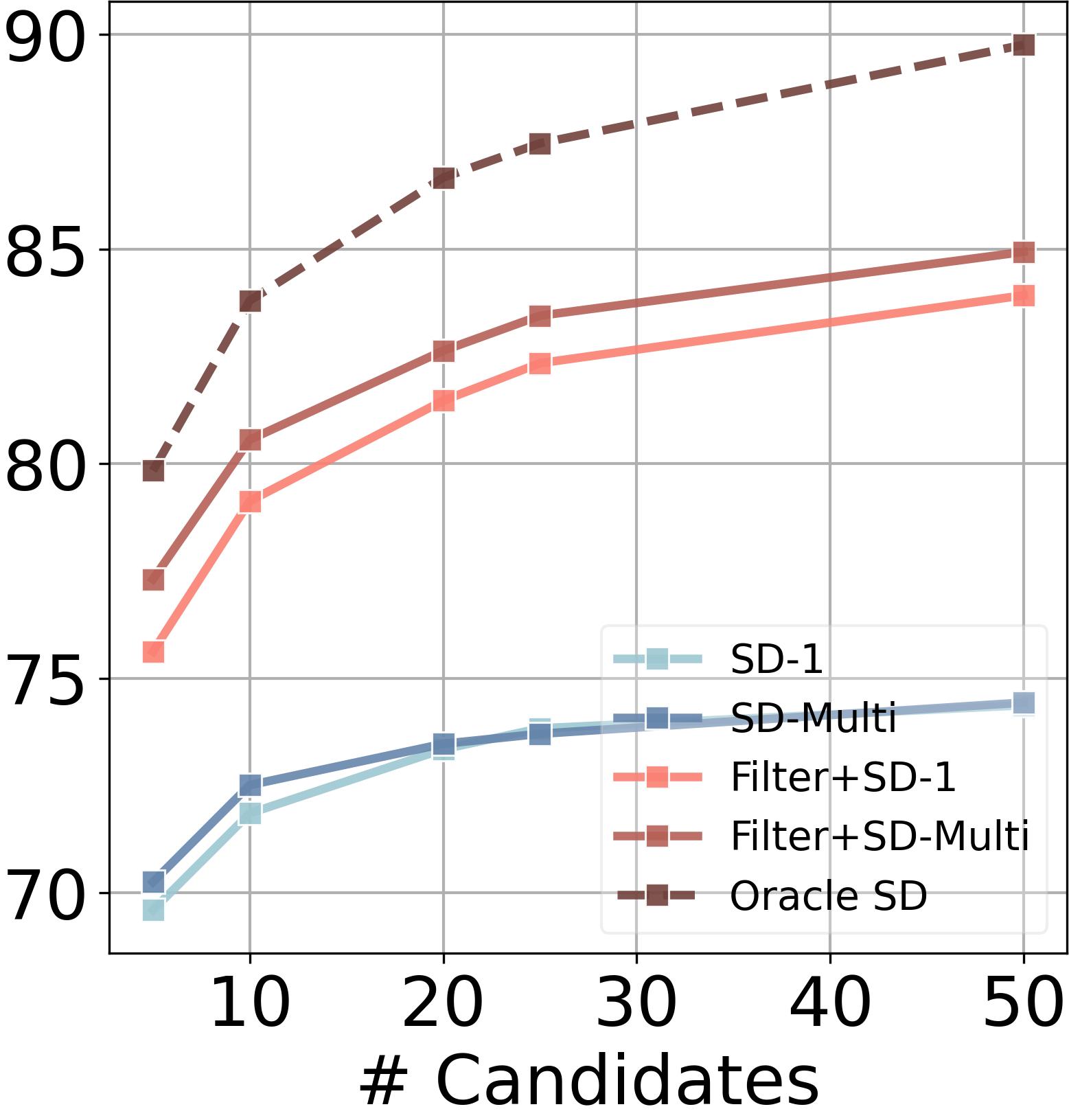}
        \caption{MBPP-S}
        \label{fig:cl_13b_sd_candidates_mbpp_base}
    \end{subfigure}
    \begin{subfigure}{0.243\textwidth}
        \centering
        \includegraphics[width=\linewidth]{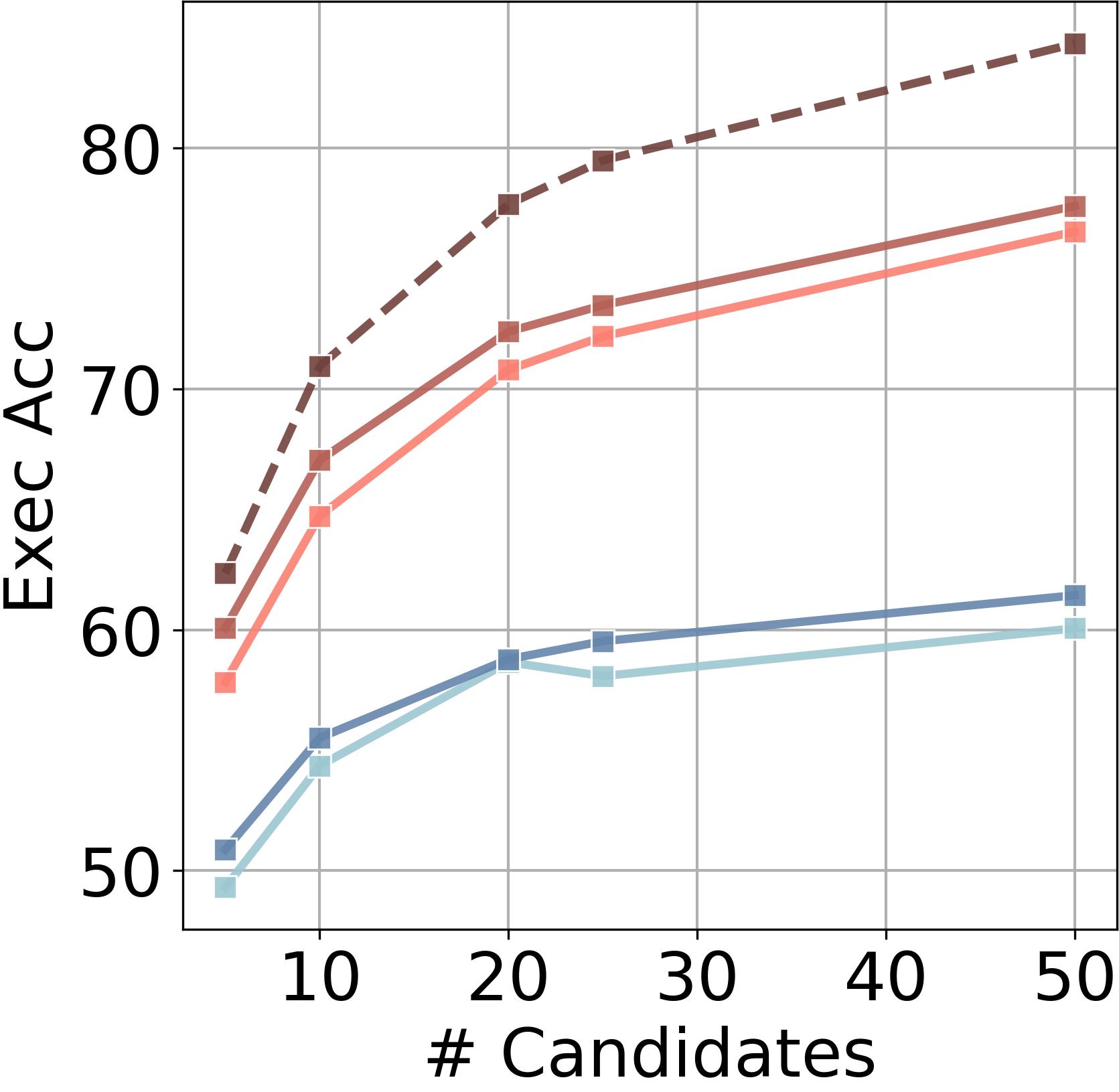}
        \caption{HumanEval+}
        \label{fig:cl_13b_sd_candidates_humaneval_plus}
    \end{subfigure}\hfill
    \begin{subfigure}{0.227\textwidth}
        \centering
        \includegraphics[width=\linewidth]{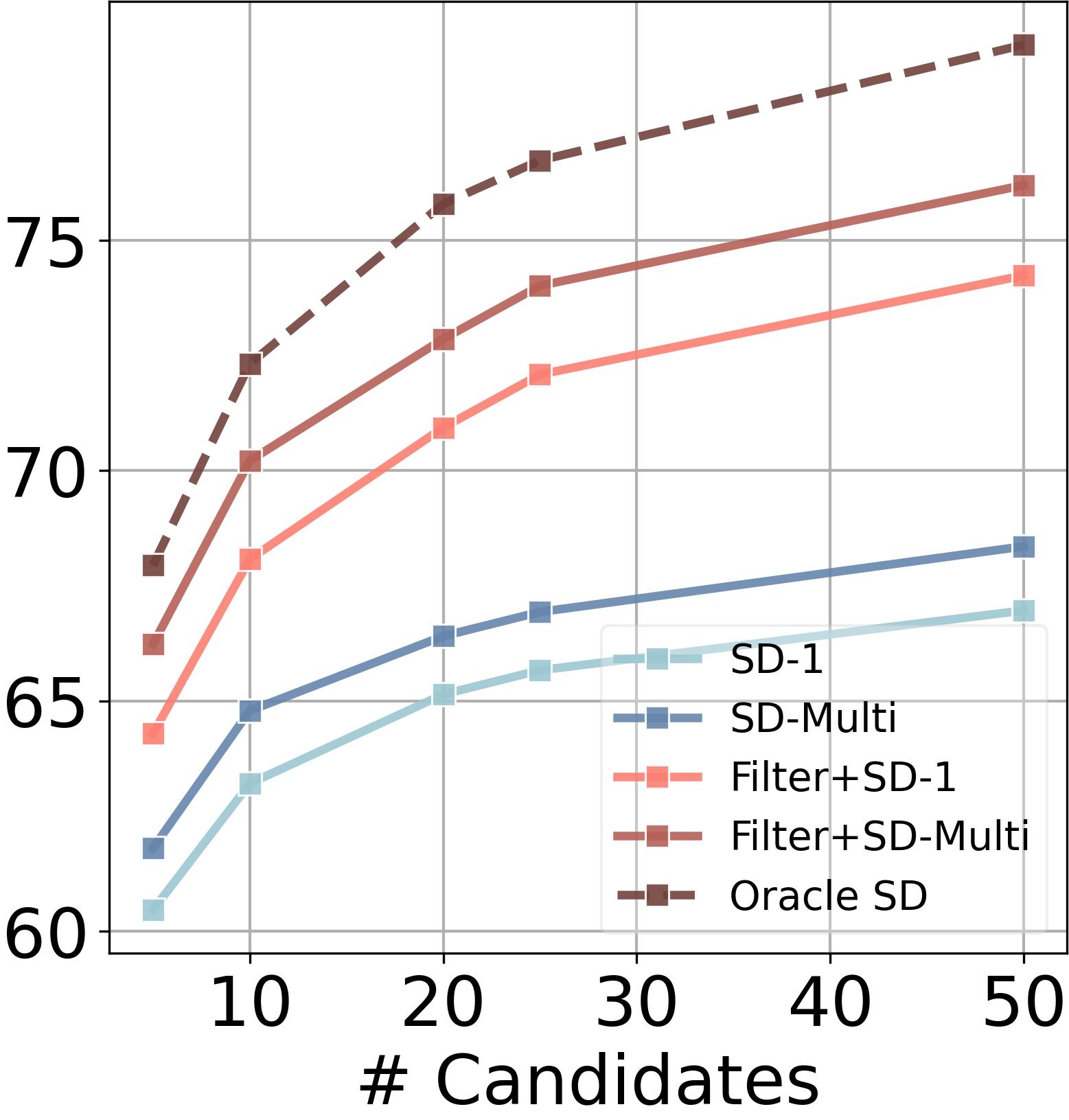}
        \caption{MBPP-S+}
        \label{fig:cl_13b_sd_candidates_mbpp_plus}
    \end{subfigure}
    \caption{Comparison of self-debugging methods over different numbers of candidates generated by CodeLlama-13B-Instruct, and debugged over $\{1,\textbf{Multi}\}$ candidates using the same LLM. We also provide the oracle after self-debugging for one round. Note that results ending with $+$ mean it is evaluated on the plus with extended test cases. Results are averaged across at least 4 runs.}
    \label{fig:cl_13b_sd_candidates}
\end{figure*}

\begin{figure*}[tbp]
    \centering
    \begin{subfigure}{0.243\textwidth}
        \centering
        \includegraphics[width=\linewidth]{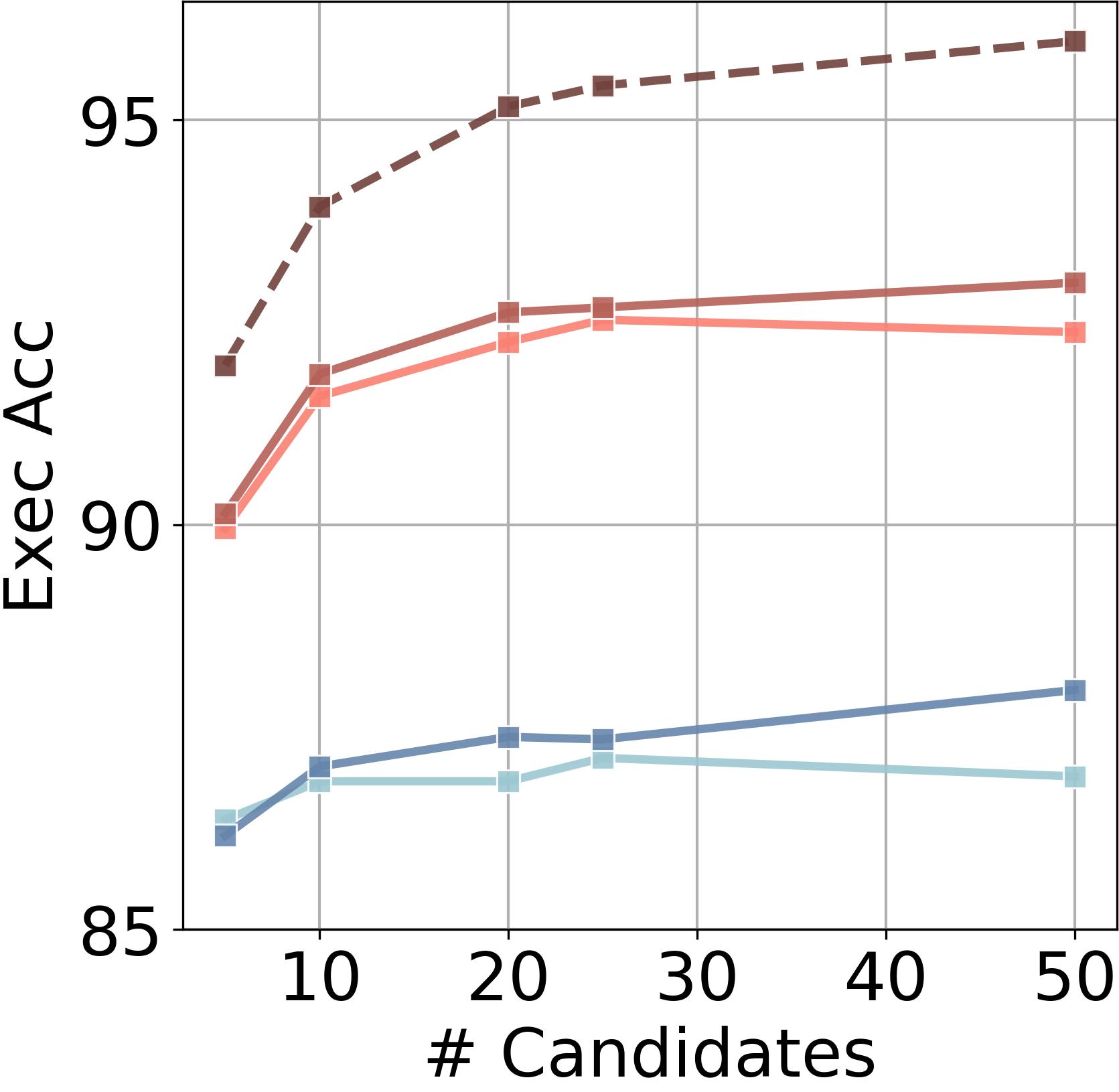}
        \caption{HumanEval}
        \label{fig:ds_16b_sd_candidates_humaneval_base}
    \end{subfigure}\hfill
    \begin{subfigure}{0.227\textwidth}
        \centering
        \includegraphics[width=\linewidth]{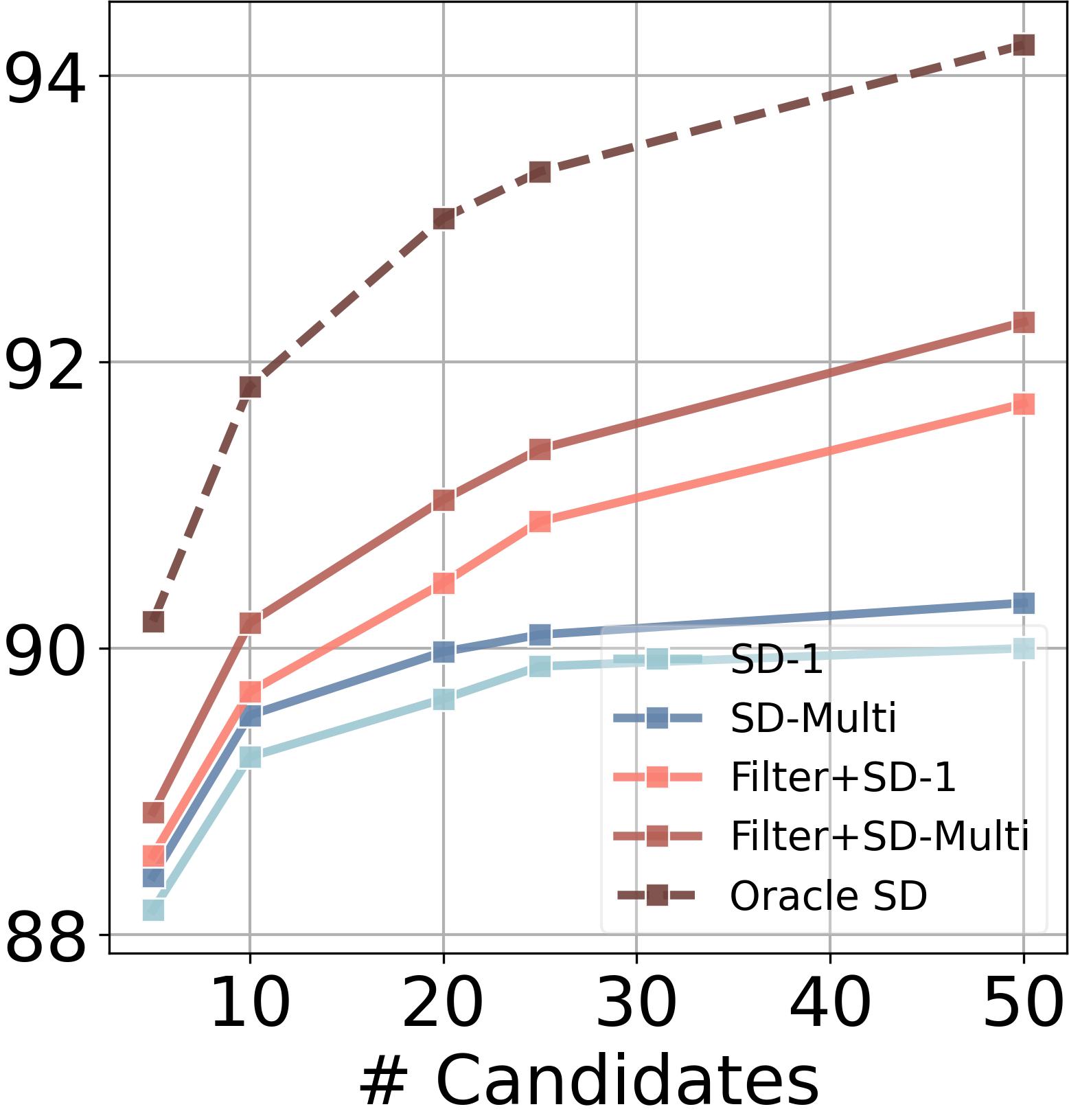}
        \caption{MBPP-S}
        \label{fig:ds_16b_sd_candidates_mbpp_base}
    \end{subfigure}
    \begin{subfigure}{0.243\textwidth}
        \centering
        \includegraphics[width=\linewidth]{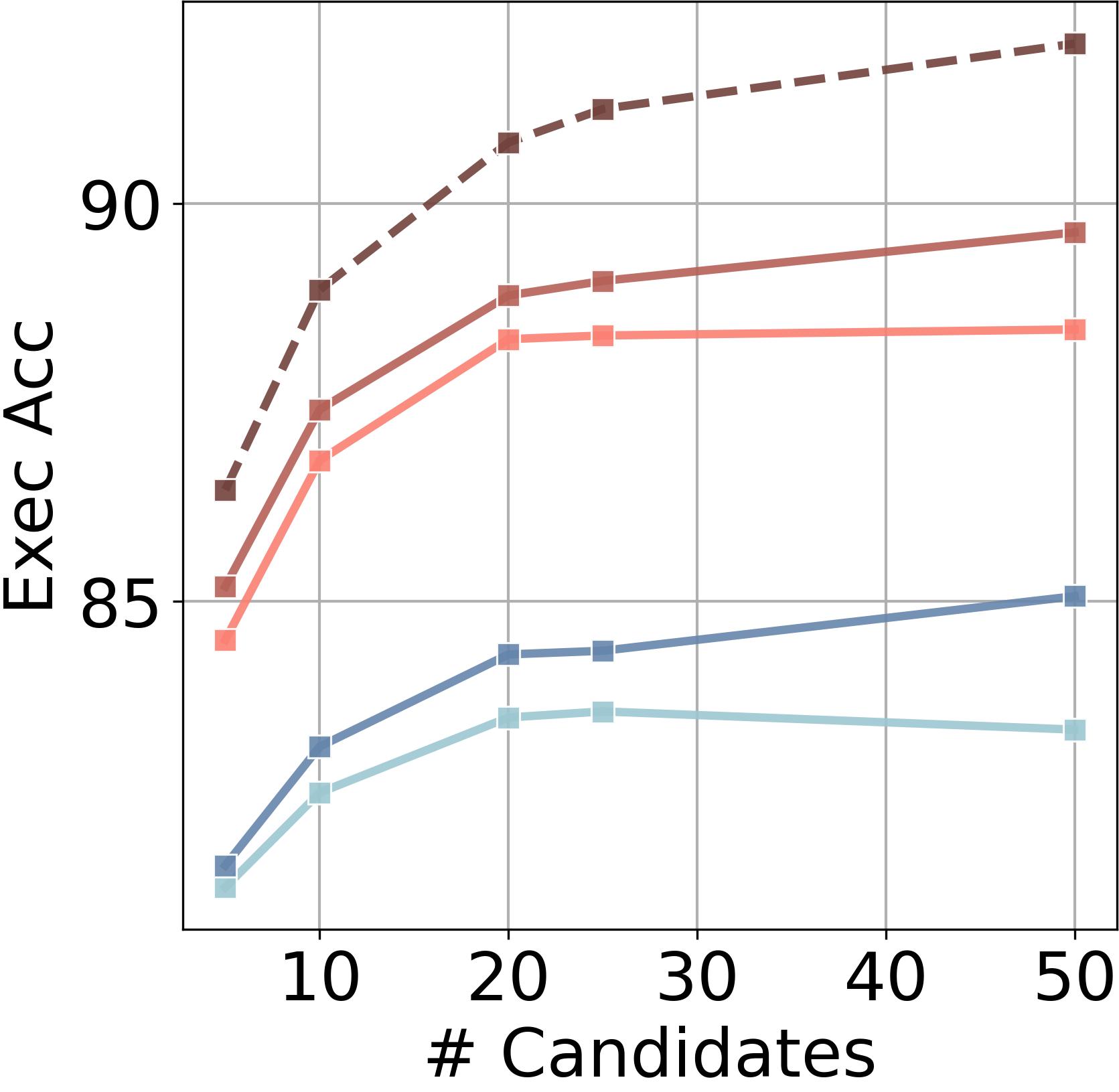}
        \caption{HumanEval+}
        \label{fig:ds_16b_sd_candidates_humaneval_plus}
    \end{subfigure}\hfill
    \begin{subfigure}{0.227\textwidth}
        \centering
        \includegraphics[width=\linewidth]{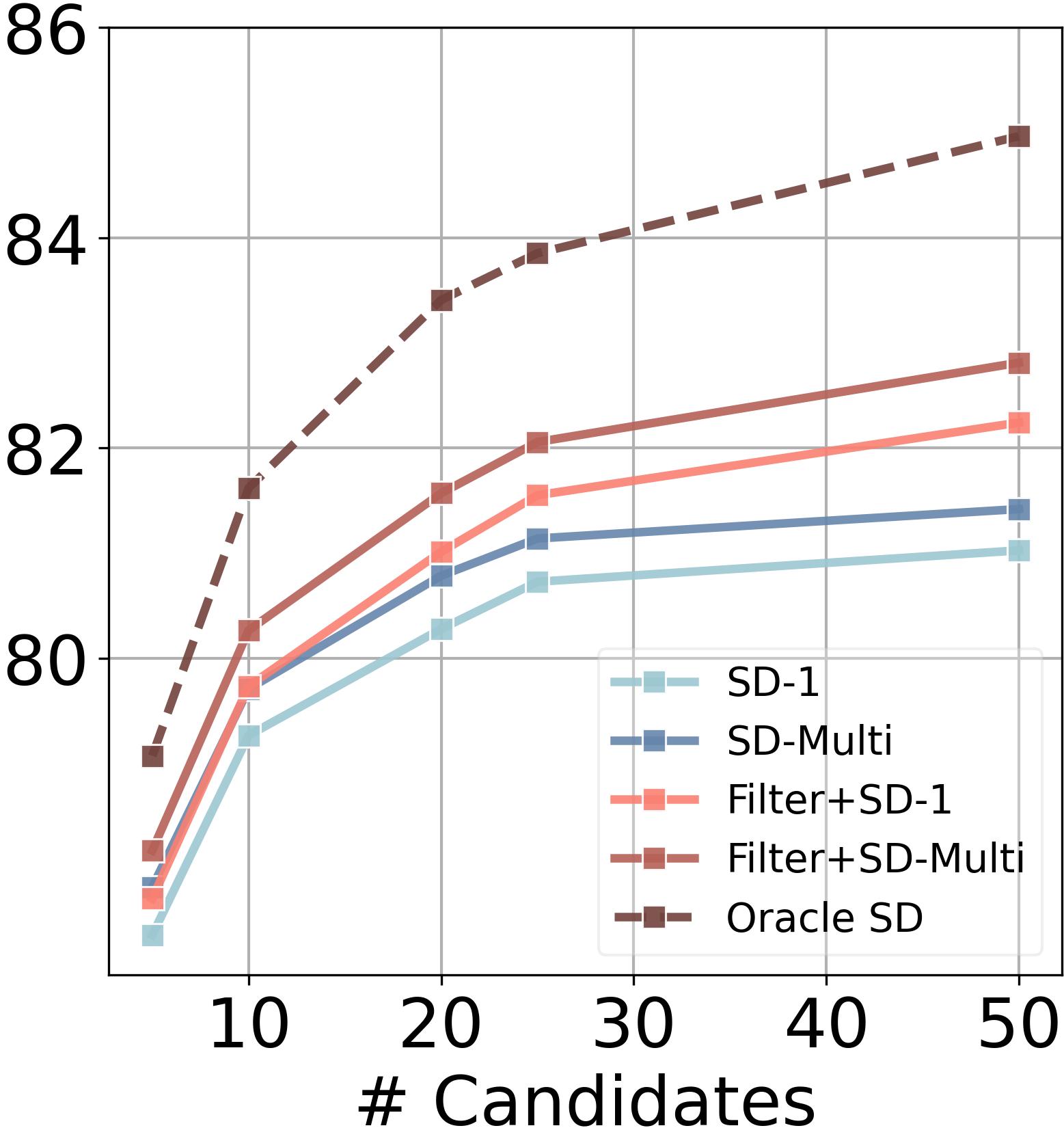}
        \caption{MBPP-S+}
        \label{fig:ds_16b_sd_candidates_mbpp_plus}
    \end{subfigure}
    \caption{Comparison of self-debugging methods over different numbers of candidates generated by DeepSeekCoder-V2-Lite-Instruct, and debugged over $\{1,\textbf{Multi}\}$ candidates using the same LLM. We also provide the oracle after self-debugging for one round. Note that results ending with $+$ mean it is evaluated on the plus with extended test cases. Results are averaged across at least 4 runs.}
    \label{fig:ds_16b_sd_candidates}
\end{figure*}

\begin{figure*}[htbp]
    \centering
    \begin{subfigure}{0.237\textwidth}
        \centering
        \includegraphics[width=\linewidth]{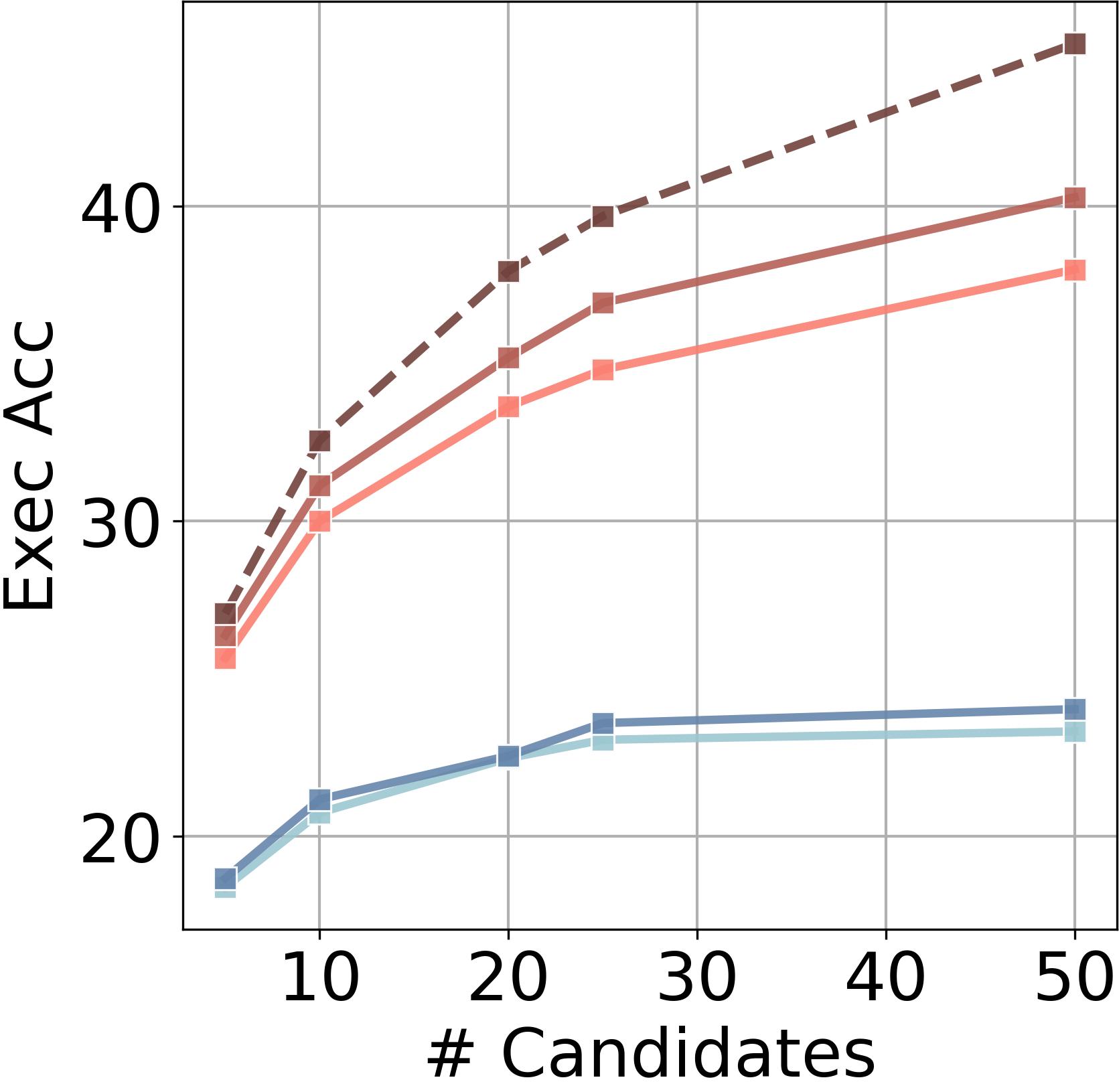}
        \caption{CL-7B}
        \label{fig:lcb_sd_candidates_original}
    \end{subfigure}\hfill
    \begin{subfigure}{0.22\textwidth}
        \centering
        \includegraphics[width=\linewidth]{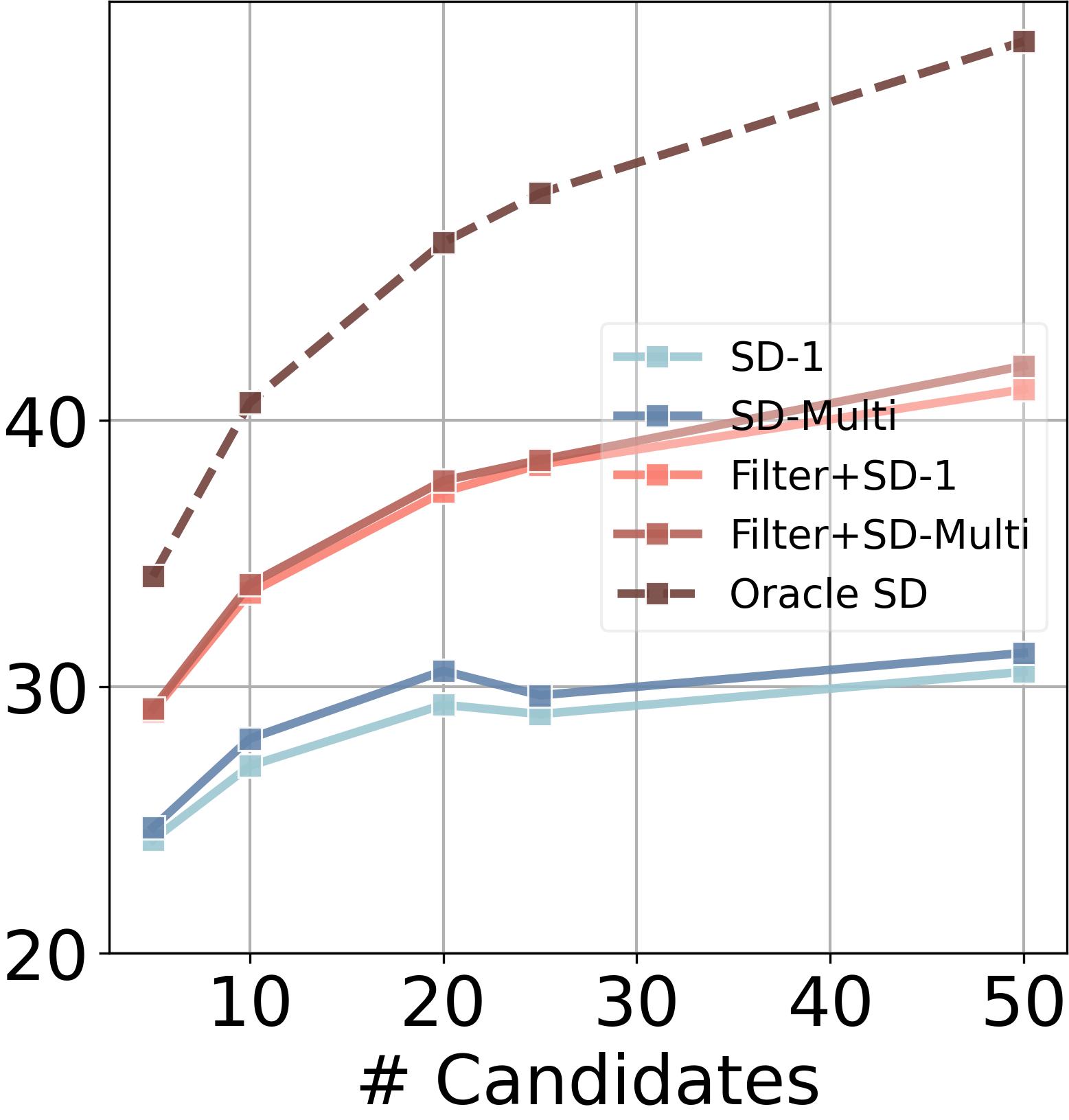}
        \caption{DS-6.7B}
        \label{fig:ds_6.7b_sd_candidates_lcb}
    \end{subfigure}\hfill
    \begin{subfigure}{0.237\textwidth}
        \centering
        \includegraphics[width=\linewidth]{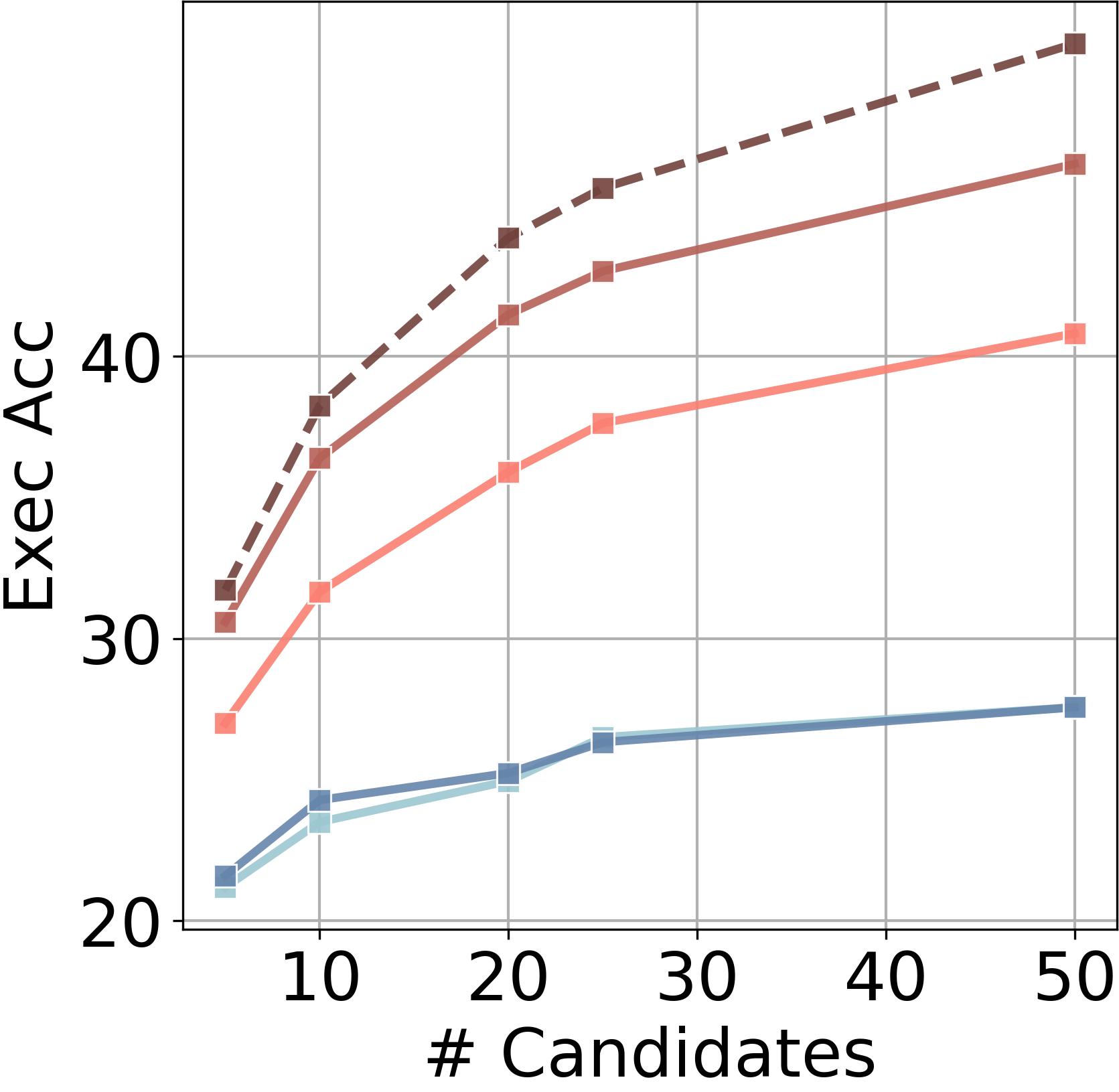}
        \caption{CL-13B}
        \label{fig:cl_13b_sd_candidates_lcb}
    \end{subfigure}\hfill
    \begin{subfigure}{0.22\textwidth}
        \centering
        \includegraphics[width=\linewidth]{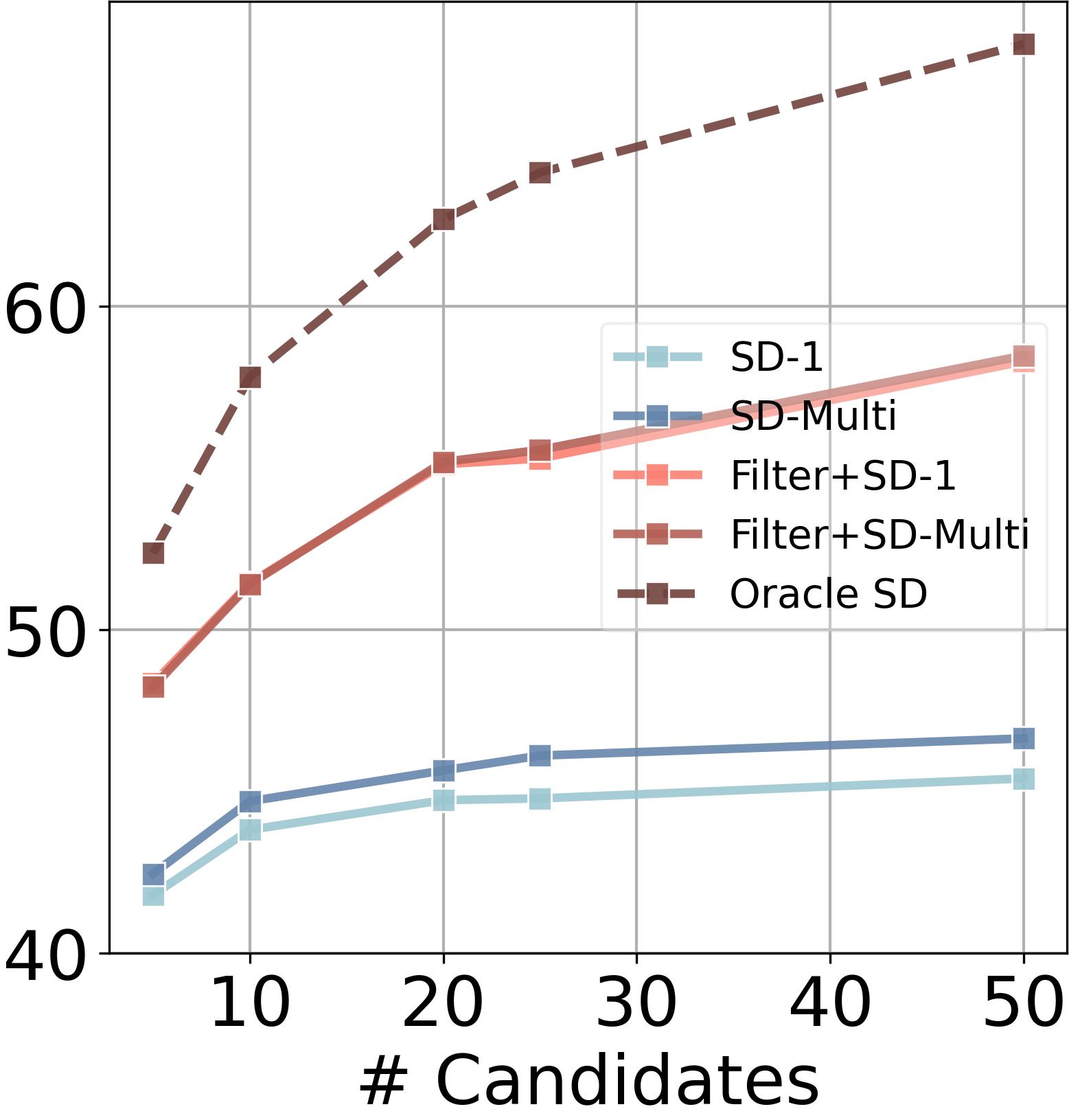}
        \caption{DS-V2-Lite}
        \label{fig:ds_16b_sd_candidates_lcb}
    \end{subfigure}

    \caption{Comparison of self-debugging methods over different numbers of candidates on LiveCodeBench, and debugged over $\{1,\textbf{Multi}\}$ candidates using the same LLM. We also provide the oracle after self-debugging for one round. Note that results ending with $+$ mean it is evaluated on the plus with extended test cases. Results are averaged across at least 2 runs.
    }
    \label{fig:lcb_sd_candidates}
\end{figure*}

\section{Dataset Statistics}\label{sec:dataset_stats}

We use the HumanEval \citep{codex} and MBPP \citep{mbpp} datasets in EvalPlus \citep{evalplus}, consisting of 164 and 395 problems respectively. For LiveCodeBench \citep{livecodebench}, we use the version that includes competitive programming problems from July 2023 to September 2024 to balance the risk of data contamination and size of the dataset. We adopt a simpler evaluation by eliminating all test cases with expected output being extreme values and set 20 unit test cases per problem at most. For easy implementation, we only include problems presented as function completion, obtaining 283 problems, among which 88 are easy problems, 135 are medium-level problems, and 60 are hard problems. 

\section{Experimental Results}\label{sec:complete_results}

\subsection{Candidate Generation}

We present the effect of the number of candidates with candidates generated by CodeLlama-\{7,13\}B-Instruct (see Figure~\ref{fig:cl_7b_num_candidates} and Figure~\ref{fig:cl_13b_num_candidates}) DeepSeekCoder-\{6.7B, V2-Lite\}-Instruct (see Figure~\ref{fig:ds_6.7b_num_candidates} and Figure~\ref{fig:ds_16b_num_candidates}) on both the original unit tests and extended test cases in EvalPlus \citep{evalplus}.  We also provide all cases for LiveCodeBench (see Figure~\ref{fig:lcb_num_candidates}). Most results align with those we present in Figure~\ref{fig:num_candidates}, where filtering based on trial unit tests improves MBR-Exec considerably and approaches the oracle performance when combined with MBR-Exec. The only exception happens on candidates generated by DeepSeekCoder-6.7B-Instruct on HumanEval (see Figure~\ref{fig:ds_6.7b_candidates_humaneval_base}) where the execution accuracy using MBR-Exec with filtering on no less than 10 candidates is already above 90.

To validate our choice of temperature, we also present the choice of sampling temperature of CodeLlama-7B-Instruct on both the original and extended test cases in Figure~\ref{fig:cl7b_temperature}. For CodeLlama-13B-Instruct and DeepSeekCoder-6.7B-Instruct, we compare results with our choice of temperature for further experiments with results using temperature 0.8. Results are presented in Table~\ref{tab:compare_temp}. All models we experiment with allow a sampling temperature over 1, with lower mean execution accuracy but higher oracle performance. Combined with filtering on trial unit tests, MBR-Exec allows constant improvement in execution accuracy when sampling with higher temperatures, which is not guaranteed without filtering.

\subsection{Comparing Reranking Methods}

We compare reranking methods with candidates generated by DeepSeekCoder-6.7B-Instruct (see Table~\ref{tab:ds_6.7b_compare_reranking}) and CodeLlama-13B-Instruct (see Table~\ref{tab:cl_13b_compare_reranking}). Findings are similar to Section~\ref{sec:compare_reranking}, where filtering based on trial unit tests is important itself and helps to boost other reranking methods, and MBR-Exec remains the best-performing method, giving close-to-oracle performance when combined with filtering. Log-likelihoods of the generated candidate and the input do not necessarily help better reranking.

\subsection{MBR-Exec with Fewer Unit Tests}

We present results using fewer unit tests using candidates generated from DeepSeekCoder-6.7B-Instruct Figure~\ref{fig:ds_6.7b_num_uts} and CodeLlama-13B-Instruct in Figure~\ref{fig:cl_13b_num_uts}. Similar to Figure~\ref{fig:num_uts}, we only need 20 unit test
cases for HumanEval+, and 5 unit test cases for MBPP-S+ to obtain optimal performance with MBR-Exec.

\subsection{Improving Oracle with Self-Debugging}

We first present results of improvement of \textit{Pass@k} that estimates oracle improvement using candidates generated and self-debugged by CodeLlama-7B-Instruct (see Figure~\ref{fig:cl_7b_sd_upperbound}), DeepSeekCoder-6.7B-Instruct (see Figure~\ref{fig:ds_6.7b_sd_upperbound}), CodeLlama-13B-Instruct (see Figure~\ref{fig:cl_13b_sd_upperbound}), and DeepSeekCoder-V2-Lite-Instruct (see Figure~\ref{fig:ds_16b_sd_upperbound}). We also provide cases for LiveCodeBench (see Figure~\ref{fig:lcb_sd_upperbound}). Our findings align with Section~\ref{sec:sd} as one round of self-debugging is enough to improve the oracle.

We then present results over different numbers of candidates generated and debugged by CodeLlama-7B-Instruct (see Figure~\ref{fig:cl_7b_sd_candidates}), DeepSeekCoder-6.7B-Instruct (see Figure~\ref{fig:ds_6.7b_sd_candidates}), CodeLlama-13B-Instruct (see TFigure~\ref{fig:cl_13b_sd_candidates}), and DeepSeekCoder-V2-Lite-Instruct (see Figure~\ref{fig:ds_16b_sd_candidates}). We also provide all cases for LiveCodeBench (see Figure~\ref{fig:lcb_sd_candidates}). We find that SD-Multi outperforms SD-1 constantly except for HumanEval(+) with candidates generated and debugged by DeepSeekCoder-6.7B-Instruct with 25 generated candidates.

\end{document}